\documentclass{article}

\PassOptionsToPackage{numbers, compress}{natbib}

\usepackage{adjustbox}  
\usepackage{amsmath}
\usepackage[pagebackref=true,breaklinks=true,colorlinks,bookmarks=false,citecolor=citecolor,linkcolor=linkcolor,urlcolor=gray]{hyperref}
\usepackage{cleveref}
\usepackage{multirow}
\usepackage{booktabs}
\usepackage{iclr2027_conference,times}

\usepackage[most]{tcolorbox}

\iclrfinalcopy
\usepackage{tikz} % draw timeline
\usepackage[utf8]{inputenc} % allow utf-8 input
\usepackage[T1]{fontenc}    % use 8-bit T1 fonts
\usepackage{url}            % simple URL typesetting
\usepackage{microtype}      % microtypography        
\usepackage{float}
\usepackage{algorithm}
\usepackage{enumitem}
\usepackage{listings}
\usepackage{wrapfig}
\usepackage{colortbl}
\usepackage{verbatim}

\usepackage{booktabs}   
\usepackage{caption} 
\usepackage{subcaption}
\usepackage{graphicx}
\usepackage{tabularx}

\usepackage{amsthm}
\usepackage{amssymb}
\usepackage{amsfonts}       % blackboard math symbols
\usepackage{mathtools}
\usepackage{xcolor}

\graphicspath{{./}{drifting/paper/}}

\usepackage{centernot}

\definecolor{citecolor}{HTML}{0071BC}
\definecolor{linkcolor}{HTML}{ED1C24}
\definecolor{lightgraybg}{gray}{0.95}
\definecolor{lightpurplebg}{HTML}{F0E6FF}
\definecolor{depthred}{HTML}{B02418}
\definecolor{myorange}{RGB}{255,127,0} % A nice orange
\definecolor{myblue}{RGB}{0,114,189}  % A nice blue
\definecolor{linkcolor}{HTML}{506B96}
\definecolor{linkpink}{HTML}{B02418}
\definecolor{propblue}{HTML}{1E4E8C}
\definecolor{brown}{HTML}{b84e1d}
\definecolor{lightgrayrow}{gray}{0.94}

\newtcolorbox{backgroundboxx}[1]{
  colback=depthred!3,
  colframe=depthred,
  title=\textbf{#1},
  fonttitle=\small,
  fontupper=\small,
  boxrule=0.8pt,
  arc=2pt,
  left=4pt,
  right=4pt,
  top=4pt,
  bottom=4pt
}

\usetikzlibrary{arrows.meta, positioning, calc}

\usepackage{tikz}
\usetikzlibrary{positioning,calc}

\hypersetup{
    colorlinks=true,
    linkcolor=brown,   % internal links (sections, equations, refs)
    citecolor=linkcolor,   % citation links
    urlcolor=linkpink     % URLs
}

\lstdefinestyle{pseudocode}{
  basicstyle=\ttfamily\small,
  keywordstyle=\color{blue}\bfseries,
  commentstyle=\color{green!60!black},
  stringstyle=\color{orange},
  numbers=left,
  numberstyle=\tiny\color{gray},
  breaklines=true,
  frame=single,
  backgroundcolor=\color{gray!10},
}

\definecolor{ourrow}{HTML}{EAF4FF}

\usepackage{textcase}
\usepackage[most]{tcolorbox}
\usepackage{algorithmicx}
\usepackage{algpseudocode}

\usepackage[scaled=0.92]{inconsolata} % nicer code font
\usepackage{array}
\tcbuselibrary{skins}
\usepackage{fontawesome5}

\usepackage{xspace}

\usepackage{pifont}

\newcommand{\cmark}{\ding{51}}  % ✓
\newcommand{\xmark}{\ding{55}}  % ✗

\def\z{\ensuremath{\mathbf{z}}}
\def\x{\ensuremath{\mathbf{x}}}

\title{\parbox{\textwidth}{\centering Depth as Time in One-Step Generative Models\par}}

\author{%
  \parbox{\textwidth}{%
    \normalfont\mdseries
    {\bfseries
      Arnold Caleb Asiimwe \hspace{0.4em}
      William Yang \hspace{0.4em}
      Sanghyuk Chun \hspace{0.4em}
      Esin Tureci \hspace{0.4em}
      Olga Russakovsky}\\[0.9em]
    \makebox[\linewidth][c]{\normalsize Princeton University}%
  }%
}
\begin{document}

\maketitle
\setcounter{footnote}{1}

\begin{abstract}
\vspace{-.5em}

The recent wave of one-step generative models, which compress the multi-step trajectory of diffusion via either distillation or learned flow maps, has reached an inflection point where they can generate high-quality images. 
Here, we ask a natural question that follows from these advances: what happens to the denoising trajectory of multi-step diffusion when generation is compressed into a single forward pass? 
We offer an empirical observation we call \textit{depth as time}: the denoising computation that multi-step diffusion performs across sampling steps appears to unfold across the depth of a single forward pass, and can be recovered by decoding intermediate layers with the model's own output head. 
Most interestingly, we show that this depthwise computation depends on the transport task a flow map is trained to solve. The most surprising case is MeanFlow, where probing shorter transport intervals reveals both denoising and renoising within a single network evaluation. In contrast, generators trained without a time-indexed transport task, such as drifting models, do not exhibit the same depthwise denoising.
Consequently, we show that models that exhibit the depthwise denoising phenomenon are more compressible across the layerwise computation: a MeanFlow \texttt{SiT-L/2} model can be compressed by $16.6\times$ in parameters into a single time-conditioned block.
% We offer an explanation for this denoise-then-renoise behavior and show that, when we treat the layerwise computation explicitly as a flow, a single time-conditioned block can be trained to denoise across layers, compressing a MeanFlow \texttt{SiT-L/2} model by $16.6\times$ in parameters.
%Finally, we show that the same insights from accelerating generative models can similarly accelerate this layerwise computation. 
Together, these results suggest that the temporal computation of diffusion is not eliminated by one-step generation, but reorganized across network depth. \href{https://github.com/princetonvisualai/depth-as-time}{Code} is available.

\end{abstract}

\begingroup
\renewcommand\thefootnote{}
\renewcommand\thefootnote{}
\renewcommand\thefootnote{}
\renewcommand\thefootnote{}
% \footnotetext{%
%   \hspace*{1.6em}%
%   \scriptsize
%   \begin{tabular}[t]{@{}l@{}}
%   Correspondence:
%   \texttt{asiimwe@cs.princeton.edu}
%   \end{tabular}%
% }
\addtocounter{footnote}{-1}
\endgroup

\section{Introduction}
% \vspace{-1.5em}

% \noindent

% \begin{figure}[H]
%   \centering

%   % Shorter image on the left, with caption below
%   \begin{minipage}[t]{0.5\linewidth}
%     \vspace{10pt}
%     \centering
%     \includegraphics[width=\linewidth]{figs/concept.pdf}

%     \caption{\textbf{Depth as time.} \textit{Left:} Multi-step diffusion loops a network over repeated calls, exposing denoising across sampling steps, whereas a one-step generator replaces this loop with a single forward pass through blocks $\mathbf{B}_1,\dots,\mathbf{B}_L$. \textit{Right:} Predictions decoded from intermediate single blocks (sb) using the model's own output readout. For a multi-step model (top; each row is one of six network calls), predictions change comparatively little across depth within each call, and denoising instead unfolds down the output column across calls. For a one-step model (bottom), a similar coarse-to-fine progression unfolds across depth within a single forward pass.}
%     \label{fig:depth-as-time}
%   \end{minipage}%
%   \hspace{0.01\linewidth}%
%   % Taller image on the right
%   \begin{minipage}[t]{0.49\linewidth}
%     \vspace{-10pt}
%     \centering
%     \includegraphics[width=\linewidth]{figs/teaser2.pdf}
%   \end{minipage}

%   \vspace{-1em}
% \end{figure}

One-step generative models based on diffusion have recently been shown to produce high-quality images without the iterative denoising trajectory of multi-step diffusion, using methods ranging from distillation~\citep{salimans2022progressive,luo2023diff,yin2024one,zhou2024score} to directly learning flow maps~\citep{song2023consistency,frans2025one,boffi2025flow,geng2025mean}. This efficiency, however, seems to remove a defining feature of diffusion- and flow-based generation: the explicit trajectory of intermediate states that evolves from coarse to fine-grained detail. This trajectory provides a natural axis for intervention and control via guidance methods that modify the generative vector field depending on the desired goal. Several downstream advantages of multi-step diffusion have followed, ranging from classifier-free guidance~\citep{ho2022classifier,wang2026cfg}, reinforcement-learning-based control~\citep{black2024training,wallace2024diffusion,fan2023dpok}, and image editing~\citep{meng2021sdedit,lee2026low,liang2026personalized} to classification~\citep{li2023your,clark2023text}, fine-grained generation~\citep{guo2026thinking,zarei2026slideredit}, and many others.

It is natural, therefore, to ask what happens to this denoising trajectory when generation is compressed into a single forward pass. We show that distilled one-step models and flow maps effectively \emph{unroll} this loop across network depth, and that decoding their intermediate blocks reveals a coarse-to-fine progression reminiscent of the one multi-step diffusion exposes across sampling steps. We call this observation \textit{depth as time}. In contrast, one-step models not based on diffusion, such as drifting models~\citep{deng2026generative}, do not appear to exhibit the same transport-organized depthwise trajectory.

Surprisingly, depth as time can be observed without any additional training. Applying the model's own output head to its intermediate representations recovers a layerwise velocity (Sec.~\ref{subsec:layer-comparisons}), from which we decode the image at each depth and reveal a progressive, denoising-like pattern (Sec.~\ref{subsec:first-evidence-denoising}).

Importantly, the depthwise trajectory depends on the transport task the model is asked to solve at inference. This becomes especially clear when the model generates a noisy intermediate $\x_r$ rather than the clean endpoint $\x_0$. Distilled one-step models under multi-step sampling nest depth as time within sampling time, with each sampler step carrying its own depthwise refinement (Sec.~\ref{subsec:distillation}). Flow maps conditioned on the transport endpoint such as MeanFlow~\citep{geng2025mean} and Shortcut models~\citep{frans2025one} instead exhibit a non-monotonic \emph{denoise-then-renoise} pattern, first revealing cleaner image structure before renoising toward the requested noisy endpoint (Sec.~\ref{subsec:flow_map}).

Finally, in Sec.~\ref{sec:explicit_depth_as_time}, we treat depth as time explicitly by training a single time-conditioned block to denoise across layers, yielding an even more compressed one-step generator. For example, we compress a MeanFlow \texttt{SiT-L/2} one-step model for ImageNet-$256 \times 256$ generation by $16.6\times$ (from $459$M to $27.6$M parameters) at FID $11.7$, which improves to $4.9$ with FD-loss post-training~\citep{yang2026representation}, compared with $4.0$ for the original model. Performing the same compression on a drifting model, by contrast, degrades it to a FID of $47.1$.

\section{Preliminaries: One-step Generative Models}
\label{sec:preliminary}

Diffusion- or flow-based generative models transport a simple noise distribution toward the data distribution along a time-indexed stochastic or deterministic process~\citep{sohl2015deep,song2020score,liu2022flow,lipman2022flow,albergo2023stochastic}. In the ordinary differential equation (ODE) formulation, the learned vector field specifies the instantaneous velocity with which samples evolve along this transport, and sampling amounts to numerically integrating it, conventionally requiring repeated network evaluations. A long-standing goal has therefore been to approximate this transport using only a few network evaluations---or even a single one---while preserving sample quality. Existing approaches broadly include distillation-based methods, which transfer the behavior of a multi-step teacher to a faster student through flow-map or distribution-level objectives~\citep{salimans2022progressive,luo2023diff,yin2024one,zhou2024score,sauer2024fast,sauer2024adversarial}, and methods that learn few-step models (or flow maps) from scratch~\citep{song2023consistency,frans2025one,boffi2025flow,geng2025mean,zhou2025inductive}.

Flow maps~\citep{boffi2025flow}, in particular, provide a useful description of recent one-step generative models based on diffusion. A flow map $\Phi_{t\rightarrow r}$ denotes the transport induced by integrating the ODE from time $t$ to time $r$. Throughout, we use transport task to refer to the specific map $\Phi_{t\rightarrow r}$ that a network is trained to approximate, i.e., the samplewise correspondence between noise level $t$ and endpoint $r$ that its output must realize; this is distinct from the downstream task (e.g., class-conditional generation) that the model serves. Learned flow maps can be classified by their transport task: approximating $\Phi_{t\rightarrow 0}(\x_t)$ for~\citet{song2023consistency} and \citet{ect}, $\Phi_{t\rightarrow r}(\x_t)$ for~\citet{kim2024consistency}, \citet{geng2025mean}, and \citet{zhou2025inductive}, and $\Phi_{t\rightarrow t-d}(\x_t)$ for~\citet{frans2025one}. The corresponding task conditioning $\boldsymbol{\tau}_\mathrm{task}$ can thus be boundary-anchored ($t\rightarrow 0$), interval-conditioned ($t\rightarrow r$), or step-conditioned ($t\rightarrow t-d$), where the latter two are alternative parameterizations related by $r=t-d$.

For simplicity, we denote the $\boldsymbol{\tau}_\mathrm{task}$-indexed network as follows, where its output may parameterize the transport map directly or through a velocity or displacement:
\[
f_\theta(\x_t; \boldsymbol{\tau}_\mathrm{task})
\quad\text{where}\quad
\boldsymbol{\tau}_\mathrm{task} =
\begin{cases}
t, & \text{e.g., } \text{consistency models \citep{song2023consistency}},\\
(t, r), & \text{e.g., } \text{MeanFlows \citep{geng2025mean}},\\
% \text{e.g., } \text{CTMs \citep{kim2024consistency}, IMM \citep{zhou2025inductive}, MeanFlow \citep{geng2025mean}},\\
(t, d), & \text{e.g., } \text{Shortcut Models \citep{frans2025one}}.
\end{cases}
\]
Here, $\x_t$ denotes the network input, $t$ the current noise level, $r$ the desired endpoint time of the transport, and $d$ the desired step size. Standard diffusion networks are also conditioned on $t$, although this conditioning alone does not specify a boundary-anchored transport map. 

We contrast these models with methods that lack an explicit time-indexed transport map at inference, e.g., drifting models~\citep{deng2026generative},
% and generative adversarial networks (GANs)~\citep{goodfellow2014generative},
whose objectives constrain the distribution of generated samples without any time-indexed transport task.
% , and generators trained by minimizing the maximum mean discrepancy (MMD) between generated and real data distributions~\citep{dziugaite2015training}. 
We use them as negative controls to distinguish generic layerwise refinement from computation organized by a $\boldsymbol{\tau}_\mathrm{task}$-indexed transport task. We provide additional details on one-step generation in Appendix~\ref{app:extra_background}.

\section{Depth as Time}
\label{sec:depth_as_time}

We introduce \emph{depth as time}, an empirical observation in which denoising-like structure unfolds across the layers of a one-step generative model. We begin by introducing our methodology for comparing the intermediate temporal outputs of a multi-step flow model with the intermediate layer-wise outputs of a one-step compressed model (Sec.~\ref{subsec:layer-comparisons}). We then use this methodology to demonstrate the first evidence of progressive denoising across layers (Sec.~\ref{subsec:first-evidence-denoising}), drawing three conclusions. First, the one-step layers exhibit patterns of denoising, which we call \emph{depth as time}. Second, the recovered layer-wise mixing schedule in one-step models is different from the standard linear schedule in the temporal progression of multi-step models. Third, we demonstrate that drifting ($\tau$-independent) models do not appear to exhibit depth as time unlike $\tau_\mathrm{task}$-conditioned models such as MeanFlows. 

% \olga{(a) any chance we have results across multiple models of each type? 
% (b) i see you're thinking of adding drifting comparison to fig 2 -- yes, that would be great, because it's not clear now why we wait until the last section to do the comparison with drifting, we can make that observation earlier in 3.2 \arnold{done}
% }

\subsection{From sampling time to network depth}
\label{subsec:layer-comparisons}

% Multi-step flow models and one-step (compressed) generative models have fundamentally different representations, so we begin by introducing a methodology for comparing their intermediate representations.
Multi-step flow models expose intermediate states across sampling time, whereas one-step generative models expose intermediate representations only across network depth. We therefore introduce a methodology for comparing these two forms of progression.

Consider a multi-step flow model that transports a noisy sample $\boldsymbol{\x}_t$ toward a clean sample $\boldsymbol{\x}_0$ by following a learned velocity field. Flow matching~\citep{lipman2022flow,albergo2023stochastic} defines a probability path between data and noise, e.g., $\boldsymbol{\x}_t=\alpha_t\boldsymbol{\x}_0+\sigma_t\boldsymbol{\epsilon}$, where $\boldsymbol{\epsilon}\sim\mathcal{N}(\mathbf{0},\mathbf{I})$ and $t\in[0,1]$. Under the linear interpolation considered here, $\alpha_t=1-t$ and $\sigma_t=t$, such that $t=0$ corresponds to data and $t=1$ corresponds to noise. Each sampled interpolation has velocity $\boldsymbol{\epsilon}-\boldsymbol{\x}_0$, and the corresponding marginal velocity field is $\boldsymbol{u}_t(\boldsymbol{x})=\mathbb{E}[\boldsymbol{\epsilon}-\boldsymbol{\x}_0\mid\boldsymbol{\x}_t=\boldsymbol{x}]$. The model is trained to approximate this field with $\boldsymbol{v}_\theta(\boldsymbol{\x}_t,t)\approx\boldsymbol{u}_t(\boldsymbol{\x}_t)$. Sampling then follows the ODE $d\boldsymbol{\x}_t/dt=\boldsymbol{v}_\theta(\boldsymbol{\x}_t,t)$, and  exposes an explicit sequence of states $\boldsymbol{\x}_{t_0},\boldsymbol{\x}_{t_1},\ldots,\boldsymbol{\x}_{t_K}$ across sampling time.

A one-step generator, by contrast, produces its output in a single network call and exposes no such external trajectory. It instead computes a sequence of hidden states \textcolor[HTML]{B02418}{$\mathbf{h}$} across network depth. Concretely, starting from a noisy input $\mathbf{z}=\boldsymbol{\x}_t$ from time step $t$, and leveraging  the task-conditioning variable $\textcolor[HTML]{000000}{\boldsymbol{\tau}_\mathrm{task}}$ and class conditioning $c$, a one-step generator with a dimension-preserving residual stream produces:
\begin{equation}
\textcolor[HTML]{B02418}{\mathbf{h}_0}
=
\mathbf{E}(\mathbf{z}),
\qquad
\textcolor[HTML]{B02418}{\mathbf{h}_{i+1}}
=
\mathbf{B}_{i+1}
\left(
\textcolor[HTML]{B02418}{\mathbf{h}_i};
\textcolor[HTML]{000000}{\boldsymbol{\tau}_\mathrm{task}}, c
\right),
\qquad
i=0,\ldots,L-1,
\label{eq:hidden_states_depth}
\end{equation}
where $\mathbf{E}$ is the input embedding, $\mathbf{B}_{i+1}$ denotes the $(i+1)$-th network block, and $L$ is the total number of blocks. Equivalently, the representation at depth $i$ is obtained by composing the first $i$ blocks, $\textcolor[HTML]{B02418}{\mathbf{h}_i}=(\mathbf{B}_i\circ\cdots\circ\mathbf{B}_1)(\textcolor[HTML]{B02418}{\mathbf{h}_0})$, with the same conditioning supplied to each block. For a velocity-parameterized generator, the model's final prediction is $\boldsymbol{v}_\theta(\z,\textcolor[HTML]{000000}{\boldsymbol{\tau}_\mathrm{task}},c)=\mathbf{W}_{\mathrm{out}}(\textcolor[HTML]{B02418}{\mathbf{h}_L};\boldsymbol{\tau}_\mathrm{task},c)$, where $\mathbf{W}_{\mathrm{out}}$ denotes the network head, which includes normalization, adaptive layer normalization modulation (AdaLN), and a linear projection. Depending on the model, this output represents either an instantaneous velocity or an interval-averaged velocity.

\textbf{Probe.} To probe how the model's prediction develops across depth, we apply the same final head to the hidden state after each block, with the original conditioning and without executing the remaining blocks $\mathbf{B}_{i+1},\ldots,\mathbf{B}_L$. For layer $i$, we define
\begin{equation}
\displaystyle
\textcolor[HTML]{B02418}{\boldsymbol{v}_i}
\coloneqq
\mathbf{W}_{\mathrm{out}}
\left(
\textcolor[HTML]{B02418}{\mathbf{h}_i};
\boldsymbol{\tau}_\mathrm{task},c
\right).
\label{eq:layer_velocity_readout}
\end{equation}
At $i=L$, this recovers the model's ordinary output, $\boldsymbol{v}_L=\boldsymbol{v}_\theta(\z,\textcolor[HTML]{000000}{\boldsymbol{\tau}_\mathrm{task}},c)$. At intermediate layers, $\boldsymbol{v}_i$ measures the velocity prediction accessible through the output head from the residual stream at depth $i$. This probe introduces no learned parameters and does not modify the generator.\footnote{This parameter-free layerwise readout is closely related to the \emph{logit lens} in language modeling~\citep{nostalgebraist2020logitlens}. In Appendix~\ref{app:readout_controls} we demonstrate that similar conclusions can be obtained using a \emph{separately} trained linear probe, suggesting a lack of reliance on the particular form of the probe, but not by a \emph{random} untrained probe. Therefore, the observed convergence is semantically meaningful rather than simply numerical. 

% \olga{I'm not entirely sure I properly explained these findings, so please check. The main reason why I'm moving these to the appendix is because they both break the flow but also raise more questions than they answer in my mind (e.g.,  what about other random probes? a combination of random probes? but why...). It kind of feels like we either need to do more here, or less here, and since it's somewhat orthogonal to our main story my instinct is to put it in the appendix for the curious reader and instead just keep the eye on the prize here.  } 

}

We then convert each layerwise velocity estimate into the corresponding endpoint prediction. For the linear flow-matching interpolation, the clean-image estimate is $\hat{\boldsymbol{\x}}_0=\boldsymbol{\x}_t-t\,\boldsymbol{v}_\theta(\boldsymbol{\x}_t,t)$; with the exact marginal velocity field, which recovers the posterior mean $\mathbb{E}[\boldsymbol{\x}_0\mid\boldsymbol{\x}_t]$. More generally, a velocity-parameterized one-step model predicts an endpoint at time $r$ through
$
\hat{\boldsymbol{\x}}_r
=
\boldsymbol{\x}_t
-
(t-r)\,\boldsymbol{v}_\theta(\boldsymbol{\x}_t,\boldsymbol{\tau}_\mathrm{task},c).
$
For an instantaneous velocity prediction, this is a single Euler step, rather than an exact integration of the ODE. For an interval-averaged velocity prediction, it is the model's finite-interval transport parameterization as in~\cite{geng2025mean}.

Applying this conversion to the layerwise readouts for a one-step input $\boldsymbol{\x}_t$ gives our layerwise endpoint probe,
\begin{equation}
\textcolor[HTML]{B02418}{\tilde{\boldsymbol{\x}}_r^{i}}
=
\boldsymbol{\x}_t
-
(t-r)\,\textcolor[HTML]{B02418}{\boldsymbol{v}_i}.
\label{eq:xr}
\end{equation}
Here, $r$ specifies the endpoint being decoded: $r=0$ gives the clean-image predictions $\textcolor[HTML]{B02418}{\tilde{\boldsymbol{\x}}_0^{i}}$, while $0<r<t$ gives a noisy intermediate predictions $\textcolor[HTML]{B02418}{\tilde{\boldsymbol{\x}}_r^{i}}$ at different depths $i$. At the final layer, $\textcolor[HTML]{B02418}{\tilde{\boldsymbol{\x}}_r^{L}}$ is the endpoint predicted by the full network for the corresponding transport problem. Simply put, $\textcolor[HTML]{B02418}{\tilde{\boldsymbol{\x}}_r^{i}}$ asks what endpoint prediction is already implied by the partially formed residual representation $\textcolor[HTML]{B02418}{\mathbf{h}_i}$ at depth $i$. Interestingly, these layerwise velocities $\textcolor[HTML]{B02418}{\boldsymbol{v}_i}$ used to construct $\textcolor[HTML]{B02418}{\tilde{\boldsymbol{\x}}_r^{i}}$ have conceptual connections to Neural ODE views of neural networks~\citep{chen2018neural} and deep equilibrium models~\citep{deq}; we discuss these connections in Appendix~\ref{app:extended_related_works}. 

% \olga{I think this section is becoming a bit long and some of this discussion can be shortened -- but I left it as-is because it's interesting and fairly linear, so if we're not short on space it can stay.}

\subsection{Evidence of progressive denoising across layers}
\label{subsec:first-evidence-denoising}

\begin{figure}[t]
  \centering
  \includegraphics[width=\linewidth]{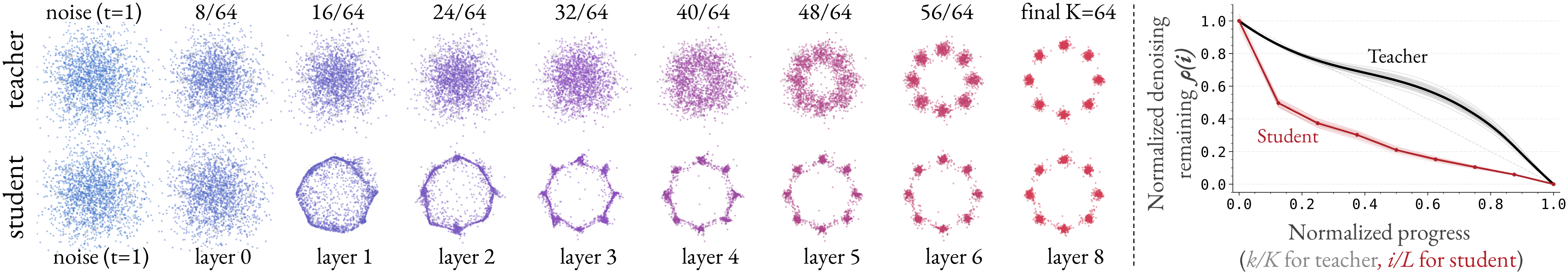}
  \vspace{-1em}
  \caption{\textbf{A one-step toy student distilled from a multi-step teacher appears to denoise across depth on the 8-Gaussians toy example.} Normalized distance (right) shows that, although both progress toward the same endpoint, the student's depthwise progression need not trace the teacher's sampling-time trajectory. $k/K$ is the normalized sampling step for the teacher, here $K=64$ steps.
  % \wy{description of the top half is missing \arnold{done}}
  }
  \label{fig:toy_depth_as_time}
\end{figure}

Using the probe methodology of Sec.~\ref{subsec:layer-comparisons}, we begin by demonstrating the emergence of depth-as-time in simple toy examples. 
% \olga{it would be reasonable to add just a bit more details here about how this experiment was set up and conducted, e.g., implementation details \arnold{done}}. 
Specifically, after progressively distilling~\citep{salimans2022progressive} a multi-step 8-Gaussians teacher into a one-step student, we plot the student's layerwise endpoint predictions $\textcolor[HTML]{B02418}{\tilde{\x}_0^i}$; for the teacher, we plot $\hat{\x}_0$-predictions for each step along its multi-step sampling trajectory in Figure~\ref{fig:toy_depth_as_time}. The teacher $\x_0$-predictions are external and therefore different from internal layerwise $\textcolor[HTML]{B02418}{\tilde{\x}_0^i}$ predictions. We also plot the normalized distance to the endpoint using Equation~\eqref{eq:depth_schedule}.

\begin{figure}[t]
  \centering
  \includegraphics[width=\linewidth]{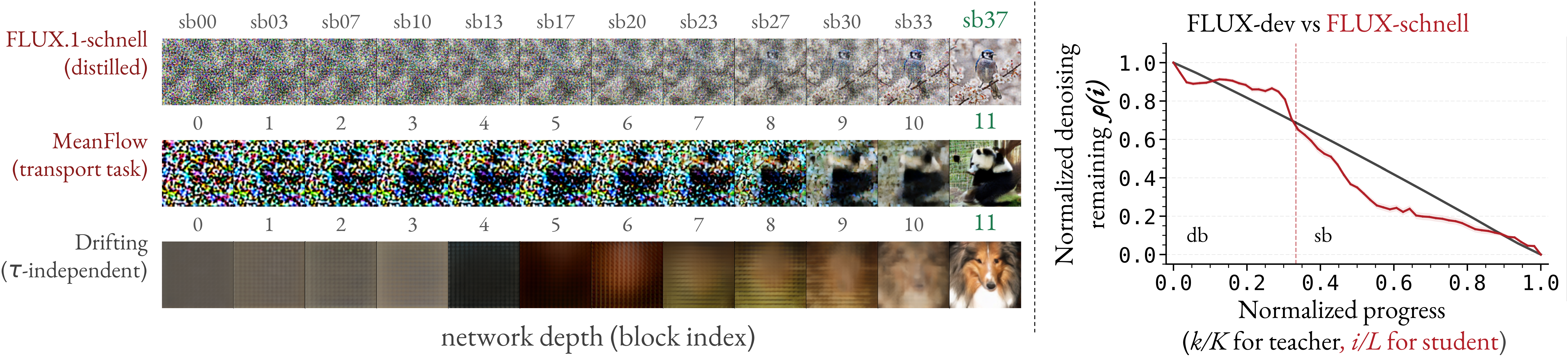}
  \caption{ \textbf{Depthwise denoising-like progression persists in large-scale one-step image generators.} Layerwise endpoint predictions from FLUX-schnell and MeanFlow progressively reveal cleaner image structure across network depth, while Drifting provides a contrasting $\tau$-independent model. For FLUX-schnell, normalized endpoint distance decreases similarly under the model's output head and an independently trained probe, whereas a random probe does not reproduce the same progression.
}
  \label{fig:scheduler}
  \vspace{-1em}
\end{figure}

The toy example in Figure~\ref{fig:toy_depth_as_time} exhibits layerwise refinement suggestive of denoising, so we next ask whether this phenomenon extends to large-scale one-step generators. We apply Equation~\eqref{eq:layer_velocity_readout} to recent one-step generative models trained with different objectives, including distillation (specifically, latent adversarial distillation for FLUX-schnell~\citep{labs2025flux}), MeanFlow~\citep{geng2025mean}, and Drifting~\citep{deng2026generative}, and visualize the resulting layerwise predictions $\textcolor[HTML]{B02418}{\tilde{\x}_0^i}$ in Figure~\ref{fig:scheduler}. For MeanFlow and Drifting, we use the released \texttt{B/2} checkpoints (12 layers, width 768) trained on ImageNet-$256\times256$~\citep{russakovsky2015imagenet}. 

Qualitatively, FLUX-schnell and MeanFlow (both $\tau_\mathrm{task}$-conditioned models; Sec.~\ref{sec:preliminary}) exhibit progressive refinement across depth. This progression is also reflected quantitatively by the approximately monotonic decrease in normalized endpoint distance from Equation~\eqref{eq:depth_schedule}. In contrast, the Drifting model does not exhibit the same systematic depthwise refinement.

%We next ask whether this progression exhibits the signal--noise structure associated with denoising. 

%\subsection{Quantifying Depthwise Progression}
%\label{subsec:denoising}

%In a conventional flow model, sampling time is associated with a prescribed probability path. As introduced above, the coefficients $\textcolor[HTML]{777777}{(\alpha_t,\sigma_t)}$ specify the signal and noise contributions at each time $t$.

%Network depth comes with no analogous calibration. Although the layerwise probe produces an ordered sequence of endpoint predictions $\tilde{\x}_r^{1},\ldots,\tilde{\x}_r^{L}$, the block index $i$ by itself says only where a representation occurs in the computation. In particular, there is no reason for normalized depth $i/L$ to correspond directly to sampling time. A visually denoising-like progression therefore motivates a quantitative question: \textit{how does the structure of these predictions change across depth, and is that change consistent with denoising?}

%We address this question in two stages, focusing here on clean-endpoint predictions with $r=0$. First, we quantify layerwise refinement by measuring how closely intermediate predictions approach the model's final output. However, this alone does not establish layerwise denoising. We, secondly, ask whether intermediate predictions can be described as mixtures of the input noise and final generated image, with their relative contributions changing across depth.

%\textbf{Normalized endpoint distance.}
In addition to plotting the constructed $\textcolor[HTML]{B02418}{\tilde{\x}_0^i}$ predictions, we also quantify layerwise refinement, by measuring how far each intermediate prediction $\textcolor[HTML]{B02418}{\tilde{\x}_0^i}$ at layer $i$ (of the clean image $\x_0$) is from the model's final prediction $\tilde{\x}_0^{L}$. For a distance or dissimilarity measure $d$, we normalize this distance by its maximum along the same trajectory:
\begin{equation}
\boldsymbol{\rho}(i)
\coloneqq
\frac{
d(
\tilde{\x}_0^{i},
\tilde{\x}_0^{L}
)
}{
\max_{1\leq j\leq L}
d(
\tilde{\x}_0^{j},
\tilde{\x}_0^{L}
)
}.
\label{eq:depth_schedule}
\end{equation}
%For trajectories with a nonzero denominator, $\boldsymbol{\rho}(i)\in[0,1]$ and $\boldsymbol{\rho}(L)=0$. \olga{add a clarification on why some trajectories might be zero? }
We normalize each trajectory separately before averaging across $32$ examples, excluding trajectories for which all distances are zero. In our experiments, we use the $\ell_2$ distance. Figures~\ref{fig:toy_depth_as_time} and~\ref{fig:scheduler} show how distance evolves across intermediate layers in the toy example and for the one-step FLUX model along with their respective teachers. 
% \olga{
% since the right half of fig 3 is going in the appendix, i'd suggest putting the left graph of figure 3 carefully to the right of figure 2 -- can remove a couple of the layers/time steps for example, while making the graph more square, and putting them side-by-side. \arnold{done} 
% Alternatively, can leave fig 2 as-is but run a few other settings and put a few graphs side-by-side within figure 3 showing the differences in schedules for other settings (but do need to describe those in the text then, and also compare/contrast those graphs and what we learn from them).}

\textbf{Recovering an effective mixing schedule.}
We note that the layerwise refinement observed above, and in particular the normalized distance to the endpoint in Equation~\eqref{eq:depth_schedule}, is not by itself evidence of denoising. To test whether the layerwise decodes exhibit a denoising-like signal--noise structure, we take inspiration from the probability-path parameterization in~\cite{lipman2022flow} and solve the inverse problem of recovering depth-dependent mixing coefficients from the layerwise predictions in Figure~\ref{fig:scheduler}. The intuition is that, if a single pair of coefficients shared across trajectories can explain intermediate predictions as mixtures of the input noise and final prediction, then a denoising-like interpretation of depth is plausible.

We focus on full one-step generation from noisy input to clean output. We recover the coefficients for FLUX-schnell, a MeanFlow \texttt{B/2} model, and a Shortcut \texttt{B/2} model~\citep{frans2025one}, and contrast these with a $\tau$-independent Drifting \texttt{B/2} model.
% \olga{We need to smooth this transition from a very simple toy example to a more complex one here. The way to do that is to beef up fig 2/3 beyond toy examples, which I think you're already planning on doing. Ideally they would include the toy example, and an example from meanflow and one from drifting, so then here it makes sense why we are building out the scheduler for the two. \arnold{line 230? done}}
Specifically, for depthwise trajectory $\textcolor[HTML]{777777}{n}$ of each model, we let $\boldsymbol{\epsilon}^{\textcolor[HTML]{777777}{n}}$ denote the input noise and $\tilde{\x}_0^{L,\textcolor[HTML]{777777}{n}}$ the final generated prediction, and assume
\begin{equation}
\tilde{\x}_0^{i,n}
=
\textcolor[HTML]{B02418}{\alpha_i}\tilde{\x}_0^{L,\textcolor[HTML]{777777}{n}}
+
\textcolor[HTML]{B02418}{\sigma_i}\boldsymbol{\epsilon}^{\textcolor[HTML]{777777}{n}}
+
\boldsymbol{\delta}_i^{\textcolor[HTML]{777777}{n}},
\label{eq:depth_mixing_model}
\end{equation}
where $\boldsymbol{\delta}_i^{\textcolor[HTML]{777777}{n}}$ is the unexplained residual. We fit a single pair of coefficients shared across 256 depthwise trajectories at each block:
$
\textcolor[HTML]{B02418}{(\widehat{\alpha}_i,\widehat{\sigma}_i)}
=
\operatorname*{arg\,min}_{\alpha,\sigma}
\sum_{\textcolor[HTML]{777777}{n}\in\mathcal{D}_{\mathrm{train}}}
\|
\tilde{\x}_0^{i,\textcolor[HTML]{777777}{n}}
-
\textcolor[HTML]{B02418}{\alpha}\tilde{\x}_0^{L,\textcolor[HTML]{777777}{n}}
-
\textcolor[HTML]{B02418}{\sigma}\boldsymbol{\epsilon}^{\textcolor[HTML]{777777}{n}}
\|_2^2
\label{eq:depth_schedule_fit}
$.

\begin{figure}[t]
  \centering
  % \vspace{-0.5em}
  \includegraphics[
    width=\linewidth,
    height=0.30\textheight,
    keepaspectratio
  ]{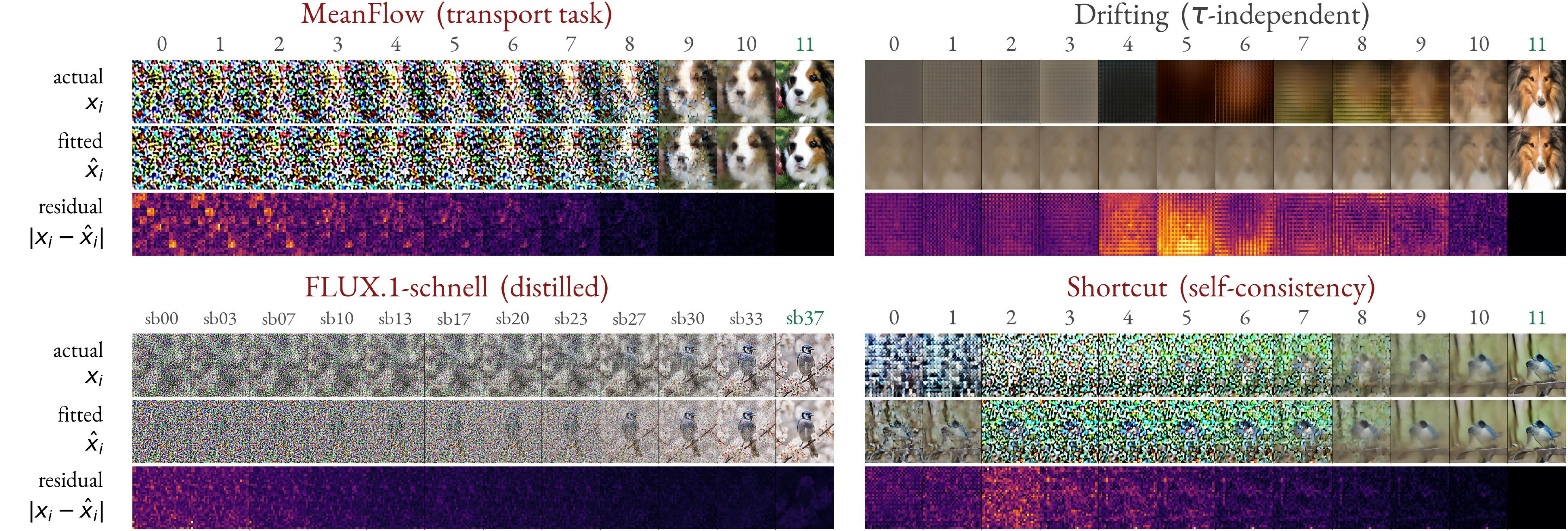}
  \caption{
  \textbf{Depthwise mixing across one-step generators.}
  A shared linear mixture of the input noise and final prediction closely reconstructs intermediate predictions for MeanFlow, FLUX-schnell, and Shortcut, with residuals decreasing across depth. In contrast, Drifting exhibits substantially larger residuals, indicating that the same two-component mixing model does not explain its depthwise trajectory as well.
  }
  \vspace{-1em}
  \label{fig:mixing_reconstructions}
\end{figure}

\begin{wrapfigure}{r}{0.5\textwidth}
  \centering
  \vspace{-1em}
  \includegraphics[width=\linewidth]{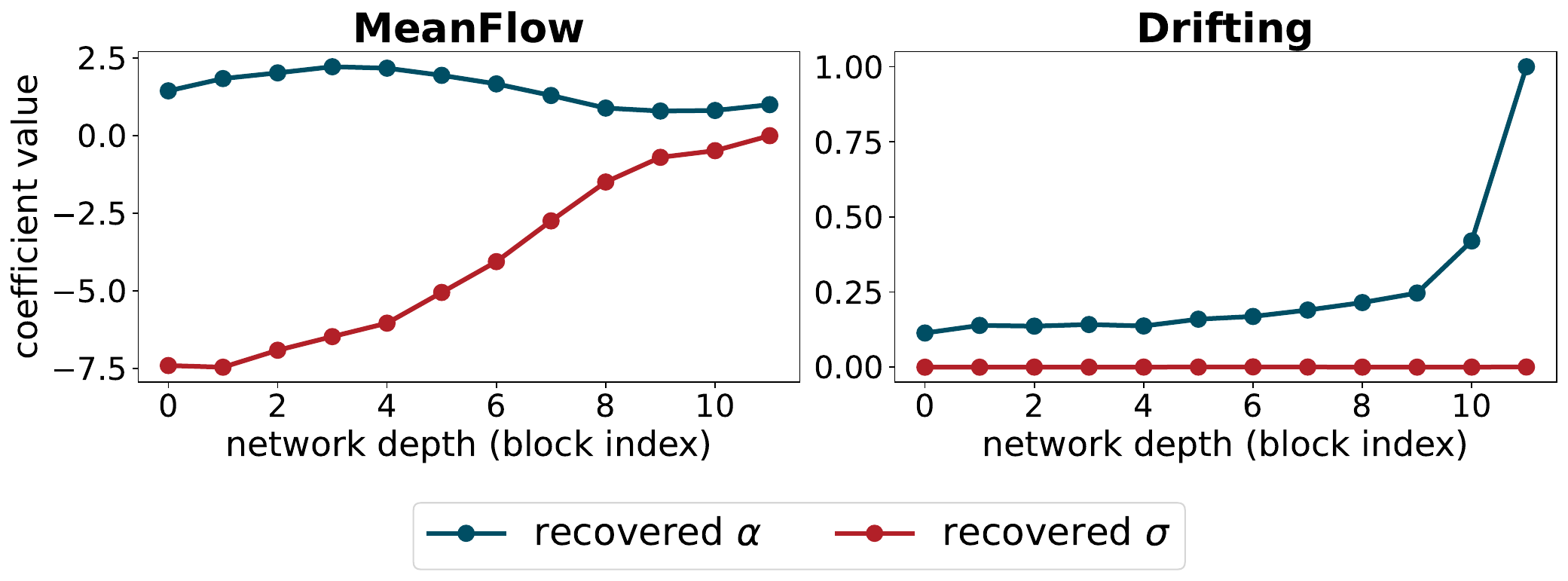}
  \vspace{-2em}
  \caption{For Meanflows, the recovered depthwise schedule is \emph{not} the flow-matching interpolation schedule. Drifting, on the other hand, maintains $\sigma_i\approx0$ at every block. }
  \label{fig:meanflow_coeffs}
  \vspace{-1.8em}
\end{wrapfigure}
The mapping $i\mapsto\textcolor[HTML]{B02418}{(\widehat{\alpha}_i,\widehat{\sigma}_i)}$ defines our \emph{effective depthwise mixing schedule}. We impose neither monotonicity nor positivity, and we do not enforce the flow-matching constraint $\textcolor[HTML]{B02418}{\alpha_i}+\textcolor[HTML]{B02418}{\sigma_i}=1$. Thus, the optimization is unconstrained and any observed difference in the signal and noise coefficients is completely attributed to the depth-as-time trajectories. We evaluate how well the fitted coefficients reconstruct decoded predictions on held-out trajectories in Figure~\ref{fig:mixing_reconstructions}.

Across the four models, Drifting stands out as having substantially larger reconstruction residuals between the observed and fitted depthwise trajectories. In fact, its recovered noise coefficient is $\sigma_i\approx0$ at every block (Figure~\ref{fig:meanflow_coeffs}), so its intermediate predictions carry no identifiable component of the input noise and do not follow a denoising-like path. In contrast, MeanFlow, FLUX-schnell, and Shortcut are each well described by a two-component signal--noise mixing model despite being trained with different objectives.

Notably, the recovered coefficients do not correspond to a standard flow-matching interpolant, for which $\alpha_i,\sigma_i\in[0,1]$ and $\alpha_i+\sigma_i=1$ along the path. For MeanFlow, for example, we recover $\alpha_4=2.18$ and $\sigma_4=-6.04$ at block 4 (Figure~\ref{fig:meanflow_coeffs}), with both coefficients outside $[0,1]$. Thus, although the low held-out reconstruction error for MeanFlow ($\approx0.10$ on average across blocks) indicates a recoverable signal--noise organization across depth, the recovered schedule is distinct from the interpolation schedule used during flow-matching training. We interpret this as depth defining its own effective denoising-like coordinate, rather than directly reproducing training time. We provide additional details and recovered schedules for the other models in Appendix~\ref{app:depthwise_mixing}.

\section{Depth As Time Generalizes to Arbitrary Transport Tasks}
\label{sec:experiments}

% A useful property of flow maps is that they can be queried for different transport tasks. Having observed \emph{depth as time} in full noise-to-clean generation, we now ask: what happens to this depthwise progression over shorter transport intervals, when the requested endpoint is still noisy? We first explore this question in distilled generators, where the sampler selects the endpoint (Sec.~\ref{subsec:distillation}), and then in interval-conditioned flow maps, where the endpoint is explicitly supplied to the network (Sec.~\ref{subsec:flow_map}).

Our findings so far show depth as time unfolding across the blocks of a single one-step evaluation that solves the full noise-to-clean task. However, one-step models based on diffusion are trained on a more general transport task that also supports multi-step generation. In this section, we show that depth as time extends to arbitrary start and end points along the generation trajectory, in two settings: distilled models in Sec.~\ref{subsec:distillation} and MeanFlow in Sec.~\ref{subsec:flow_map} (and relegate Shortcut models to Appendix~\ref{app:shortcut}).

In distilled models, we find that each sampling step contains its own depthwise refinement, nesting depth as time within sampling time. In MeanFlow, the depthwise trajectory adapts to the requested transport interval with an interesting property: when the prescribed endpoint is noisy, intermediate layers first denoise to reveal cleaner structure and then gradually \textit{renoise} toward that endpoint.

% \wy{TODO for Arnold, write 1-2 sentence here that is the takeaway of the distilled model and the mean-flow section}

% For the remainder of this section, we decode intermediate residual states using Equation~\eqref{eq:layer_velocity_readout} and recover normalized depth schedules using Equation~\eqref{eq:depth_schedule}. We study two settings: distilled one-step models $f_\theta(z,\tau_\mathrm{task}=t)$ in Sec.~\ref{subsec:distillation} and flow-map models trained without a teacher in Sec.~\ref{subsec:flow_map}. 
% ~\footnote{We provide the detailed background on one-step generative models in Appendix~\ref{app:distilled}}.

% $\boldsymbol{s_\theta}(\z,\textcolor[HTML]{B02418}{\tau=(t,d)})$ in \S~\ref{subsec:flow_map}, and $\tau$-independent endpoint-only distribution-matching models of the form $\x=\boldsymbol{f_\theta}(\boldsymbol{\epsilon})$ in \S~\ref{subsec:drifting}. We also show that the depthwise trajectory reflects the training objective $\boldsymbol{u_\theta}(\z,\textcolor[HTML]{B02418}{\tau=(t,r)})$ in \S~\ref{subsec:objective}.

\subsection{Distilled one-step generators nest depth as time in sampling time}
\label{subsec:distillation}

We first consider distilled one-step models, which are conditioned only on the current state and time $f_\theta(\x_t,t)$. Because they do not explicitly receive an endpoint $r$, multi-step generation relies on an external sampler that uses each network function evaluation (NFE) to advance from the current time $t$ to a prescribed next time $r$. We refer to the resulting transport problem $t\rightarrow r$ as a \emph{subtask} of the full $1\rightarrow0$ generation task.

To probe how computation unfolds within these subtasks, we study two large-scale distilled text-to-image models, FLUX-schnell and SANA-Sprint~0.6B~\citep{chen2025sana}. For each subtask, we decode both the endpoint prediction $\textcolor[HTML]{B02418}{\tilde{\x}_r^i}$ and the diagnostic clean prediction $\textcolor[HTML]{B02418}{\tilde{\x}_0^i}$ following Equation~\eqref{eq:xr}. The former shows how the layerwise prediction evolves toward the next state prescribed by the sampler; the latter shows what clean image is implied by the same intermediate representation. 

\begin{figure}[t]
  \centering
  % \vspace{-0.5em}
  \includegraphics[width=0.9\linewidth]{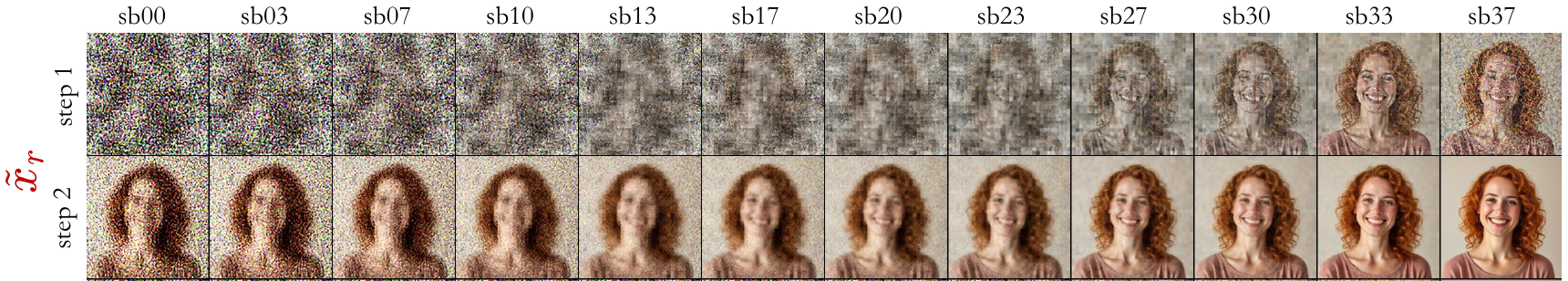}
  \vspace{0em}

    \caption{\textbf{Each sampler step exhibits its own depthwise denoising progression.} Layerwise endpoint predictions $\textcolor[HTML]{B02418}{\tilde{\x}_r^i}$ for two-step FLUX-schnell, decoded after each single block (sb). Within each NFE, the prediction is refined across depth toward the endpoint prescribed by the sampler: a still-noisy intermediate state in step~1, and the clean image in step~2.}

  \vspace{-0.5em}
  \label{fig:q7_two_step}
\end{figure}

\begin{figure*}[t]
  \centering

  \includegraphics[width=\linewidth]{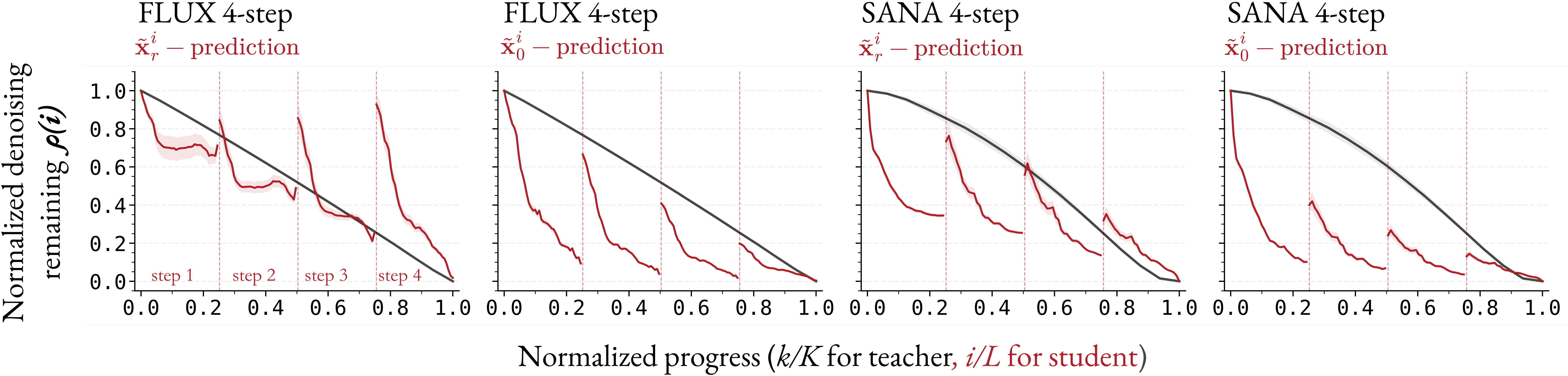}
  \caption{\textbf{Sampling time and network depth encode distinct but nested progressions.} Normalized distance $\rho(i)$ (Equation~\eqref{eq:depth_schedule}) for four-step FLUX-schnell and SANA-Sprint, computed from the endpoint predictions $\textcolor[HTML]{B02418}{\tilde{\x}_r^i}$ and the diagnostic clean predictions $\textcolor[HTML]{B02418}{\tilde{\x}_0^i}$. Red curves show the layerwise progression within each NFE, gray curves show the multi-step teacher's sampling trajectory, and dashed lines mark sampler steps. The $\textcolor[HTML]{B02418}{\tilde{\x}_r^i}$ curves reset at each step as the sampler advances, whereas the $\textcolor[HTML]{B02418}{\tilde{\x}_0^i}$ curves largely reach the final clean image within the first step.}
  \label{fig:distillation}
  \vspace{-0.5em}
\end{figure*}

Figure~\ref{fig:q7_two_step} shows the layerwise endpoint predictions for FLUX-schnell with generation split into two subtasks, and Figure~\ref{fig:distillation} quantifies both predictions using Equation~\eqref{eq:depth_schedule} for FLUX-schnell and SANA-Sprint across four subtasks.
%We analyze each model following \hyperref[alg:depth-probe]{Alg.~1}.

Each subtask exhibits its own depthwise refinement. In Figure~\ref{fig:q7_two_step}, the prediction in step~1 sharpens toward a still-noisy intermediate state; step~2 then restarts from that state and refines toward the clean image. In Figure~\ref{fig:distillation}, this appears as the $\textcolor[HTML]{B02418}{\tilde{\x}_r^i}$ curves jumping back up each time the sampler advances. This reset, however, applies only to the sampler state. The diagnostic $\textcolor[HTML]{B02418}{\tilde{\x}_0^i}$ predictions show that even during the first subtask, $1\rightarrow0.75$, the final layerwise representation already largely implies the clean image at $t=0$ (Figure~\ref{fig:distillation}, second and fourth panels). The amount of refinement in $\textcolor[HTML]{B02418}{\tilde{\x}_0^i}$ also shrinks in later subtasks as the sampler itself approaches the clean endpoint.

The endpoint predictions $\textcolor[HTML]{B02418}{\tilde{\x}_r^i}$, by contrast, do not progress as far toward the clean image within each step and can flatten or even reverse within depth, most visibly for FLUX (Figure~\ref{fig:distillation}, first panel). This is consistent with the model returning toward the prescribed noisy endpoint. In general, this suggests that sampling time and network depth are distinct but nested axes of progression. We examine the reversal in $\textcolor[HTML]{B02418}{\tilde{\x}_r^i}$ more directly in explicitly interval-conditioned MeanFlow models in Sec.~\ref{subsec:flow_map}.

\subsection{MeanFlows Denoise, Then Renoise, Toward Noisy Endpoints}
\label{subsec:flow_map}
\begin{figure*}[t]
  \centering

  \begin{subfigure}{0.46\textwidth}
    \centering
    \includegraphics[width=\linewidth]{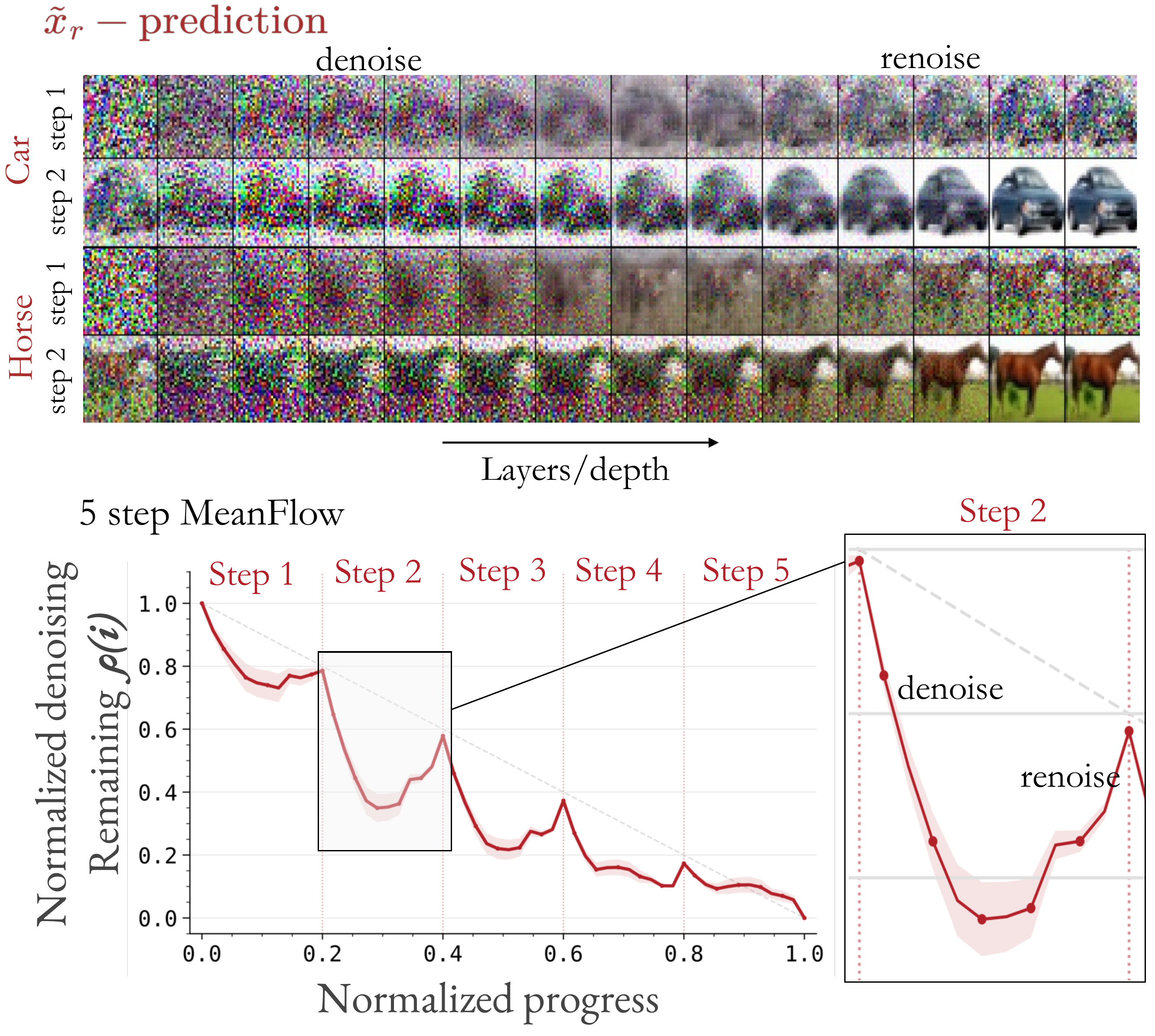}
    \caption{MeanFlow on CIFAR-10 (\texttt{DiT-B/2})}
    \label{fig:meanflow_cifar}
  \end{subfigure}
  \hfill
  \begin{subfigure}{0.53\textwidth}
    \centering
    \includegraphics[width=\linewidth]{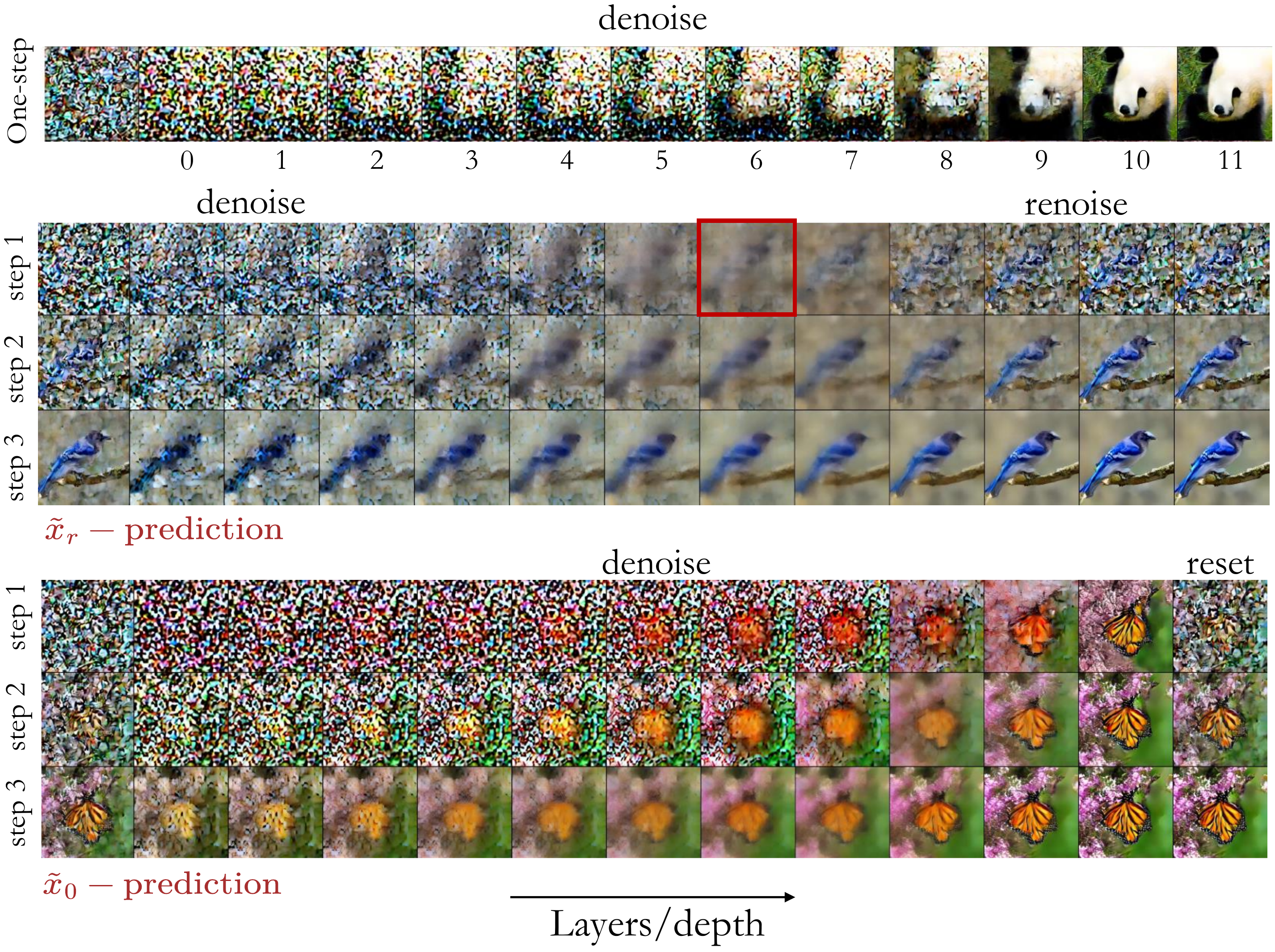}
    \caption{MeanFlow on ImageNet (\texttt{SiT-B/2})}
    \label{fig:meanflow_imagenet}
  \end{subfigure}

  \caption[Depth as time in MeanFlow.]{\textbf{
  Multi-step MeanFlow models trained on CIFAR-10 and ImageNet exhibit the \emph{denoise-then-renoise} pattern.}
  \textbf{(a)} Visualizing $\textcolor[HTML]{B02418}{\tilde{\x}_r}$-predictions on the two-step generation trajectory  across depth for a MeanFlow trained on CIFAR-10\footnotemark reveals the denoise-then-renoise pattern (\emph{top}). Quantitatively, the recovered depth schedule $\boldsymbol{\rho}(i)$ for a five-step generation trajectory , which decreases and then increases within each step (\emph{bottom}).
  \textbf{(b)} We observe the depth as time on a MeanFlow with \texttt{SiT-B/2} trained on ImageNet with one step generation (\emph{top}). Using the same model but on three-step generation, we observe the same denoise and renoise pattern with $\textcolor[HTML]{B02418}{\tilde{\x}_r}$-predictions (\emph{middle}) and $\textcolor[HTML]{B02418}{\tilde{\x}_0}$-predictions (\emph{bottom}).
}
  \label{fig:meanflow_depth_transport}
  \vspace{-0.5em}
\end{figure*}
\footnotetext{Most CIFAR-10 diffusion and one-step generation codebases use DDPM++/NCSN++-style U-Nets~\citep{song2020score}; we train our own \texttt{DiT-B/2} on CIFAR-10 for this experiment.}

Unlike the distilled models of Sec.~\ref{subsec:distillation}, which
receive the endpoint $r$ only implicitly through the external sampler,
MeanFlow \citep{geng2025mean} is explicitly conditioned on the interval
$(t, r)$ and predicts the average velocity $u$ over it. This lets us ask
directly how the requested endpoint shapes the depthwise trajectory. The probe of Sec.~\ref{subsec:distillation} carries over directly, except
that the final output head now estimates the interval-averaged velocity
$\textcolor[HTML]{B02418}{\boldsymbol{u}_i}$ not an instantaneous one.
Decoding it gives the layerwise endpoint prediction
$
    \textcolor[HTML]{B02418}{\tilde{\x}_r^{\,i}}
    =\z_t-(t-r)\textcolor[HTML]{B02418}{\boldsymbol{u}_i}
$
for $r>0$, and a diagnostic clean-image estimate from the same velocity,
$
    \textcolor[HTML]{B02418}{\tilde{\x}_0^{\,i}}
    = \z_t-t\textcolor[HTML]{B02418}{\boldsymbol{u}_i}
$ for $r = 0$.

For one-step generation, the requested endpoint is clean ($r = 0$), and
MeanFlow behaves as in Sec.~\ref{subsec:first-evidence-denoising}: on ImageNet \citep{russakovsky2015imagenet}, the layerwise predictions progress approximately monotonically from noise toward the clean image (Figure~\ref{fig:meanflow_depth_transport}b, top row). Under multi-step sampling, each network evaluation instead solves a subtask
$t \to r$ with $r > 0$. The diagnostic clean predictions
$\textcolor[HTML]{B02418}{\tilde{\x}_0^{\,i}}$ show that depth as time
persists within each subtask: at every step, the layerwise predictions again
progress toward the clean image, nesting depthwise refinement within sampling
time as in the distilled models of Sec.~\ref{subsec:distillation}
(Figure~\ref{fig:meanflow_depth_transport}b, bottom rows).

The endpoint predictions $\textcolor[HTML]{B02418}{\tilde{\x}_r^{\,i}}$,
which track what the network \emph{actually} outputs for each subtask, reveal a
different pattern. Because the prescribed endpoint is now noisy, the
depthwise trajectory becomes non-monotonic: intermediate layers first reveal
a cleaner image structure than the desired target, and later layers
gradually \textit{renoise} the prediction toward $\z_r$. We observe this on both
CIFAR-10 \citep{krizhevsky2009learning} and ImageNet
(Figure~\ref{fig:meanflow_depth_transport}a, top;
Figure~\ref{fig:meanflow_depth_transport}b, middle rows), where the cleanest
prediction appears in the intermediate layers (highlighted in \textcolor[HTML]{B02418}{red}) rather than the noisy endpoint. 

The normalized distance $\rho(i)$ for a 5-step MeanFlow on CIFAR-10 quantifies the same effect: within each step, $\rho(i)$ first falls and then rises again before the next step begins (Figure~\ref{fig:meanflow_depth_transport}a, bottom). We refer to this pattern as \emph{denoise-then-renoise}. It suggests that even when asked for only a partial transport, the network passes through an estimate
of the clean image. %We next offer an explanation for why.

%Full probing details are given in \hyperref[alg:meanflow-depth-probe]{Alg.~2}.

\textbf{Compositional hypothesis for \textit{denoise-then-renoise}.} At first glance, this trajectory is surprising. An interval-conditioned model
is asked to solve the direct transport $\z_t\mapsto\z_r$, yet its intermediate
predictions detour through the clean endpoint estimate before returning to the
requested noisy one. One candidate explanation is that flow maps satisfy a
\emph{compositional consistency} property~\citep{boffi2025flow} (or a one-parameteric semigroup). Writing
$\Phi_{t\rightarrow r}$ for the transport from time $t$ to time $r$, any
intermediate time $s$ gives
$\Phi_{t\rightarrow r}=\Phi_{s\rightarrow r}\circ\Phi_{t\rightarrow s}$.
Composing through the clean boundary, in particular, yields
$\Phi_{t\rightarrow r}(\z_t)=\Phi_{0\rightarrow r}\!\left(\Phi_{t\rightarrow 0}(\z_t)\right)$,
a factorization $\z_t\rightarrow\z_0\rightarrow\z_r$ that matches the
observed trajectory. Yet, compositional consistency alone cannot be the full
explanation since it holds equally for every intermediate time
$s$ and therefore does not single out the clean endpoint.

The clean endpoint estimate is instead singled out by the structure of the training target. For example, under the linear path $\z_t=(1-t)\x_0+t\epsilon$, the marginal velocity is affine in the posterior clean estimate:
$
    v(\z_t,t)
    =\mathbb{E}[\epsilon-\x_0\mid\z_t]
    =\frac{\z_t-\textcolor[HTML]{1E4E8C}{\mathbb{E}[\x_0\mid\z_t]}}{t}
    \label{eq:x0_param}
$ \citep{lipman2022flow, albergo2023stochastic}\footnote{The same holds for any affine path $\z_t=\alpha_t\x_0+\sigma_t\epsilon$, since $\mathbb{E}[\epsilon\mid\z_t]=\big(\z_t-\alpha_t\textcolor[HTML]{1E4E8C}{\mathbb{E}[\x_0\mid\z_t]}\big)/\sigma_t$.}.
A single Euler step to any endpoint $r$ therefore gives
$\z_t-(t-r)\,v(\z_t,t)
=\tfrac{r}{t}\z_t+\left(1-\tfrac{r}{t}\right)\textcolor[HTML]{1E4E8C}{\mathbb{E}[\x_0\mid\z_t]}$,
so every endpoint estimate with $r<t$ is an affine combination of $\z_t$ and
$\textcolor[HTML]{1E4E8C}{\mathbb{E}[\x_0\mid\z_t]}$. The posterior clean
estimate is thus a non-trivial computation shared across all of these tasks.

For MeanFlow, this extends beyond a single Euler step: the average
velocity $u(\z_t,t,r)=\frac{1}{t-r}\int_r^t v(\z_s,s)\,ds$ satisfies the
MeanFlow identity:
$
    u(\z_t,t,r)
    =v(\z_t,t)-(t-r)\,\frac{d}{dt}u(\z_t,t,r)
    \label{eq:meanflow_identity}
$ \citep{geng2025mean}.
The first term, $v(\z_t,t)$, is independent of $r$ and, determined by the clean estimate
$\textcolor[HTML]{1E4E8C}{\mathbb{E}[\x_0\mid\z_t]}$; the second is an
interval-dependent correction that vanishes as $r\rightarrow t$. Therefore, a network trained jointly over many $(t,r)$ pairs has a natural computation to reuse: first resolve the clean sample implied by $\z_t$, then apply the $r$-specific correction that carries it to the requested endpoint. This two-stage computation matches the observed denoise-then-renoise
trajectory in Figure~\ref{fig:meanflow_depth_transport}.

\section{Making Depth as Time Explicit}
\label{sec:explicit_depth_as_time}

So far we demonstrated that task-conditioned one-step models refine their predictions across layers in a denoising-like manner. A natural next question is whether this layerwise computation can itself be explicitly modeled as a \emph{flow}, with a single weight-shared block applied repeatedly and conditioned on a time variable that loosely corresponds to the layer index. If so, the one-step model could be compressed further in its number of parameters, while also gaining the flexibility to vary the number of \emph{depthwise} steps rather than being tied to a fixed number of layers $L$.

Concretely, the one-step model has a sequence of layer-specific blocks $\{\mathbf{B}_i\}_{i=1}^{L}$, with hidden layers $\mathbf{h}_{i+1}
=
\mathbf{B}_{i+1} (
\mathbf{h}_i; \boldsymbol{\tau}_\mathrm{task}, c)
$ (\cref{eq:hidden_states_depth}). We have demonstrated that these blocks loosely correspond to denoising via depth as time. Making this explicit, we can rewrite the model as:
\begin{equation}
\mathbf{h}_{\tau_{i+1}}
=
\textcolor[HTML]{B02418}{\mathbf{B}_\theta}
\left(
\mathbf{h}_{\tau_i},
\textcolor[HTML]{B02418}{(\tau_i,\tau_{i+1})};
\textcolor[HTML]{777777}{\boldsymbol{\tau}_{\mathrm{task}},c}
\right),
\label{eq:explicit_depth_as_time}
\end{equation}
with the time interval conditioning
$\textcolor[HTML]{B02418}{(\tau_i,\tau_{i+1})}$
replacing the layer-specific transformations. Intuitively, we can view this sequence as a flow in the representation space: each application of $\textcolor[HTML]{B02418}{\mathbf{B}_\theta}$ over the computational interval $(\tau_i,\tau_{i+1})$ transports the hidden state $\mathbf{h}_{\tau_i}$ to $\mathbf{h}_{\tau_{i+1}}$. If we set $\textcolor[HTML]{B02418}{\tau_i}=i/L$, we maintain $L$ steps corresponding to the $L$ layers.

We train the parameters \textcolor[HTML]{B02418}{$\mathbf{B}_\theta$} using the supervision signal from the $L$ hidden layers $\mathbf{h}_1, \dots, \mathbf{h}_L$ of the one-step model. We embed the conditioning variables, including the interval \textcolor[HTML]{B02418}{$(\tau_i,\tau_{i+1})$}, and inject them into the shared block through adaptive layer normalization (AdaLN) modulation. This training is data-free and uses only random noise $z\sim\mathcal{N}(\mathbf{0},\mathbf{I})$, sampled class labels, and the one-step teacher model representations. See Appendix~\ref{app:training} for training details.

\textbf{Parameter compression.}
Using this formulation, we compress the MeanFlow \texttt{SiT-B/2} and \texttt{SiT-L/2} one-step models for ImageNet $256{\times}256$ from $131$M and $459$M to $16.0$M and $27.6$M parameters, respectively, corresponding to $8.2\times$ and $16.6\times$ compression. The compressed models reach FID-50k $15.6$ and $11.7$, which improve to $8.2$ and $4.9$ with FD-loss post-training~\citep{yang2026representation}, compared with $6.2$ and $4.0$ for the original models. Using two unique blocks instead of one trades some compression for quality, reaching FID $5.84$ with $26.6$M parameters ($4.9\times$ compression) for \texttt{SiT-B/2} and $4.04$ with $46.5$M parameters ($9.9\times$) for \texttt{SiT-L/2} after post-training. We ablate the effect of conditioning on the interval \textcolor[HTML]{B02418}{$(\tau_i,\tau_{i+1})$} in Appendix~\ref{app:looping_ablation} and show that on CIFAR-10, removing it from an otherwise identical weight-shared block more than doubles FID-50k.

\textbf{Compute compression.}
Because \textcolor[HTML]{B02418}{$\mathbf{B}_\theta$} now defines a flow over depth, we can also borrow intuition from accelerating diffusion models to reduce inference-time compute, training over coarse intervals \textcolor[HTML]{B02418}{$(\tau_a,\tau_b)$} that span multiple layers so that \textcolor[HTML]{B02418}{$\mathbf{B}_\theta$} runs fewer than $L$ times. For \texttt{SiT-L/2} with two unique blocks, halving the number of block applications gives FID $5.25$, and reducing it by $3\times$ gives $6.11$, against a teacher FID of $4.01$ (after post-training). In contrast, simply running the native-depth student with fewer applications degrades FID (Appendix~\ref{app:compression_results}). Much like sampling steps in diffusion, the number of depthwise steps thus becomes a knob that trades compute for quality within a single forward pass.

\textbf{MeanFlow models are more compressible than drifting models.}
Finally, recall that drifting models, which lack a time-indexed transport task or training signal, do not exhibit depth as time (Sec.~\ref{subsec:first-evidence-denoising}), even though their predictions still sharpen in later blocks. Consistent with this, making depth as time explicit fails for them: with an analogous procedure (Appendix~\ref{app:drifting}), the drifting \texttt{B/2} student reaches only FID-50k $47.09$, about three times worse than the MeanFlow \texttt{SiT-B/2} student ($15.61$) at a similar compression ratio, despite distilling from a stronger teacher (FID $1.75$). It therefore seems that refinement along depth is not enough to result in a compressible one-step generator, and that the time-indexed transport task is what organizes this refinement into a denoising process compact enough for a single time-conditioned block.

Full training details and results are in Appendix~\ref{app:acceleration}.

\vspace{-.5em}
\section{Conclusion}
\vspace{-.5em}

We began by asking what happens to the denoising trajectory of diffusion when generation is compressed into a single forward pass. Through a series of experiments, we showed that one-step generative models exhibit a denoising-like progression across network depth, a phenomenon we call \emph{depth as time}. Decoding intermediate representations further revealed that this progression reflects the transport task being solved, from approximately monotonic denoising toward clean endpoints to \emph{denoise-then-renoise} when the prescribed endpoint remains noisy. Finally, we made this depthwise structure explicit and reveal that models with explicit time-indexed transport maps are more compressible.  Leveraging this property with weight sharing and coarse depthwise steps, we can further compress both parameters and compute, resulting in a more compute efficient one-step model.

% \subsection*{AI use statement}

% In this work, we used generative AI tools (Claude Code and Cursor) to implement methods, specifically for writing and debugging experimental code. We have not used generative AI tools to propose or refine hypotheses, develop conceptual frameworks, formulate mathematical claims, design research methodology or experiments, or interpret results, and generating synthetic datasets, assisting with proofs, translation, dataset cleaning, and qualitative data analysis are not applicable to this work. Additionally, we used generative AI tools (Claude Opus 5.5 and GPT-6 Sol) for feedback on the presentation and readability of earlier drafts, for formatting LaTeX tables and figures, and for cross-checking reported numbers against training and evaluation logs. We have reviewed all AI-assisted work: all AI-assisted code was reviewed and tested by the authors, all reported numbers were verified against the original evaluation logs, all citations were manually checked against the original papers, and all text in the final version was written by the authors. We take responsibility for the final content of this work, including text, claims or artifacts produced with the aid of generative AI.

\subsection*{Acknowledgements}

We thank Prof. Tim Buschman, Prof. Ryan P. Adams, Tyler Zhu, Kaleb Newman, Kun Wang, Arash Vahdat, and Julius Berner for their insightful discussions at various points of this project. This work was supported in part by the Princeton First-Year Fellowship awarded to ACA, the Princeton Francis Robbins Upton Fellowship to SC, and Solidigm AI SW. 

% and used the Ionic, Neuronic and Della computing clusters maintained by the Department of Computer Science at Princeton University.

% \bibliographystyle{plainnat}
\bibliographystyle{plainnat}
\bibliography{ref}

@String(NeurIPS  = "{Neural Information Processing Systems (NeurIPS)}")

@String(AAAI  = "{Association for the Advancement of Artificial Intelligence (AAAI)}")

@String(ICML  = "{International Conference on Machine Learning (ICML)}")

@article{geng2025mean,
  title={Mean Flows for One-step Generative Modeling},
  author={Geng, Zhengyang and Deng, Mingyang and Bai, Xingjian and Kolter, J Zico and He, Kaiming},
  journal={arXiv preprint arXiv:2505.13447},
  year={2025}
}

@article{zhou2025inductive,
  title={Inductive Moment Matching},
  author={Zhou, Linqi and Ermon, Stefano and Song, Jiaming},
  journal={arXiv preprint arXiv:2503.07565},
  year={2025}
}

@article{ho2022classifier,
  title={Classifier-free Diffusion Guidance},
  author={Ho, Jonathan and Salimans, Tim},
  journal={arXiv preprint arXiv:2207.12598},
  year={2022}
}

@article{lipman2022flow,
  title={Flow Matching for Generative Modeling},
  author={Lipman, Yaron and Chen, Ricky TQ and Ben-Hamu, Heli and Nickel, Maximilian and Le, Matt},
  journal={arXiv preprint arXiv:2210.02747},
  year={2022}
}

@article{song2023consistency,
  title={Consistency Models},
  author={Song, Yang and Dhariwal, Prafulla and Chen, Mark and Sutskever, Ilya},
  year={2023},
  journal={arXiv preprint arXiv:2303.01469}
}

@article{salimans2022progressive,
  title={Progressive Distillation for Fast Sampling of Diffusion Models},
  author={Salimans, Tim and Ho, Jonathan},
  journal={arXiv preprint arXiv:2202.00512},
  year={2022}
}

@article{boffi2025flow,
  title={Flow Map Matching with Stochastic Interpolants: A Mathematical Framework for Consistency Models},
  author={Boffi, Nicholas Matthew and Albergo, Michael Samuel and Vanden-Eijnden, Eric},
  journal={TMLR},
  year={2025}
}

@inproceedings{
song2020score,
title={Score-Based Generative Modeling through Stochastic Differential Equations},
author={Yang Song and Jascha Sohl-Dickstein and Diederik P Kingma and Abhishek Kumar and Stefano Ermon and Ben Poole},
booktitle={International Conference on Learning Representations},
year={2021},
url={https://openreview.net/forum?id=PxTIG12RRHS}
}

@article{liu2022flow,
  title={Flow Straight and Fast: Learning to Generate and Transfer Data with Rectified Flow},
  author={Liu, Xingchao and Gong, Chengyue and Liu, Qiang},
  journal={arXiv preprint arXiv:2209.03003},
  year={2022}
}

@article{albergo2023stochastic,
  title={Stochastic Interpolants: A Unifying Framework for Flows and Diffusions},
  author={Albergo, Michael S and Boffi, Nicholas M and Vanden-Eijnden, Eric},
  journal={arXiv preprint arXiv:2303.08797},
  year={2023}
}

@article{luo2023diff,
  title={{Diff-Instruct}: A Universal Approach for Transferring Knowledge from Pre-trained Diffusion Models},
  author={Luo, Weijian and Hu, Tianyang and Zhang, Shifeng and Sun, Jiacheng and Li, Zhenguo and Zhang, Zhihua},
  journal={NeurIPS},
  volume={36},
  pages={76525--76546},
  year={2023}
}

@inproceedings{sohl2015deep,
  title={Deep Unsupervised Learning using Nonequilibrium Thermodynamics},
  author={Sohl-Dickstein, Jascha and Weiss, Eric and Maheswaranathan, Niru and Ganguli, Surya},
  booktitle={ICML},
  pages={2256--2265},
  year={2015},
  organization={pmlr}
}

@article{goodfellow2014generative,
  title={Generative Adversarial Nets},
  author={Goodfellow, Ian J and Pouget-Abadie, Jean and Mirza, Mehdi and Xu, Bing and Warde-Farley, David and Ozair, Sherjil and Courville, Aaron and Bengio, Yoshua},
  journal={NeurIPS},
  year={2014}}

@inproceedings{black2024training,
  title={Training diffusion models with reinforcement learning},
  author={Black, Kevin and Janner, Michael and Du, Yilun and Kostrikov, Ilya and Levine, Sergey},
  booktitle={International Conference on Learning Representations},
  volume={2024},
  pages={4965--4987},
  year={2024}
}

@inproceedings{wallace2024diffusion,
  title={Diffusion model alignment using direct preference optimization},
  author={Wallace, Bram and Dang, Meihua and Rafailov, Rafael and Zhou, Linqi and Lou, Aaron and Purushwalkam, Senthil and Ermon, Stefano and Xiong, Caiming and Joty, Shafiq and Naik, Nikhil},
  booktitle={Proceedings of the IEEE/CVF Conference on Computer Vision and Pattern Recognition},
  pages={8228--8238},
  year={2024}
}

@article{fan2023dpok,
  title={Dpok: Reinforcement learning for fine-tuning text-to-image diffusion models},
  author={Fan, Ying and Watkins, Olivia and Du, Yuqing and Liu, Hao and Ryu, Moonkyung and Boutilier, Craig and Abbeel, Pieter and Ghavamzadeh, Mohammad and Lee, Kangwook and Lee, Kimin},
  journal={Advances in Neural Information Processing Systems},
  volume={36},
  pages={79858--79885},
  year={2023}
}

@article{meng2021sdedit,
  title={Sdedit: Guided image synthesis and editing with stochastic differential equations},
  author={Meng, Chenlin and He, Yutong and Song, Yang and Song, Jiaming and Wu, Jiajun and Zhu, Jun-Yan and Ermon, Stefano},
  journal={arXiv preprint arXiv:2108.01073},
  year={2021}
}

@inproceedings{li2023your,
  title={Your diffusion model is secretly a zero-shot classifier},
  author={Li, Alexander C and Prabhudesai, Mihir and Duggal, Shivam and Brown, Ellis and Pathak, Deepak},
  booktitle={Proceedings of the IEEE/CVF International Conference on Computer Vision},
  pages={2206--2217},
  year={2023}
}

@article{clark2023text,
  title={Text-to-image diffusion models are zero shot classifiers},
  author={Clark, Kevin and Jaini, Priyank},
  journal={Advances in Neural Information Processing Systems},
  volume={36},
  pages={58921--58937},
  year={2023}
}

@inproceedings{guo2026thinking,
  title={Thinking-while-generating: Interleaving textual reasoning throughout visual generation},
  author={Guo, Ziyu and Zhang, Renrui and Li, Hongyu and Zhang, Manyuan and Chen, Xinyan and Wang, Sifan and Feng, Yan and Pei, Peng and Heng, Pheng-Ann},
  booktitle={Proceedings of the IEEE/CVF Conference on Computer Vision and Pattern Recognition},
  pages={26295--26305},
  year={2026}
}

@inproceedings{zarei2026slideredit,
  title={Slideredit: Continuous image editing with fine-grained instruction control},
  author={Zarei, Arman and Basu, Samyadeep and Pournemat, Mobina and Nag, Sayan and Rossi, Ryan A and Feizi, Soheil},
  booktitle={Proceedings of the IEEE/CVF Conference on Computer Vision and Pattern Recognition},
  pages={14430--14439},
  year={2026}
}

@inproceedings{wang2026cfg,
  title={CFG-Ctrl: Control-Based Classifier-Free Diffusion Guidance},
  author={Wang, Hanyang and Liu, Yiyang and Chi, Jiawei and Liu, Fangfu and Xue, Ran and Duan, Yueqi},
  booktitle={Proceedings of the IEEE/CVF Conference on Computer Vision and Pattern Recognition},
  pages={11437--11447},
  year={2026}
}

@inproceedings{lee2026low,
  title={Low-Resolution Editing is All You Need for High-Resolution Editing},
  author={Lee, Junsung and Lee, Hyunsoo and Lee, Yong Jae and Han, Bohyung},
  booktitle={Proceedings of the IEEE/CVF Conference on Computer Vision and Pattern Recognition},
  pages={16216--16225},
  year={2026}
}

@article{chen2018neural,
  title={Neural ordinary differential equations},
  author={Chen, Ricky TQ and Rubanova, Yulia and Bettencourt, Jesse and Duvenaud, David K},
  journal={Advances in neural information processing systems},
  volume={31},
  year={2018}
}

@article{deng2026generative,
  title={Generative Modeling via Drifting},
  author={Deng, Mingyang and Li, He and Li, Tianhong and Du, Yilun and He, Kaiming},
  journal={arXiv preprint arXiv:2602.04770},
  year={2026}
}

@inproceedings{chen2025sana,
  title={Sana-sprint: One-step diffusion with continuous-time consistency distillation},
  author={Chen, Junsong and Xue, Shuchen and Zhao, Yuyang and Yu, Jincheng and Paul, Sayak and Chen, Junyu and Cai, Han and Han, Song and Xie, Enze},
  booktitle={Proceedings of the IEEE/CVF International Conference on Computer Vision},
  pages={16185--16195},
  year={2025}
}

@article{labs2025flux,
  title={FLUX. 1 Kontext: Flow Matching for In-Context Image Generation and Editing in Latent Space},
  author={{Black Forest Labs} and Batifol, Stephen and Blattmann, Andreas and Boesel, Frederic and Consul, Saksham and Diagne, Cyril and Dockhorn, Tim and English, Jack and English, Zion and Esser, Patrick and others},
  journal={arXiv preprint arXiv:2506.15742},
  year={2025}
}

@inproceedings{zhou2024score,
  title={Score identity distillation: Exponentially fast distillation of pretrained diffusion models for one-step generation},
  author={Zhou, Mingyuan and Zheng, Huangjie and Wang, Zhendong and Yin, Mingzhang and Huang, Hai},
  booktitle={Forty-first International Conference on Machine Learning},
  year={2024}
}

@inproceedings{sauer2024fast,
  title={Fast high-resolution image synthesis with latent adversarial diffusion distillation},
  author={Sauer, Axel and Boesel, Frederic and Dockhorn, Tim and Blattmann, Andreas and Esser, Patrick and Rombach, Robin},
  booktitle={SIGGRAPH Asia 2024 Conference Papers},
  pages={1--11},
  year={2024}
}

@inproceedings{sauer2024adversarial,
  title={Adversarial diffusion distillation},
  author={Sauer, Axel and Lorenz, Dominik and Blattmann, Andreas and Rombach, Robin},
  booktitle={European Conference on Computer Vision},
  pages={87--103},
  year={2024},
  organization={Springer}
}

@inproceedings{yin2024one,
  title={One-step diffusion with distribution matching distillation},
  author={Yin, Tianwei and Gharbi, Micha{\"e}l and Zhang, Richard and Shechtman, Eli and Durand, Fredo and Freeman, William T and Park, Taesung},
  booktitle={Proceedings of the IEEE/CVF conference on computer vision and pattern recognition},
  pages={6613--6623},
  year={2024}
}

@inproceedings{kim2024consistency,
  title={Consistency trajectory models: Learning probability flow ode trajectory of diffusion},
  author={Kim, Dongjun and Lai, Chieh-Hsin and Liao, WeiHsiang and Murata, Naoki and Takida, Yuhta and Uesaka, Toshimitsu and He, Yutong and Mitsufuji, Yuki and Ermon, Stefano},
  booktitle={International Conference on Learning Representations},
  volume={2024},
  pages={44493--44525},
  year={2024}
}

@inproceedings{
frans2025one,
title={One Step Diffusion via Shortcut Models},
author={Kevin Frans and Danijar Hafner and Sergey Levine and Pieter Abbeel},
booktitle={The Thirteenth International Conference on Learning Representations},
year={2025},
url={https://openreview.net/forum?id=OlzB6LnXcS}
}

@inproceedings{
shing2026diffusionblocks,
title={DiffusionBlocks: Block-wise Neural Network Training via Diffusion Interpretation},
author={Makoto Shing and Masanori Koyama and Takuya Akiba},
booktitle={The Fourteenth International Conference on Learning Representations},
year={2026},
url={https://openreview.net/forum?id=pwVSmK71cS}
}

@article{lai2025principles,
  title={The principles of diffusion models},
  author={Lai, Chieh-Hsin and Song, Yang and Kim, Dongjun and Mitsufuji, Yuki and Ermon, Stefano},
  journal={arXiv preprint arXiv:2510.21890},
  year={2025}
}

@article{field1987relations,
  title={Relations between the statistics of natural images and the response properties of cortical cells},
  author={Field, David J},
  journal={Journal of the Optical Society of America A},
  volume={4},
  number={12},
  pages={2379--2394},
  year={1987},
  publisher={Optical Society of America}
}

@article{simoncelli2001natural,
  title={Natural image statistics and neural representation},
  author={Simoncelli, Eero P and Olshausen, Bruno A},
  journal={Annual review of neuroscience},
  volume={24},
  number={1},
  pages={1193--1216},
  year={2001},
  publisher={Annual Reviews 4139 El Camino Way, PO Box 10139, Palo Alto, CA 94303-0139, USA}
}

@article{liang2026personalized,
  title={Personalized Generative Models for Contextual Debiasing},
  author={Liang, Xinran and Tureci, Esin and Sinha, Prachi and Zhu, Ye and Ramaswamy, Vikram V and Russakovsky, Olga},
  journal={arXiv preprint arXiv:2605.26353},
  year={2026}
}

@techreport{krizhevsky2009learning,
  title={Learning multiple layers of features from tiny images},
  author={Krizhevsky, Alex},
  year={2009},
  institution={University of Toronto}
}

@misc{nostalgebraist2020logitlens,
  author       = {nostalgebraist},
  title        = {Interpreting GPT: The Logit Lens},
  year         = {2020},
  month        = aug,
  howpublished = {LessWrong},
  note         = {Posted 31 August 2020}
}

@article{ect,
  title={Consistency models made easy},
  author={Geng, Zhengyang and Pokle, Ashwini and Luo, William and Lin, Justin and Kolter, J Zico},
  journal={arXiv preprint arXiv:2406.14548},
  year={2024}
}

@inproceedings{attention_vs_md,
    title={Is Attention Better Than Matrix Decomposition?},
    author={Zhengyang Geng and Meng-Hao Guo and Hongxu Chen and Xia Li and Ke Wei and Zhouchen Lin},
    booktitle={International Conference on Learning Representations},
    year={2021},
}

@inproceedings{train_implicit,
      title={On Training Implicit Models}, 
      author={Zhengyang Geng and Xin-Yu Zhang and Shaojie Bai and Yisen Wang and Zhouchen Lin},
      booktitle={Advances in Neural Information Processing Systems},
      volume={34},
      pages={24247--24260},
      year={2021}
}

@inproceedings{implicit_graph_neural_networks,
      title={Implicit Graph Neural Networks},
      author={Fangda Gu and Heng Chang and Wenwu Zhu and Somayeh Sojoudi and Laurent El Ghaoui},
      booktitle={Advances in Neural Information Processing Systems},
      volume={33},
      pages={11984--11995},
      year={2020}
}

@inproceedings{jfb,
      title={JFB: Jacobian-Free Backpropagation for Implicit Networks}, 
      author={Samy Wu Fung and Howard Heaton and Qiuwei Li and Daniel McKenzie and Stanley Osher and Wotao Yin},
      booktitle={Proceedings of the AAAI Conference on Artificial Intelligence},
      volume={36},
      pages={6648--6656},
      year={2022},
      doi={10.1609/aaai.v36i6.20619},
}

@inproceedings{deq,
  title={Deep Equilibrium Models},
  author={Bai, Shaojie and Kolter, J Zico and Koltun, Vladlen},
  booktitle={Advances in Neural Information Processing Systems},
  volume={32},
  year={2019}
}

@inproceedings{ut,
  title={Universal transformers},
  author={Dehghani, Mostafa and Gouws, Stephan and Vinyals, Oriol and Uszkoreit, Jakob and Kaiser, {\L}ukasz},
  booktitle={International Conference on Learning Representations},
  year={2019}
}

@inproceedings{pie,
      title={Path Independent Equilibrium Models Can Better Exploit Test-Time Computation}, 
      author={Cem Anil and Ashwini Pokle and Kaiqu Liang and Johannes Treutlein and Yuhuai Wu and Shaojie Bai and Zico Kolter and Roger Grosse},
      booktitle={Advances in Neural Information Processing Systems},
      volume={35},
      year={2022},
}

@inproceedings{looped_transformers_programmable_computers,
      title={Looped Transformers as Programmable Computers},
      author={Giannou, Angeliki and Rajput, Shashank and Sohn, Jy-yong and Lee, Kangwook and Lee, Jason D. and Papailiopoulos, Dimitris},
      booktitle={Proceedings of the 40th International Conference on Machine Learning},
      series={Proceedings of Machine Learning Research},
      volume={202},
      pages={11398--11442},
      publisher={PMLR},
      year={2023},
}

@inproceedings{looped_transformers_expressive_power,
      title={On Expressive Power of Looped Transformers: Theoretical Analysis and Enhancement via Timestep Encoding},
      author={Kevin Xu and Issei Sato},
      booktitle={Proceedings of the 42nd International Conference on Machine Learning},
      year={2025},
}

@inproceedings{latent_thoughts_looped_transformers,
      title={Reasoning with Latent Thoughts: On the Power of Looped Transformers},
      author={Nikunj Saunshi and Nishanth Dikkala and Zhiyuan Li and Sanjiv Kumar and Sashank J. Reddi},
      booktitle={International Conference on Learning Representations},
      year={2025},
}

@misc{latent_cot_depth_recurrent,
      title={Latent Chain-of-Thought? Decoding the Depth-Recurrent Transformer},
      author={Wenquan Lu and Yuechuan Yang and Kyle Lee and Yanshu Li and Enqi Liu},
      year={2025},
      eprint={2507.02199},
      archivePrefix={arXiv},
      primaryClass={cs.CL},
}

@article{two_scale_latent_dynamics,
  title={Two-scale latent dynamics for recurrent-depth transformers},
  author={Pappone, Francesco and Crisostomi, Donato and Rodol{\`a}, Emanuele},
  journal={arXiv preprint arXiv:2509.23314},
  year={2025}
}

@misc{inverse_depth_scaling,
      title={Inverse Depth Scaling From Most Layers Being Similar},
      author={Yizhou Liu and Sara Kangaslahti and Ziming Liu and Jeff Gore},
      year={2026},
      eprint={2602.05970},
      archivePrefix={arXiv},
      primaryClass={cs.LG},
}

@misc{loop_think_generalize,
      title={Loop, Think, \& Generalize: Implicit Reasoning in Recurrent-Depth Transformers},
      author={Harsh Kohli and Srinivasan Parthasarathy and Huan Sun and Yuekun Yao},
      year={2026},
      eprint={2604.07822},
      archivePrefix={arXiv},
      primaryClass={cs.CL},
}

@misc{stability_generalization_looped_transformers,
      title={Stability and Generalization in Looped Transformers},
      author={Asher Labovich},
      year={2026},
      eprint={2604.15259},
      archivePrefix={arXiv},
      primaryClass={cs.LG},
}

@misc{iso_depth_scaling_looped_lms,
      title={How Much Is One Recurrence Worth? Iso-Depth Scaling Laws for Looped Language Models},
      author={Kristian Schwethelm and Daniel Rueckert and Georgios Kaissis},
      year={2026},
      eprint={2604.21106},
      archivePrefix={arXiv},
      primaryClass={cs.LG},
}

@inproceedings{looped_transformers_learning_algorithms,
      title={Looped Transformers are Better at Learning Learning Algorithms},
      author={Liu Yang and Kangwook Lee and Robert Nowak and Dimitris Papailiopoulos},
      booktitle={International Conference on Learning Representations},
      year={2024},
}

@article{yang2026representation,
  title={Representation Fr$\backslash$'echet Loss for Visual Generation},
  author={Yang, Jiawei and Geng, Zhengyang and Ju, Xuan and Tian, Yonglong and Wang, Yue},
  journal={arXiv preprint arXiv:2604.28190},
  year={2026}
}

@article{russakovsky2015imagenet,
  title={Imagenet large scale visual recognition challenge},
  author={Russakovsky, Olga and Deng, Jia and Su, Hao and Krause, Jonathan and Satheesh, Sanjeev and Ma, Sean and Huang, Zhiheng and Karpathy, Andrej and Khosla, Aditya and Bernstein, Michael and others},
  journal={International journal of computer vision},
  volume={115},
  number={3},
  pages={211--252},
  year={2015},
  publisher={Springer}
}

\appendix
\newpage
\appendix

{\Large{\textbf{Appendix}}}

\vspace{1.2em}

\begingroup
\small
\setlength{\parindent}{0pt}
\setlength{\parskip}{0.22em}

\definecolor{tocsub}{HTML}{3A3A3A}  % dark grey for subsections; adjust to taste

\newcommand{\appendixtocsection}[3]{%
  \vspace{0.45em}%
  \noindent\makebox[2.0em][l]{\textbf{\hyperref[#3]{#1}}}%
  \textbf{\hyperref[#3]{#2}}%
  \leaders\hbox to 0.65em{\hss.\hss}\hfill
  \textbf{\pageref{#3}}\par
}

\newcommand{\appendixtocsubsection}[3]{%
  \noindent\hspace*{2.0em}\makebox[3.0em][l]{\hyperref[#3]{\textcolor{tocsub}{#1}}}%
  \hyperref[#3]{\textcolor{tocsub}{#2}}%
  {\color{tocsub}\leaders\hbox to 0.65em{\hss.\hss}\hfill
  \pageref{#3}}\par
}

\appendixtocsection{A}{Related Work and Discussion}{app:extended_related_works}

\appendixtocsection{B}{Readout Controls and Depthwise Mixing Details}{app:readout_and_mixing}
\appendixtocsubsection{B.1}{Readout controls}{app:readout_controls}
\appendixtocsubsection{B.2}{Depthwise mixing details}{app:depthwise_mixing}

\appendixtocsection{C}{Implementation Details for Making Depth as Time Explicit}{app:acceleration}
\appendixtocsubsection{C.1}{Distilling the depthwise trajectory}{app:training}
\appendixtocsubsection{C.2}{FD-loss post-training}{app:fd}
\appendixtocsubsection{C.3}{Results}{app:compression_results}
\appendixtocsubsection{C.4}{Ablating the internal time conditioning}{app:looping_ablation}
\appendixtocsubsection{C.5}{Drifting models as a negative control}{app:drifting}

\appendixtocsection{D}{Extra Background on One-Step Generators}{app:extra_background}
\appendixtocsubsection{D.1}{Distillation}{app:bg_distillation}
\appendixtocsubsection{D.2}{Boundary-anchored and interval-conditioned flow maps}{app:bg_flowmaps}
\appendixtocsubsection{D.3}{Step-conditioned flow maps: Shortcut models}{app:bg_shortcut}

\appendixtocsection{E}{Extra Analyses on Flow Maps}{app:flow_maps}
\appendixtocsubsection{E.1}{Shortcut Models}{app:shortcut}
% \appendixtocsubsection{F.1}{<title>}{<label>}   % fill in once F's subsections exist
% \appendixtocsubsection{F.2}{<title>}{<label>}
\endgroup
\vspace{1.2em}

\newpage
\section{Related Work and Discussion}
\label{app:extended_related_works}

This appendix expands the related work in the main text. Multi-step generators compute across network calls in sampling time, while a one-step generator computes across the layers of a single forward pass. We ask how the former is reorganized into the latter, and whether the resulting depthwise trajectory can itself be modeled and compressed as a flow.

\paragraph{Depth as a dynamical process.}
Neural ODEs~\citep{chen2018neural} view a residual update $\mathbf{h}_{i+1}=\mathbf{h}_i+g(\mathbf{h}_i)$ as an Euler step, with depth playing the role of time. Our layerwise predictions $\boldsymbol{v}_i=\mathbf{W}_{\mathrm{out}}(\mathbf{h}_i;\boldsymbol{\tau}_\mathrm{task},c)$, obtained by applying the model's output layer (normalization, AdaLN modulation, and linear projection) to each intermediate state, make this view measurable in generative models. Universal Transformers~\citep{ut} apply one shared block repeatedly with a per-step timestep embedding, the closest precedent for our explicit model (Section~\ref{sec:explicit_depth_as_time}); our block is instead conditioned on an interval $(\tau_a,\tau_b)$ of an internal clock, which allows steps of different sizes. DiffusionBlocks~\citep{shing2026diffusionblocks} treats residual blocks as denoising steps by design, assigning each block a range of noise levels so that blocks can be trained independently; we find that this correspondence already emerges in one-step generators trained end-to-end. Deep equilibrium models~\citep{deq,pie} and related implicit models~\citep{attention_vs_md,train_implicit,jfb,implicit_graph_neural_networks,stability_generalization_looped_transformers} define the representation as a fixed point $\z^\star=f_\theta(\z^\star;\x)$.

\paragraph{Layerwise predictions and probing.}
Computing layerwise predictions with the model's own output layer is the generative analogue of the \emph{logit lens}~\citep{nostalgebraist2020logitlens}, which decodes the intermediate states of language models with the unembedding and reveals a gradual refinement toward the final prediction. Because what such decodings reveal can depend strongly on the layer and on the choice of decoder~\citep{latent_cot_depth_recurrent}, we confirm our results with a trained and a random linear probe (Appendix~\ref{app:readout_controls}). Representations of diffusion models have mostly been studied across sampling time~\citep{li2023your,clark2023text}; we study them across depth within a single evaluation.

\paragraph{Depth as iterative refinement.}
Depth often implements incremental refinement rather than distinct transformations~\citep{inverse_depth_scaling,two_scale_latent_dynamics}. We find that in one-step generators this refinement is shaped by the transport task: it follows a denoising-like schedule (Section~\ref{subsec:first-evidence-denoising}), adapts to the requested endpoint (Section~\ref{subsec:flow_map}), and, unlike in drifting models, can be compressed into a single time-conditioned block (Section~\ref{sec:explicit_depth_as_time}, Appendix~\ref{app:drifting}).

\paragraph{Weight-tied and looped computation.}
Weight tying decouples computation from parameters. Looped Transformers can emulate programmable computers~\citep{looped_transformers_programmable_computers}, learn iterative algorithms~\citep{looped_transformers_learning_algorithms}, match deeper models~\citep{latent_thoughts_looped_transformers}, and extrapolate to more iterations than seen in training~\citep{loop_think_generalize}, though an extra iteration is worth less than an extra layer~\citep{iso_depth_scaling_looped_lms}. Timestep encodings recover iteration-dependent behavior in looped models~\citep{looped_transformers_expressive_power}, and our internal clock plays this role for generation (Appendix~\ref{app:looping_ablation}).

\paragraph{One-step generative models.}
One-step generators are obtained by distilling a multi-step teacher~\citep{salimans2022progressive,luo2023diff,yin2024one,zhou2024score,sauer2024fast,sauer2024adversarial} or by learning flow maps directly~\citep{song2023consistency,kim2024consistency,frans2025one,boffi2025flow,geng2025mean,zhou2025inductive}, while GANs~\citep{goodfellow2014generative} and drifting models~\citep{deng2026generative} match only the output distribution (Appendix~\ref{app:extra_background}). These methods compress sampling time; we ask where that computation goes, and show that it can be compressed further across depth.
\newpage
\section{Readout Controls and Depthwise mixing details}
\label{app:readout_and_mixing}

\subsection{Readout Controls}
\label{app:readout_controls}

\begin{wrapfigure}{r}{0.4\textwidth}
  \centering
  \vspace{-1.5em}
  \includegraphics[width=\linewidth]{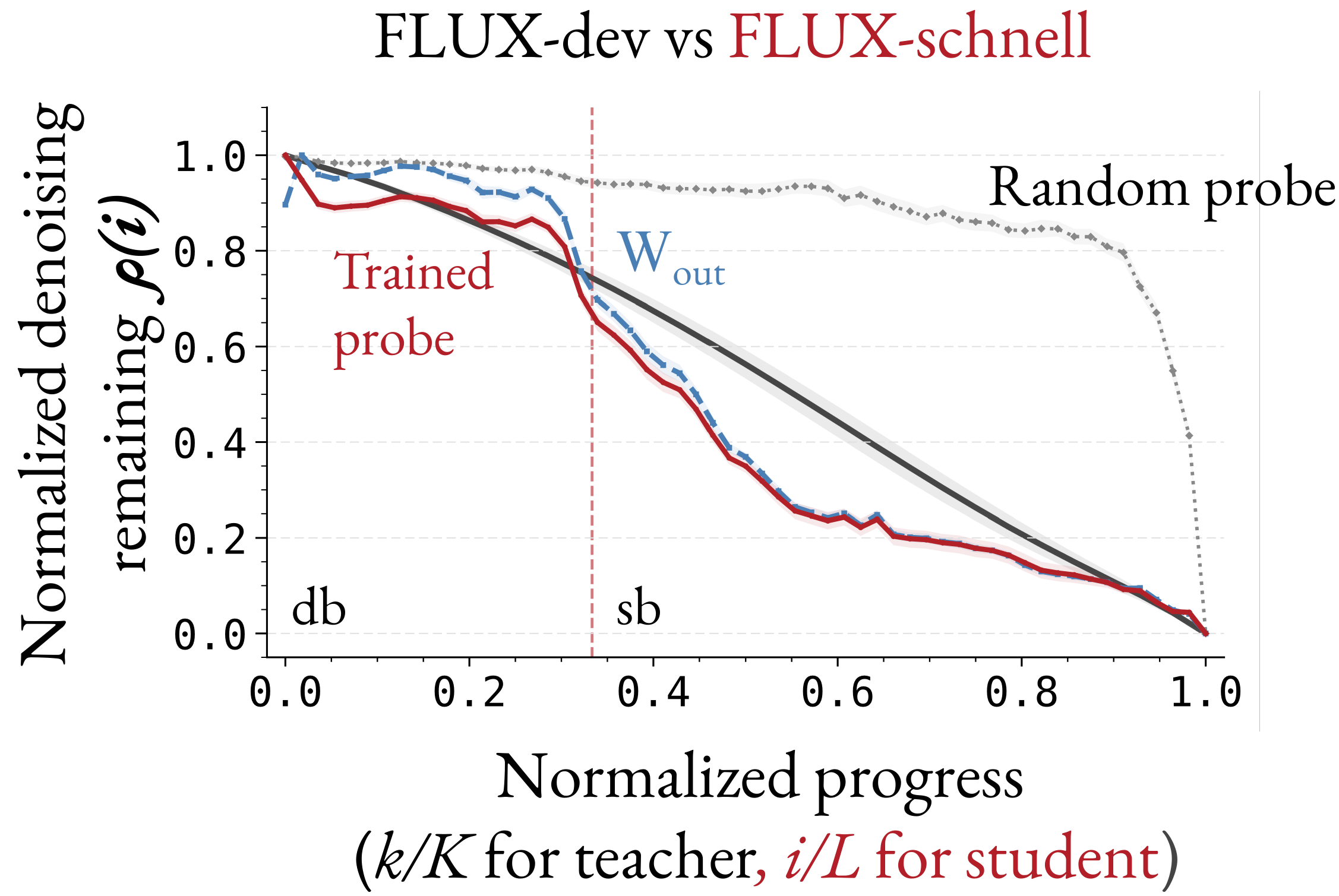}
  \caption{\textbf{Readout controls on one-step FLUX-schnell.} Normalized distance $\rho(i)$ to the final prediction under the model's own readout $\mathbf{W}_{\mathrm{out}}$ (blue), a trained linear probe (red), and a random linear probe (grey), alongside the FLUX-dev teacher's sampling trajectory (black). The dashed line marks the transition from double-stream (db) to single-stream (sb) blocks.}
  \label{fig:readout_controls}
  \vspace{-2em}
\end{wrapfigure}

A natural concern is that the endpoint-convergence pattern measured by Eq.~\eqref{eq:depth_schedule} may be an artifact of applying the model's final readout $\mathbf{W}_{\mathrm{out}}$ to intermediate residual states. Since this readout is optimized only for the final representation $\textcolor[HTML]{B02418}{\mathbf{h}_L}$, intermediate states might gradually become more compatible with it regardless of whether they encode anything meaningful. To assess how the pattern depends on the readout, we compare $\mathbf{W}_{\mathrm{out}}$ with a trained and a random linear probe.

\textbf{Trained probe.}
We fit an independent linear map $\mathbf{V}$ by closed-form ridge regression on the layer-normalized final hidden state, such that
$
\mathbf{V}\bigl(
\mathrm{LayerNorm}(
\textcolor[HTML]{B02418}{\mathbf{h}_L})
\bigr)
\approx
\mathbf{W}_{\mathrm{out}}\bigl(
\textcolor[HTML]{B02418}{\mathbf{h}_L}
\bigr).
$
The target is the model's own final prediction, which may be a velocity $\boldsymbol{v}$, noise $\boldsymbol{\epsilon}$, or image $\boldsymbol{\x}$. We then freeze $\mathbf{V}$ and apply it to every intermediate hidden state $\textcolor[HTML]{B02418}{\mathbf{h}_i}$; the probe is never trained on intermediate layers. On one-step FLUX-schnell, the trained probe traces nearly the same trajectory as $\mathbf{W}_{\mathrm{out}}$ (Fig.~\ref{fig:readout_controls}), indicating that the pattern does not depend on the particular form of the readout.

\textbf{Random probe.}
We also evaluate a frozen random linear map with the same input and output dimensions as $\mathbf{V}$. We sample
$\mathbf{V}_{\mathrm{rand}}
\sim\mathcal{N}(0,1)^{D\times d_{\mathrm{out}}}$
and rescale its weights to match the entrywise standard deviation of $\mathbf{V}$. At each layer, we compute
$
\boldsymbol{v}_i^{\mathrm{rand}}
=
\mathbf{V}_{\mathrm{rand}}
\bigl(
\mathrm{LayerNorm}(
\textcolor[HTML]{B02418}{\mathbf{h}_i})
\bigr),
$
without training the probe to match the model's predictions. The random probe does not reproduce the convergence pattern (Fig.~\ref{fig:readout_controls}): its $\rho(i)$ stays close to $1$ through nearly all blocks and drops only in the last few, where it reaches zero simply because each trajectory is normalized to its own endpoint. The progressive convergence observed with $\mathbf{W}_{\mathrm{out}}$ is therefore semantically meaningful rather than a numerical artifact of applying any linear map.

\subsection{Depthwise Mixing Details}
\label{app:depthwise_mixing}

\textbf{Mixing model.}
For trajectory $n$, let $\boldsymbol{\epsilon}^n$ be the Gaussian input latent and $\tilde{\x}_0^{L,n}$ the model's final endpoint prediction. At each block $i$, we fit one coefficient pair $(\alpha_i,\sigma_i)$, shared across trajectories, by unconstrained least squares over all spatial and latent dimensions:
\[
\tilde{\x}_0^{i,n}
\;\approx\;
\alpha_i\,\tilde{\x}_0^{L,n}
+
\sigma_i\,\boldsymbol{\epsilon}^n.
\]
Coefficients are fit on a training split, and for $i<L$ we report the held-out reconstruction error
\[
R_i
=
\frac{
\sum_n \big\|
\tilde{\x}_0^{i,n}
-\widehat{\alpha}_i\,\tilde{\x}_0^{L,n}
-\widehat{\sigma}_i\,\boldsymbol{\epsilon}^n
\big\|_2^2
}{
\sum_n \big\|
\tilde{\x}_0^{i,n}
-\tilde{\x}_0^{L,n}
\big\|_2^2
},
\]
normalized by the error of predicting the final output directly, so that $R_i=1$ means no improvement over this baseline.

\textbf{Setup.}
We analyze MeanFlow \texttt{B/2} ($L=12$), Drifting \texttt{B/2} ($L=12$), FLUX-schnell (single-stream blocks, $L=38$), SANA-Sprint ($L=28$), and Shortcut \texttt{B/2} ($L=12$). For each, we extract $N=256$ trajectories from a single network evaluation with a fixed seed and split them $70/30$ into $179$ training and $77$ held-out trajectories.

\begin{figure*}[h]
  \centering
  \vspace{-1em}
  \includegraphics[width=\linewidth]{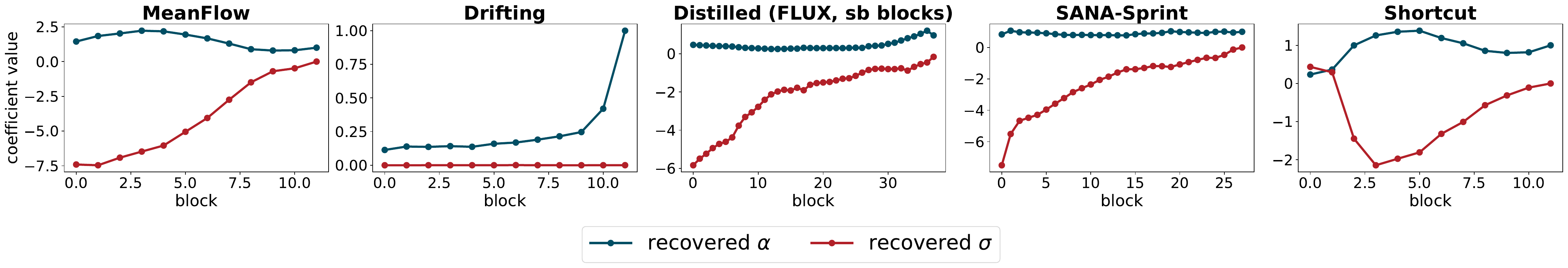}
  \caption{\textbf{Recovered depthwise mixing coefficients} $(\alpha_i,\sigma_i)$. For the task-conditioned models, $\sigma_i$ rises toward zero across depth; for Drifting, $\sigma_i\approx0$ at every block.}
  \label{fig:mixing_coefficients}
\end{figure*}

\textbf{Results.}
For the four task-conditioned models, $\sigma_i$ rises from strongly negative values toward zero across depth while $\alpha_i$ remains of order one (Figure~\ref{fig:mixing_coefficients}), a systematic signal--noise mixing structure. For MeanFlow, this structure is fit with low held-out error (mean $R\approx0.10$), and its coefficients extend well outside the $[0,1]$ range of a flow-matching interpolant, so depth follows its own effective schedule. For Drifting, $\sigma_i\approx0$ at every block, so the input noise plays no identifiable role, and the fit is substantially worse (mean $R\approx0.47$).

\begin{figure*}[h]
  \centering
  \vspace{-1em}
  \includegraphics[width=0.8\linewidth]{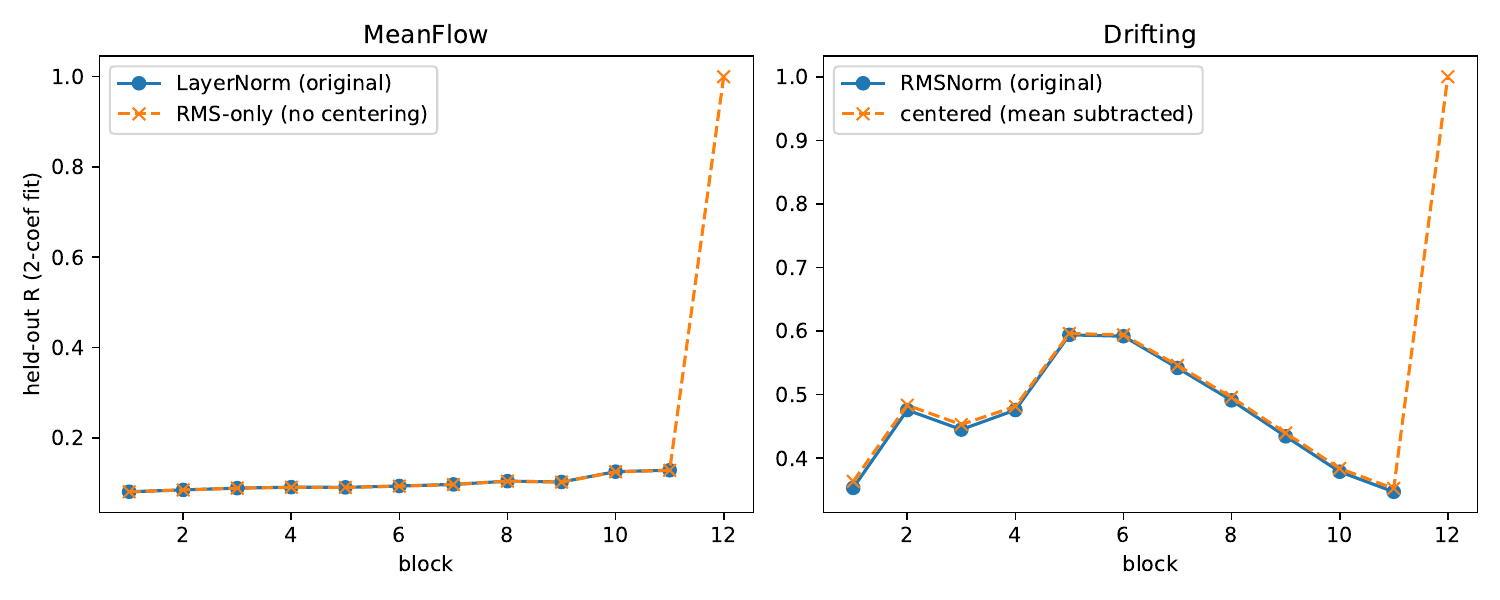}
  \caption{\textbf{Centering ablation.} For MeanFlow, removing centering leaves the held-out reconstruction error essentially
  unchanged across intermediate blocks. For Drifting, adding centering does not improve
  the fit and instead makes it marginally worse. The final block is excluded from the
  normalized $R_i$ comparison because its baseline error is zero by construction.}
  \label{fig:centering_ablation}
\end{figure*}

\textbf{Normalization control.}
MeanFlow's readout uses LayerNorm, which centers its input, whereas Drifting's uses RMSNorm, which does not, so centering could in principle induce the mixing structure. We therefore swap only the normalization, keeping the trained AdaLN modulation and linear projection: we remove centering from MeanFlow and add it to Drifting, then refit the coefficients ($N=192$, same protocol). Removing centering leaves MeanFlow's $R_i$ essentially unchanged, and adding it to Drifting makes the fit marginally worse (Figure~\ref{fig:centering_ablation}), so the contrast is not an artifact of normalization.

\newcommand{\hlbox}[1]{{\setlength{\fboxsep}{1pt}\colorbox[HTML]{F6DEDB}{#1}}}
\newcommand{\hlboxg}[1]{{\setlength{\fboxsep}{1pt}\colorbox[HTML]{DFF0E3}{#1}}}

\section{Implementation Details for Making Depth as Time Explicit}
\label{app:acceleration}

Here, we provide the details behind Section~\ref{sec:explicit_depth_as_time}. We first describe how the weight-shared block \textcolor[HTML]{B02418}{$\mathbf{B}_\theta$} is trained by distilling the depthwise trajectory of a one-step teacher, and how it is trained and run with coarser steps at inference (App.~\ref{app:training}). We then describe FD-loss post-training (App.~\ref{app:fd}) and report full results for parameter and compute compression, including qualitative samples (App.~\ref{app:compression_results}). Finally, we ablate the internal clock on CIFAR-10 (App.~\ref{app:looping_ablation}) and detail the drifting control (App.~\ref{app:drifting}).

% ---------------------------------------------------------------------
\subsection{Distilling the depthwise trajectory}
\label{app:training}
% ---------------------------------------------------------------------

\textbf{Architecture.} The student replaces the $L$ layer-specific blocks of a one-step MeanFlow teacher (\texttt{SiT-B/2}, $L=12$; \texttt{SiT-L/2}, $L=24$) with $n\in\{1,2\}$ unique blocks, applied in alternation for $K$ steps. Each application is conditioned on its interval of the internal clock \textcolor[HTML]{B02418}{$(\tau_a,\tau_b)$}, together with the task conditioning $\boldsymbol{\tau}_\mathrm{task}=(t,r)=(1,0)$ for one-step generation and the class label $c$, all injected through AdaLN modulation. The input embedding $\mathbf{E}$ and output readout $\mathbf{W}_{\mathrm{out}}$ are copied from the teacher and kept frozen, so only the shared blocks and the clock embedding are trained. The shared blocks are warm-started from the teacher's middle block.

\textbf{Objective.} Let $\{\mathbf{H}_j\}_{j=1}^{L}$ denote the teacher's hidden states for an input $\z\sim\mathcal{N}(\mathbf{0},\mathbf{I})$ and label $c$. A student with \textcolor[HTML]{B02418}{$K$} applications uses stride $s=L/\textcolor[HTML]{B02418}{K}$, and the state $\mathbf{h}_k$ reached after its $k$-th application is trained to match the teacher state $\mathbf{H}_{ks}$. Because the norms of the teacher's hidden states grow with depth, an unnormalized loss would be dominated by the last layers and effectively ignore early ones, so we normalize each term by the norm of its teacher state. We additionally anchor the final decoded prediction to the teacher's one-step sample $\x_0^{T}$:
\begin{equation}
\mathcal{L}_{\mathrm{comp}}(\theta)
=
\mathbb{E}_{\z,c}\!\left[
\frac{1}{\textcolor[HTML]{B02418}{K}}
\sum_{k=1}^{\textcolor[HTML]{B02418}{K}}
\frac{\|\mathbf{h}_k-\mathbf{H}_{ks}\|_2^2}{\|\mathbf{H}_{ks}\|_2^2}
+
\lambda\,
\big\|(\z-\boldsymbol{v})-\x_0^{T}\big\|_2^2
\right],
\label{eq:comp_loss}
\end{equation}
where $\boldsymbol{v}=\mathbf{W}_{\mathrm{out}}(\mathbf{h}_{\textcolor[HTML]{B02418}{K}};\boldsymbol{\tau}_\mathrm{task},c)$ is decoded from the final student state and $\lambda=1$. Since the teacher provides its full deterministic trajectory, no self-consistency objective is required, and training is data-free: it uses only noise $\z\sim\mathcal{N}(\mathbf{0},\mathbf{I})$ and sampled class labels, which are replaced by the null label with probability $p_{\mathrm{unc}}=0.1$ to support classifier-free guidance (CFG).

\begin{wrapfigure}{r}{0.45\textwidth}
  \centering
  \vspace{-2em}
  \includegraphics[width=\linewidth]{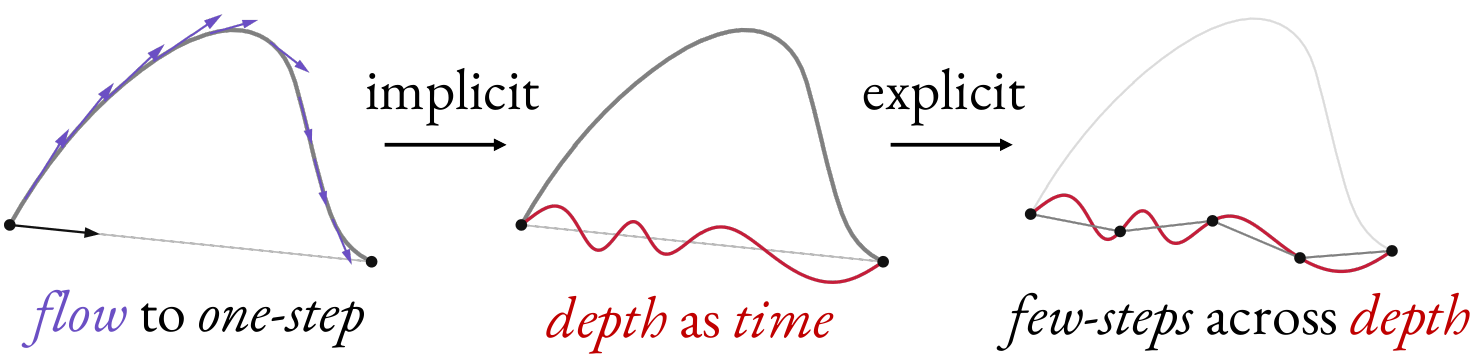}
  \caption{A single application of \textcolor[HTML]{B02418}{$\mathbf{B}_\theta$} over a coarse interval \textcolor[HTML]{B02418}{$(\tau_a,\tau_b)$} replaces several fine-grained transitions.}
  \label{fig:compressing}
  \vspace{-1em}
\end{wrapfigure}

\textbf{Fixed-$K$ and elastic training.} A \emph{fixed-$K$} student is trained for a single number of applications (Alg.\hyperref[alg:fixed-k]{~C.1}). An \emph{elastic} student samples \textcolor[HTML]{B02418}{$K$} uniformly from a set $\mathcal{K}$ at every iteration (Alg.\hyperref[alg:elastic]{~C.2}), so that one model can be run at any depth in the set; we use $\mathcal{K}=\{2,4,6,12\}$ for \texttt{B/2} and $\mathcal{K}=\{4,6,8,12,24\}$ for \texttt{L/2}. The parameter-compression models in Section~\ref{sec:explicit_depth_as_time} are fixed-$K$ students with $K=L$, so that $s=1$ and every teacher state is supervised. When $K<L$, a few applications over coarse intervals replace several fine-grained transitions along depth (Fig.~\ref{fig:compressing}).

\textbf{Hyperparameters.} All fixed-$K$ students, for both \texttt{B/2} and \texttt{L/2} and every number of unique blocks, are trained for $1.2$M iterations with batch size $32$, learning rate $10^{-4}$ with $2{,}000$ warmup steps, and EMA decay $0.9999$. The elastic students use the same settings.

\begin{figure*}[h]
\centering

% ===================== LEFT: Fixed-K =====================
\begin{minipage}[t]{0.49\textwidth}
\vspace{0pt}

\phantomsection
\label{alg:fixed-k}
\noindent{\footnotesize\bfseries Algorithm C.1}
\footnotesize Fixed-$K$ Training
\par\vspace{0.15em}

\begin{tcolorbox}[
  enhanced,
  colback=white,
  colframe=white,
  boxrule=0pt,
  borderline north={0.8pt}{0pt}{black},
  borderline south={0.8pt}{0pt}{black},
  left=4pt,
  right=4pt,
  top=4pt,
  bottom=4pt,
  before skip=0pt,
  after skip=0pt,
]

{\scriptsize\ttfamily
\renewcommand{\arraystretch}{1.15}

\begin{tabularx}{\linewidth}{
  @{}
  r
  @{\hspace{1.3em}}
  >{\raggedright\arraybackslash}X
  @{}
}

    & \textbf{req:} teacher $T$ (depth $D$),
      target $\textcolor[HTML]{B02418}{K}$ \\

\cellcolor{gray!15}
\textcolor[HTML]{62A978}{1}
    &
\cellcolor{gray!15}
      $s \leftarrow D/\textcolor[HTML]{B02418}{K}$
      \hfill
      \makebox[9em][l]{\color{gray}\# fixed depth}\hspace{0.6em} \\

\textcolor{gray}{2}
    & \textbf{repeat} \\

\textcolor{gray}{3}
    & \quad $z \sim \mathcal{N}(0,I),\;\; y \sim p(y)$
      \hfill
      \makebox[9em][l]{\color{gray}\# $y\!\to\!\varnothing$ w.p. $p_{\text{unc}}$}\hspace{0.6em} \\

\textcolor{gray}{4}
    & \quad $\{H_j\}_{j=1}^{D} \leftarrow T(z,y)$
      \hfill
      \makebox[9em][l]{\color{gray}\# frozen}\hspace{0.6em} \\

\textcolor{gray}{5}
    & \quad $\tau_a \leftarrow
      \mathrm{lin}(0,1,\textcolor[HTML]{B02418}{K}),\;\;
      \tau_b \leftarrow [\tau_a^{2:},1]$ \\

\textcolor{gray}{6}
    & \quad $\{\mathbf{h}_k\}_{k=1}^{\textcolor[HTML]{B02418}{K}},\,v
      \leftarrow f_\theta(z,y;\tau_a,\tau_b)$ \\

\textcolor{gray}{7}
    & \quad $\mathcal{L}_\mathbf{h} \leftarrow
      \tfrac{1}{K}\!\sum_k
      \lVert \hat{\mathbf{h}}_k-\hat{H}_{ks}\rVert^2$ \\

\textcolor{gray}{8}
    & \quad $\mathcal{L}_e \leftarrow
      \lVert (z-v)-\x_0^{T}\rVert^2$ \\

\textcolor{gray}{9}
    & \quad $\theta \leftarrow \theta-\eta\,\nabla_\theta
      (\mathcal{L}_h+\lambda\,\mathcal{L}_e)$ \\

\textcolor{gray}{10}
    & \textbf{until} converged

\end{tabularx}
}

\end{tcolorbox}

\end{minipage}
\hfill
%
% ===================== RIGHT: Elastic =====================
\begin{minipage}[t]{0.49\textwidth}
\vspace{0pt}

\phantomsection
\label{alg:elastic}
\noindent{\footnotesize\bfseries Algorithm C.2}
\footnotesize Elastic Training
\par\vspace{0.15em}

\begin{tcolorbox}[
  enhanced,
  colback=white,
  colframe=white,
  boxrule=0pt,
  borderline north={0.8pt}{0pt}{black},
  borderline south={0.8pt}{0pt}{black},
  left=4pt,
  right=4pt,
  top=4pt,
  bottom=4pt,
  before skip=0pt,
  after skip=0pt,
]

{\scriptsize\ttfamily
\renewcommand{\arraystretch}{1.15}

\begin{tabularx}{\linewidth}{
  @{}
  r
  @{\hspace{1.3em}}
  >{\raggedright\arraybackslash}X
  @{}
}

    & \textbf{req:} teacher $T$ (depth $D$), set $\mathcal{K}$ \\

\textcolor{gray}{1}
    & \textbf{repeat} \\

\textcolor{gray}{2}
    & \quad $z \sim \mathcal{N}(0,I),\;\; y \sim p(y)$
      \hfill
      \makebox[9em][l]{\color{gray}\# $y\!\to\!\varnothing$ w.p. $p_{\text{unc}}$}\hspace{0.6em} \\

\textcolor{gray}{3}
    & \quad $\{H_j\}_{j=1}^{D} \leftarrow T(z,y)$
      \hfill
      \makebox[9em][l]{\color{gray}\# frozen}\hspace{0.6em} \\

\cellcolor{gray!15}
\textcolor[HTML]{62A978}{4}
    &
\cellcolor{gray!15}
      \quad $\textcolor[HTML]{B02418}{K} \sim
      \mathrm{Unif}(\mathcal{K}),\;\;
      s \leftarrow D/\textcolor[HTML]{B02418}{K}$ \\

\textcolor{gray}{5}
    & \quad $\tau_a \leftarrow
      \mathrm{lin}(0,1,\textcolor[HTML]{B02418}{K}),\;\;
      \tau_b \leftarrow [\tau_a^{2:},1]$ \\

\textcolor{gray}{6}
    & \quad $\{\mathbf{h}_k\}_{k=1}^{\textcolor[HTML]{B02418}{K}},\,v
      \leftarrow f_\theta(z,y;\tau_a,\tau_b)$ \\

\textcolor{gray}{7}
    & \quad $\mathcal{L}_\mathbf{h} \leftarrow
      \tfrac{1}{K}\!\sum_k
      \lVert \hat{\mathbf{h}}_k-\hat{H}_{ks}\rVert^2$ \\

\textcolor{gray}{8}
    & \quad $\mathcal{L}_e \leftarrow
      \lVert (z-v)-\x_0^{T}\rVert^2$ \\

\textcolor{gray}{9}
    & \quad $\theta \leftarrow \theta-\eta\,\nabla_\theta
      (\mathcal{L}_h+\lambda\,\mathcal{L}_e)$ \\

\textcolor{gray}{10}
    & \textbf{until} converged

\end{tabularx}
}

\end{tcolorbox}

\end{minipage}

\end{figure*}

\textbf{Inference and compute accounting.} Alg.\hyperref[alg:inference]{~C.3} generates a sample by running the student for $K$ applications and decoding once with $\mathbf{W}_{\mathrm{out}}$. With CFG, this is done twice. All FLOPs in Table~\ref{tab:compute_compression} are reported relative to the same student at its native depth $K=L$, with CFG in both cases. Note that the MeanFlow teacher incorporates guidance into training and is evaluated with a single pass ($\omega=1$), so a student at its native depth costs roughly twice the teacher per sample.

\begin{wrapfigure}{r}{0.52\linewidth}
% \vspace{-1.5em}

\phantomsection
\label{alg:inference}
\noindent{\footnotesize\bfseries Algorithm C.3}
\footnotesize Inference (1 NFE Generation)
\par\vspace{0.15em}

\begin{tcolorbox}[
  enhanced,
  colback=white,
  colframe=white,
  boxrule=0pt,
  borderline north={0.8pt}{0pt}{black},
  borderline south={0.8pt}{0pt}{black},
  left=4pt,
  right=4pt,
  top=4pt,
  bottom=4pt,
  before skip=0pt,
  after skip=0pt,
]

{\scriptsize\ttfamily
\renewcommand{\arraystretch}{1.15}

\begin{tabularx}{\linewidth}{
  @{}
  r
  @{\hspace{1.3em}}
  >{\raggedright\arraybackslash}X
  @{}
}

    & \textbf{req:} tied block $b_\theta$, guidance $\omega$ \\

\textcolor{gray}{1}
    &
      choose $\textcolor[HTML]{B02418}{K}$
      \hfill
      {\color{gray}\# fixed: $K$; elastic: $K\!\in\!\mathcal{K}$}
      \hspace{0.6em} \\

\textcolor{gray}{2}
    & $z \sim \mathcal{N}(0,I),\;\; y \sim p(y)$ \\

\textcolor{gray}{3}
    & $\textcolor[HTML]{B02418}{\tau_a}
      \leftarrow
      \mathrm{lin}(0,1,\textcolor[HTML]{B02418}{K}),\;\;
      \textcolor[HTML]{B02418}{\tau_b}
      \leftarrow
      [\textcolor[HTML]{B02418}{\tau_a}^{2:},1]$ \\

\textcolor{gray}{4}
    & \textbf{for } $c \in \{y,\varnothing\}$:
      \hfill
      \makebox[8.5em][l]{\color{gray}\# cond/uncond}
      \hspace{0.6em} \\

\textcolor{gray}{5}
    & \quad $h \leftarrow \mathbf{E}(z)$ \\

\cellcolor{gray!15}
\textcolor[HTML]{62A978}{6}
    &
\cellcolor{gray!15}
      \quad \textbf{for } $k=1..\textcolor[HTML]{B02418}{K}$:\;\;
      $h \leftarrow
      b_\theta
      \!\left(
      h;\,
      \textcolor[HTML]{B02418}{\tau_a}[k],
      \textcolor[HTML]{B02418}{\tau_b}[k]; \tau_\mathrm{task}, c
      \right)$ \\

\textcolor{gray}{7}
    & \quad $v_c \leftarrow \mathbf{W}_{\text{out}}(h)$
      \hfill
      \makebox[8.5em][l]{\color{gray}\# decode once}
      \hspace{0.6em} \\

\textcolor{gray}{8}
    & $v \leftarrow v_{\varnothing}
      + \omega\,(v_y-v_{\varnothing})$
      \hfill
      \makebox[8.5em][l]{\color{gray}\# guidance}
      \hspace{0.6em} \\

\textcolor{gray}{9}
    & $\x_0 \leftarrow z-v$
      \hfill
      \makebox[8.5em][l]{\color{gray}\# clean sample}
      \hspace{0.6em} \\

\textcolor{gray}{10}
    & \textbf{return} $\x_0$

\end{tabularx}
}

\end{tcolorbox}

% \vspace{-1em}
\end{wrapfigure}

% ---------------------------------------------------------------------
\subsection{FD-loss post-training}
\label{app:fd}
% ---------------------------------------------------------------------

After distillation, we optionally fine-tune each student with the FD loss of~\citet{yang2026representation}, which backpropagates directly through the Fr\'echet distance between generated and real feature statistics rather than through a per-sample loss. We compute the distance jointly in three feature spaces, SigLIP, Inception, and MAE, each normalized so that no single space dominates, against precomputed moments of real ImageNet features. The population moments of generated features are tracked with an EMA of decay $0.999$. We fine-tune for $20{,}000$ steps with learning rate $10^{-6}$ and effective batch size $32$ ($4\times8$ gradient accumulation), keeping the CFG scale fixed at the optimum of the distilled model. Unlike distillation, this stage uses statistics of real data.

% ---------------------------------------------------------------------
\subsection{Results}
\label{app:compression_results}
% ---------------------------------------------------------------------

\begin{figure*}[!ht]
\centering
\begin{minipage}{0.56\linewidth}
\includegraphics[width=\textwidth]{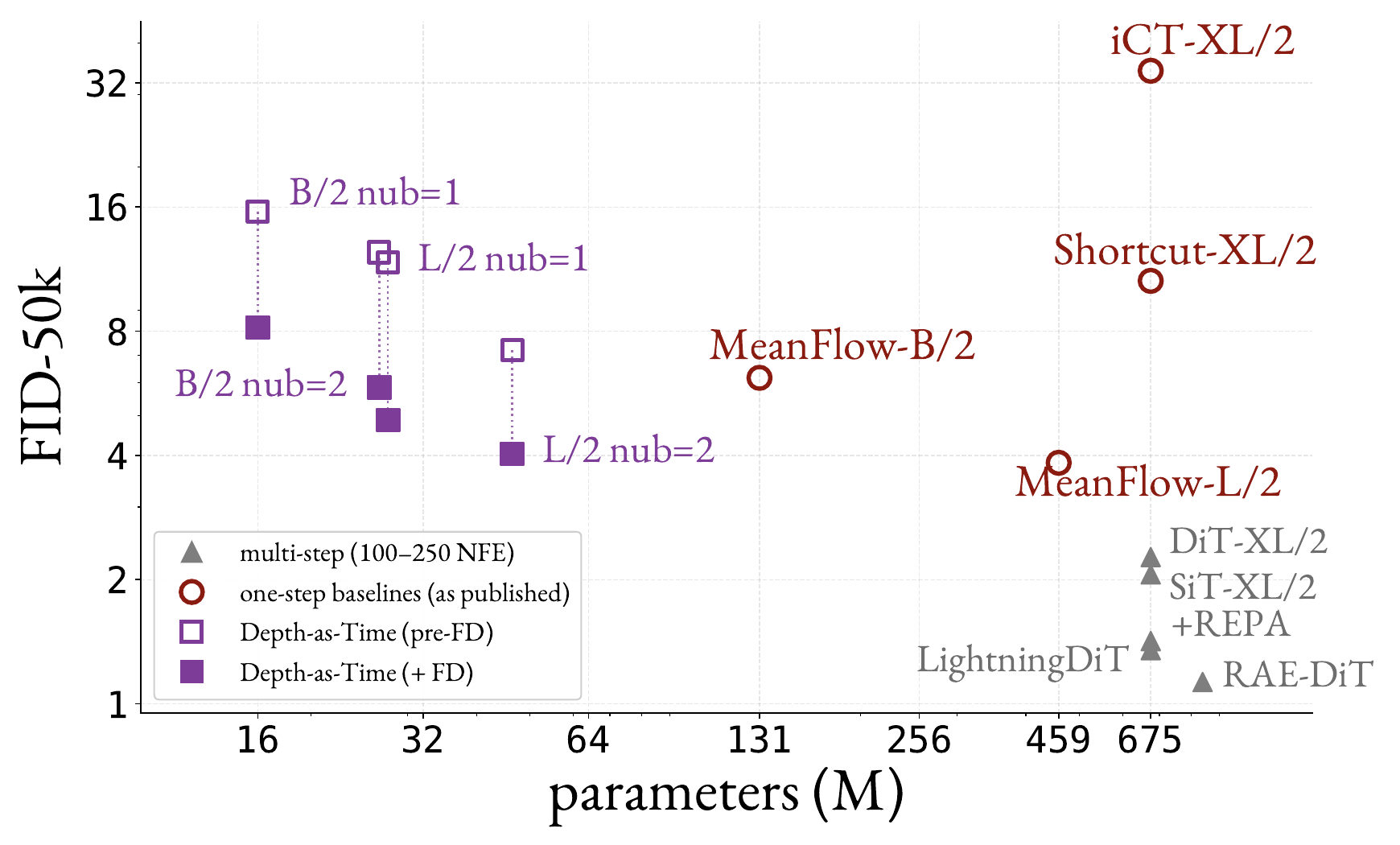}
\end{minipage}
\begin{minipage}{0.43\linewidth}
\caption{%
  \textbf{Parameter efficiency of \textit{depth as time} on ImageNet $256{\times}256$.}
  FID-50k is reported at each model's optimal CFG scale.
  Hollow markers denote models before FD-loss post-training; filled markers denote the same
  models after. All baselines are as reported in their original papers. Our students reach competitive FID with $4.9$--$16.6\times$ fewer parameters than their teachers.
}
\label{fig:params_vs_fid}
\end{minipage}
\end{figure*}

\textbf{Parameter compression.} Table~\ref{tab:param_compression} reports all fixed-$K$ students at the teacher's depth. Adding unique blocks improves FID at the cost of less compression: for \texttt{B/2}, pre-FD FID improves from $15.61$ with one block to $12.43$ with two. FD-loss post-training improves every model by $3$--$7.5$ FID, with larger gains for weaker starting models. Figure~\ref{fig:params_vs_fid} places these models among existing one-step generators, and Figures~\ref{fig:sit_l2_samples}--\ref{fig:sit_l2_samples_r5} show paired samples from the \texttt{SiT-L/2} teacher and its students.

\begin{table}[h]
\centering
\scriptsize
\setlength{\tabcolsep}{4pt}
\begin{tabular}{l|c|c|c|c|c|c|c}
Model & Unique blocks & Params & $K$ & Compression & FID (pre-FD) & FID (post-FD) & $\omega$ \\
\specialrule{\lightrulewidth}{0pt}{0.5pt}
\rowcolor{gray!15}
MeanFlow \texttt{SiT-B/2} (teacher) & 12 & 131M & 12 & -- & 6.17 & -- & -- \\
\rowcolor[HTML]{DFF0E3}
\quad Ours & 1 & 16.0M & 12 & $\mathbf{8.2\times}$ & 15.61 & 8.17 & 1.4 \\
\quad Ours & 2 & 26.6M & 12 & $4.9\times$ & 12.43 & 5.84 & 1.2 \\
\specialrule{\lightrulewidth}{0pt}{0.5pt}
\rowcolor{gray!15}
MeanFlow \texttt{SiT-L/2} (teacher) & 24 & 459M & 24 & -- & 4.01 & -- & -- \\
\rowcolor[HTML]{DFF0E3}
\quad Ours & 1 & 27.6M & 24 & $\mathbf{16.6\times}$ & 11.75 & 4.87 & 1.4 \\
\quad Ours & 2 & 46.5M & 24 & $9.9\times$ & 7.19 & 4.04 & 1.3 \\
\end{tabular}
\caption{\textbf{Parameter compression on ImageNet $256{\times}256$.} FID-50k of fixed-$K$ students at the teacher's depth, before and after FD-loss post-training, each at its optimal CFG scale $\omega$. Compression is relative to the teacher's parameter count. \hlboxg{Highlighted} rows use a single shared block and are the results reported in Section~\ref{sec:explicit_depth_as_time}; grey rows are the teachers. Adding unique blocks trades compression for quality.}
\label{tab:param_compression}
\end{table}

\begin{figure*}[t]
\vspace{-.5em}
\centering
\begin{minipage}[t]{0.32\linewidth}
    \centering
    \includegraphics[width=\linewidth]{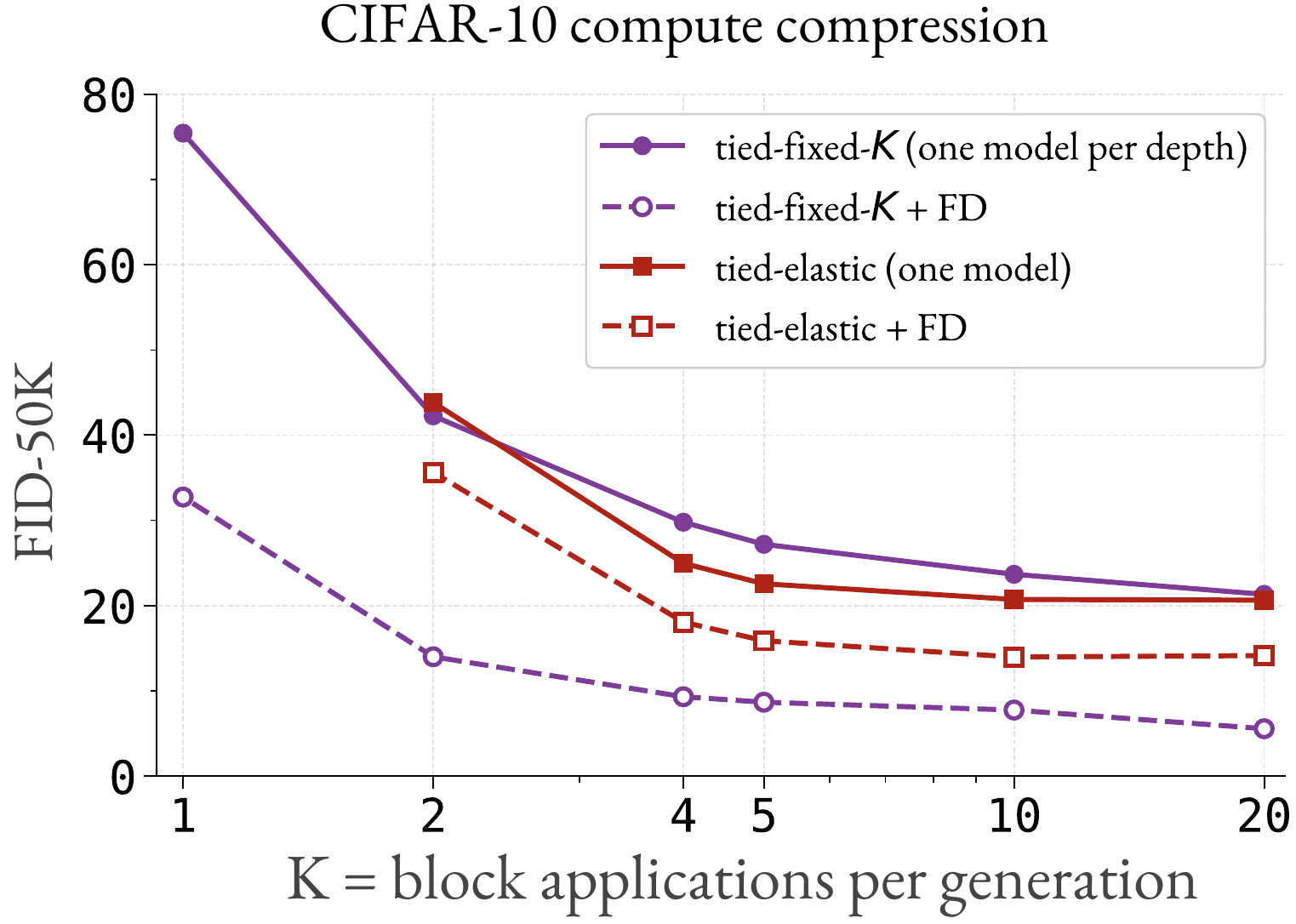}
\end{minipage}
\hfill
\begin{minipage}[t]{0.32\linewidth}
    \centering
    \includegraphics[width=\linewidth]{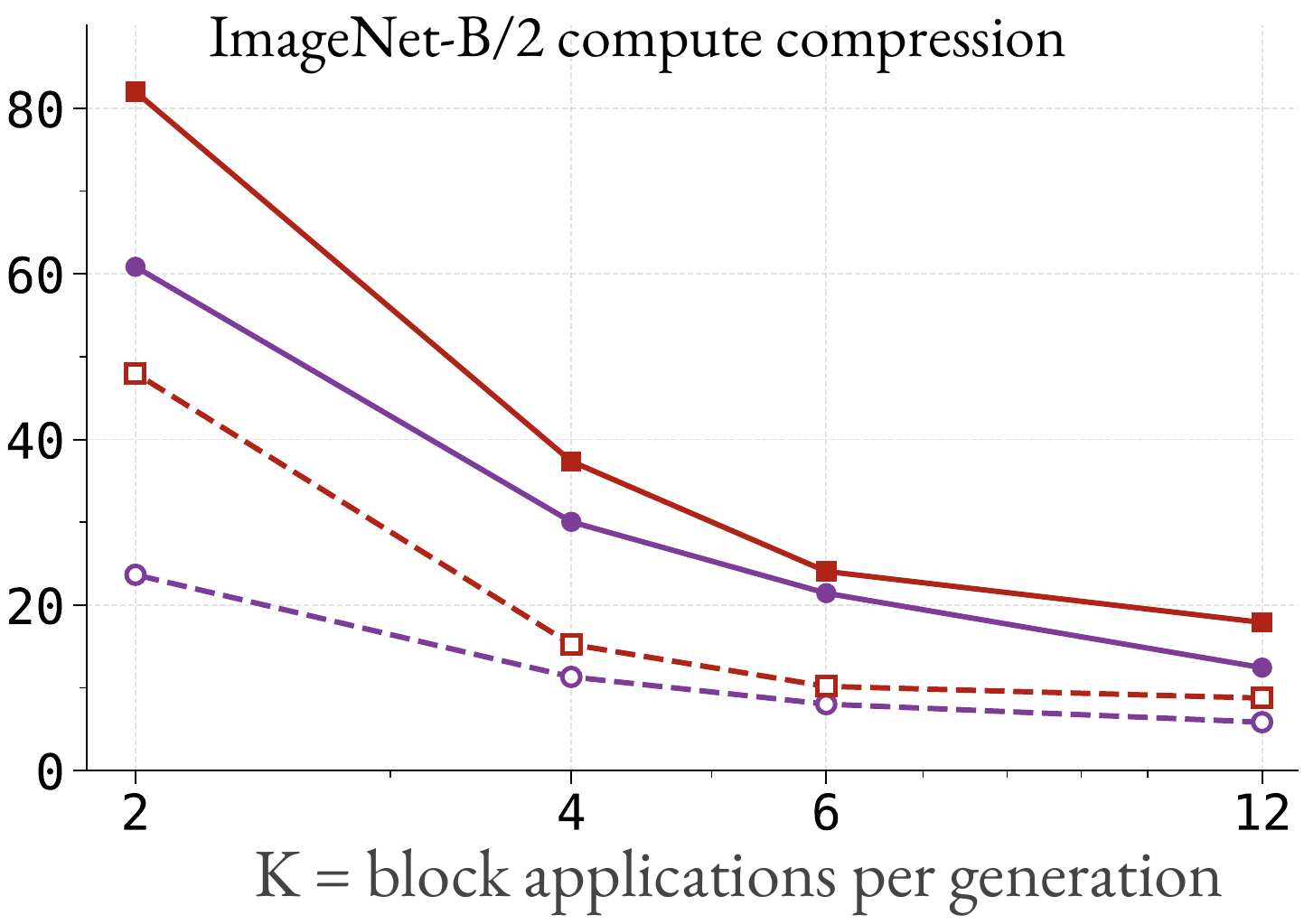}
\end{minipage}
\hfill
\begin{minipage}[t]{0.32\linewidth}
    \centering
    \includegraphics[width=\linewidth]{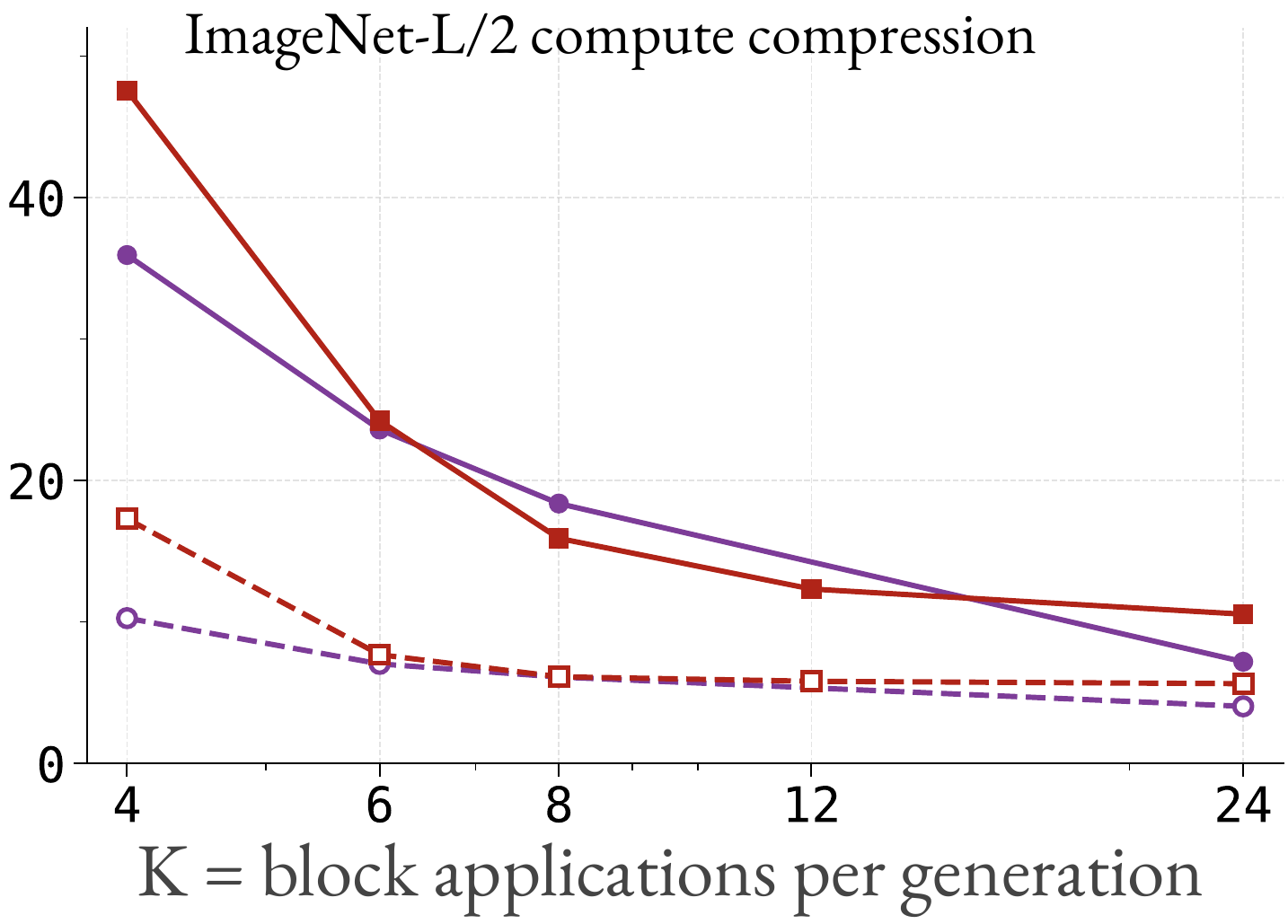}
\end{minipage}
\caption{%
\textbf{Compute compression of depth as time.}
FID-50k at the optimal CFG scale against the number of block applications $K$ per generation; FLOPs scale approximately linearly with $K$. \emph{Fixed-$K$} is a separate tied model trained for each depth; \emph{elastic} is a single tied model run at every $K$. Solid markers are pre-FD, hollow markers post-FD.
}
\label{fig:compute_scaling}
\vspace{-.5em}
\end{figure*}

\textbf{Compute compression.} Table~\ref{tab:compute_compression} reports FID after FD-loss post-training as the number of block applications decreases, for two-block students; Table~\ref{tab:elastic_prefd} gives the corresponding pre-FD numbers for the elastic models, and Figure~\ref{fig:compute_scaling} plots both. Three observations stand out. First, a model trained at its native depth cannot simply be run with fewer, coarser applications: probing the native-depth model below its training depth collapses to FID above $290$. Second, students trained for coarse intervals degrade gracefully, especially at larger scale: the \texttt{L/2} fixed-$K$ student reaches FID $5.25$ with half the block applications and $6.11$ with a third, against a teacher FID of $4.01$. Third, a single elastic model tracks the separately trained fixed-$K$ models closely at moderate compression (\textit{e.g.}, $6.13$ vs.\ $6.11$ for \texttt{L/2} at $K=8$), but falls behind at the most aggressive setting.

\begin{table*}[h]
\centering
\scriptsize
\setlength{\tabcolsep}{2.5pt}

% ===================== LEFT: B/2 =====================
\begin{tabular}[t]{@{}l|c|c|>{\columncolor[HTML]{DFF0E3}}c|>{\columncolor[HTML]{F6DEDB}}c}
\multicolumn{5}{c}{\textbf{\texttt{B/2}} (teacher FID $6.17$)} \\[2pt]
$K$
    & FLOPs $\downarrow$
    & elastic $\downarrow$
    & fixed-$K$ $\downarrow$
    & native, probed $\downarrow$ \\
\specialrule{\lightrulewidth}{0pt}{0.5pt}
12 (native) & $1.00\times$ & 8.75  & \textbf{5.84}  & \textbf{5.84} \\
6           & $0.50\times$ & 10.17 & \textbf{8.02}  & \textcolor[HTML]{808080}{294.68} \\
4           & $0.33\times$ & 15.24 & \textbf{11.31} & \textcolor[HTML]{808080}{313.64} \\
2           & $0.17\times$ & 48.01 & \textbf{23.65} & \textcolor[HTML]{808080}{436.60} \\
\end{tabular}
\hspace{1.2em}
%
% ===================== RIGHT: L/2 =====================
\begin{tabular}[t]{@{}l|c|c|>{\columncolor[HTML]{DFF0E3}}c|>{\columncolor[HTML]{F6DEDB}}c}
\multicolumn{5}{c}{\textbf{\texttt{L/2}} (teacher FID $4.01$)} \\[2pt]
$K$
    & FLOPs $\downarrow$
    & elastic $\downarrow$
    & fixed-$K$ $\downarrow$
    & native, probed $\downarrow$ \\
\specialrule{\lightrulewidth}{0pt}{0.5pt}
24 (native) & $1.00\times$ & 5.64  & \textbf{4.04}  & \textbf{4.04} \\
12          & $0.50\times$ & 5.81  & \textbf{5.25}  & \textcolor[HTML]{808080}{474.65} \\
8           & $0.33\times$ & 6.13  & \textbf{6.11}  & \textcolor[HTML]{808080}{461.09} \\
6           & $0.25\times$ & 7.69  & \textbf{7.04}  & \textcolor[HTML]{808080}{613.43} \\
4           & $0.17\times$ & 17.29 & \textbf{10.27} & \textcolor[HTML]{808080}{637.97} \\
\end{tabular}

\vspace{-0.5em}
\caption{%
\textbf{Compute compression on ImageNet $256{\times}256$.}
FID-50k after FD-loss post-training for two-block students, each at its optimal CFG scale. FLOPs are relative to the same student at its native depth, with CFG in both cases. \emph{Elastic} is a single model trained over the set $\mathcal{K}$ and run at each $K$. \hlboxg{\emph{Fixed-$K$}} is a separate model trained for each $K$, and degrades gracefully as compute decreases. \hlbox{\emph{Native, probed}} is the native-depth model run with fewer, coarser applications than it was trained for, which collapses. 
}
\label{tab:compute_compression}
\vspace{-1em}
\end{table*}

\begin{table}[h]
\centering
\small
\setlength{\tabcolsep}{4pt}
\begin{tabular}[t]{@{}l|cccc@{}}
\multicolumn{5}{c}{\textbf{\texttt{B/2} elastic}, pre-FD} \\[2pt]
$K$ & 12 & 6 & 4 & 2 \\
\specialrule{\lightrulewidth}{0pt}{0.5pt}
FID-50k & 17.89 & 24.06 & 37.36 & 82.02 \\
$\omega$ & 1.5 & 1.5 & 1.5 & 1.7 \\
\end{tabular}
\hspace{1em}
\begin{tabular}[t]{@{}l|ccccc@{}}
\multicolumn{6}{c}{\textbf{\texttt{L/2} elastic}, pre-FD} \\[2pt]
$K$ & 24 & 12 & 8 & 6 & 4 \\
\specialrule{\lightrulewidth}{0pt}{0.5pt}
FID-50k & 10.55 & 12.33 & 15.91 & 24.23 & 47.56 \\
$\omega$ & 1.5 & 1.5 & 1.5 & 1.5 & 1.7 \\
\end{tabular}
\caption{\textbf{Pre-FD FID-50k of the elastic students} in Table~\ref{tab:compute_compression}, at the optimal CFG scale $\omega$ for each $K$.}
\label{tab:elastic_prefd}
\end{table}

% ---------------------------------------------------------------------
\subsection{Ablating the internal time conditioning}
\label{app:looping_ablation}
% ---------------------------------------------------------------------

To test whether the benefit of weight sharing comes from treating depth as time, we compare two weight-tied models that differ only in whether they receive the internal clock. No publicly released DiT-based MeanFlow model exists for CIFAR-10 to distill from, so both are trained from scratch with the MeanFlow $(t,r)$ objective; the CIFAR-10 MeanFlow model analyzed in Fig.~\ref{fig:meanflow_depth_transport} is likewise our own.

\textbf{Setup.} Both models use a \texttt{DiT-B/2} backbone (hidden size $768$, $12$ heads, patch size $2$), class conditioning over the $10$ CIFAR-10 classes with label dropout $0.1$, and EMA decay $0.9999$, and both are trained for $400$k iterations. We compare (i) a single tied block applied $K=20$ times with the internal clock, and (ii) the identical architecture with the clock removed from the AdaLN modulation, so that the block acts identically at every step. Samples are generated in one step ($t=1$, $r=0$) without guidance, and FID is computed against CIFAR-10 training-set statistics.

\begin{table}[h]
\centering
\small
\setlength{\tabcolsep}{3pt}
\begin{tabular}{@{}l|c|c|c|c|c|c|c|c@{}}
Model & Blocks & $K$ & Params & Clock & LR (schedule) & Batch & Steps & FID-50k \\
\specialrule{\lightrulewidth}{0pt}{0.5pt}
Tied + clock & 1 & 20 & 14.4M & \cmark & $10^{-4}$ (cosine) & 64 & 400k & \textbf{21.27} \\
Tied, no clock & 1 & 20 & 14.4M & \xmark & $10^{-4}$ (cosine) & 64 & 400k & \textbf{45.45} \\
\end{tabular}
\caption{\textbf{Internal-clock ablation on CIFAR-10}, trained from scratch. The two models share architecture, parameters, compute, and optimization settings, and differ only in the internal clock. Removing it more than doubles FID-50k ($21.27\to45.45$), showing that the clock is essential to the tied block.}
\label{tab:clock_ablation}
\end{table}

\textbf{Results.} Removing the clock from the tied block more than doubles FID-50k ($21.27\to45.45$; Table~\ref{tab:clock_ablation}). Weight tying therefore works only when each application is told where it is along the computational trajectory.

\newpage
% ---------------------------------------------------------------------
\subsection{Drifting models as a negative control}
\label{app:drifting}
% ---------------------------------------------------------------------

\begin{figure}[h]
    \centering
    \includegraphics[width=0.75\linewidth]{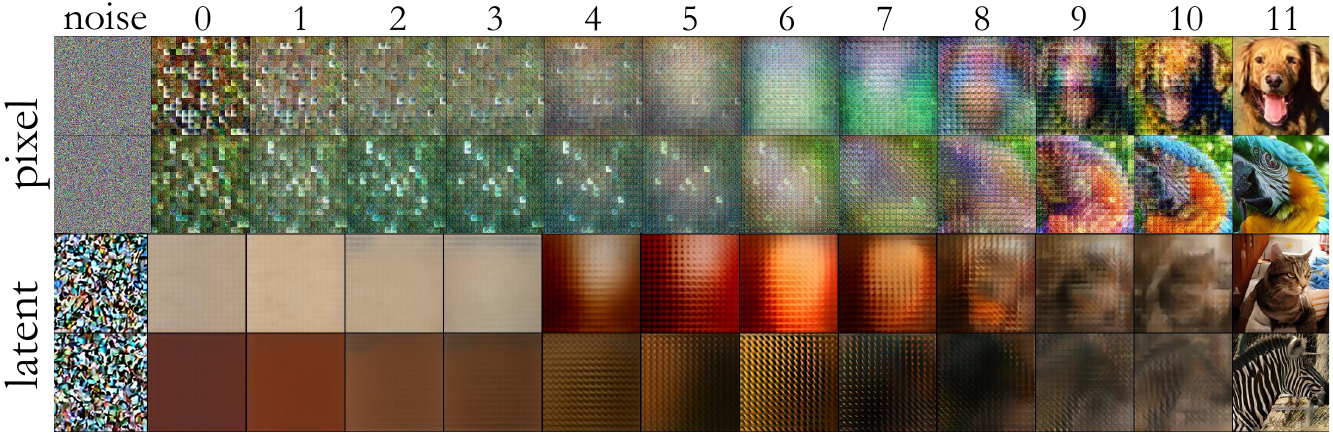}
  \caption{Applying $\mathbf{W}_{\mathrm{out}}$ to the residual states of pixel- and latent-space drifting \texttt{DiT-B/2} models does not reveal the depth-as-time structure observed in task-conditioned models.}
  \label{fig:image_depth_as_time_drift_main}
  \vspace{-1em}
\end{figure}

Drifting models~\citep{deng2026generative} constrain the endpoint distribution without prescribing a time-indexed transport task. Although they can still refine their predictions over depth, their intermediate residual states do not form the same readable trajectory under the model's own $\mathbf{W}_{\mathrm{out}}$ readout; recognizable structure emerges primarily in later blocks (Fig.~\ref{fig:image_depth_as_time_drift_main}).

\textbf{Distillation recipe.} We distill the latent-space drifting \texttt{DiT-B/2} ($12$ blocks, hidden size $768$, $12$ heads, $16$ register tokens) into a single tied block with $K=12$ and the internal clock. The recipe follows the MeanFlow one with two differences. First, because the norm of the teacher's residual stream grows by roughly $141.5\times$ across depth (as opposed to MeanFlow's $22.9\times$), we match $\ell_2$-normalized states, supervising their direction. Second, because drifting models have no transport conditioning, the student receives only the class label and the interval \textcolor[HTML]{B02418}{$(\tau_i,\tau_{i+1})$}, and its output head predicts the image directly.

\begin{table}[h]
\centering
\small
\begin{tabular}{@{}l|c|c|c|c@{}}
Model & Params & Compression & FID-50k & Degradation \\
\specialrule{\lightrulewidth}{0pt}{0.5pt}
Drifting \texttt{B/2} (teacher) & 132.7M & -- & $1.75$ & -- \\
\quad one tied block, $K=12$ & 15.8M & $8.4\times$ & 47.09 & \hlbox{\textbf{${\approx}27\times$}} \\
\specialrule{\lightrulewidth}{0pt}{0.5pt}
MeanFlow \texttt{SiT-B/2} (teacher) & 131M & -- & 6.17 & -- \\
\quad one tied block, $K=12$ & 16.0M & $8.2\times$ & 15.61 & \hlboxg{\textbf{$2.5\times$}} \\
\end{tabular}
\caption{\textbf{Compressing a drifting model versus a MeanFlow model} into a single time-conditioned block (pre-FD, ImageNet $256{\times}256$). Both teachers are compressed by a similar factor, yet the \hlbox{drifting} student degrades far more than the \hlboxg{MeanFlow} student, despite distilling from a stronger teacher.}
\label{tab:drifting_compression}
\end{table}

\textbf{Results.} As Table~\ref{tab:drifting_compression} shows, this procedure does not produce an effective compressed drifting model: at a similar compression ratio, the drifting student reaches FID $47.09$, about three times worse than the MeanFlow student ($15.61$).

 \begin{figure*}[p]
\centering
\setlength{\tabcolsep}{3pt}
\renewcommand{\arraystretch}{1.0}
\begin{tabular}{ccc}
{\small Teacher (\texttt{SiT-L/2}, 459M)} &
{\small Ours, 1 unique block (27.6M)} &
{\small Ours, 2 unique blocks (46.5M)} \\[-1pt]
{\scriptsize FID-50k $4.01$} &
{\scriptsize FID-50k $4.87$ ($16.6\times$ fewer params)} &
{\scriptsize FID-50k $4.04$ ($9.9\times$ fewer params)} \\[2pt]
\includegraphics[width=0.27\textwidth]{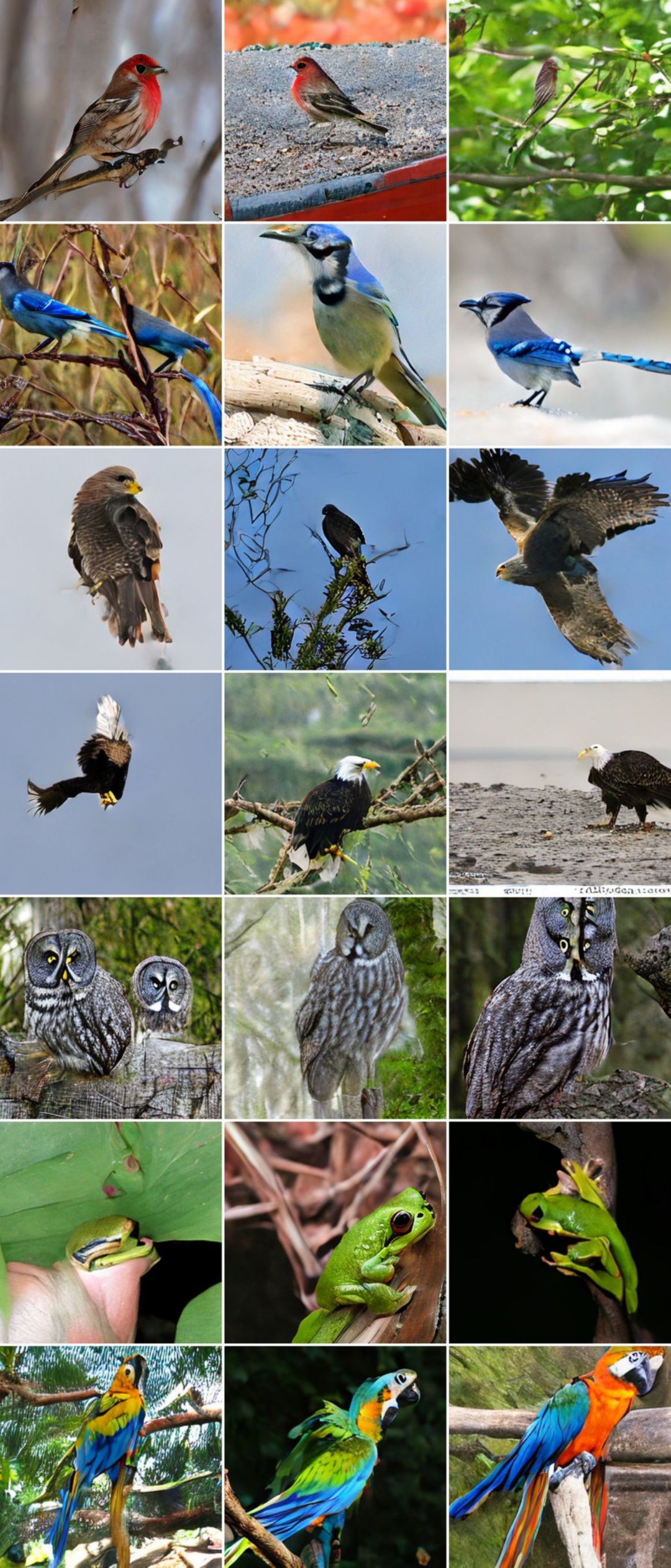} &
\includegraphics[width=0.27\textwidth]{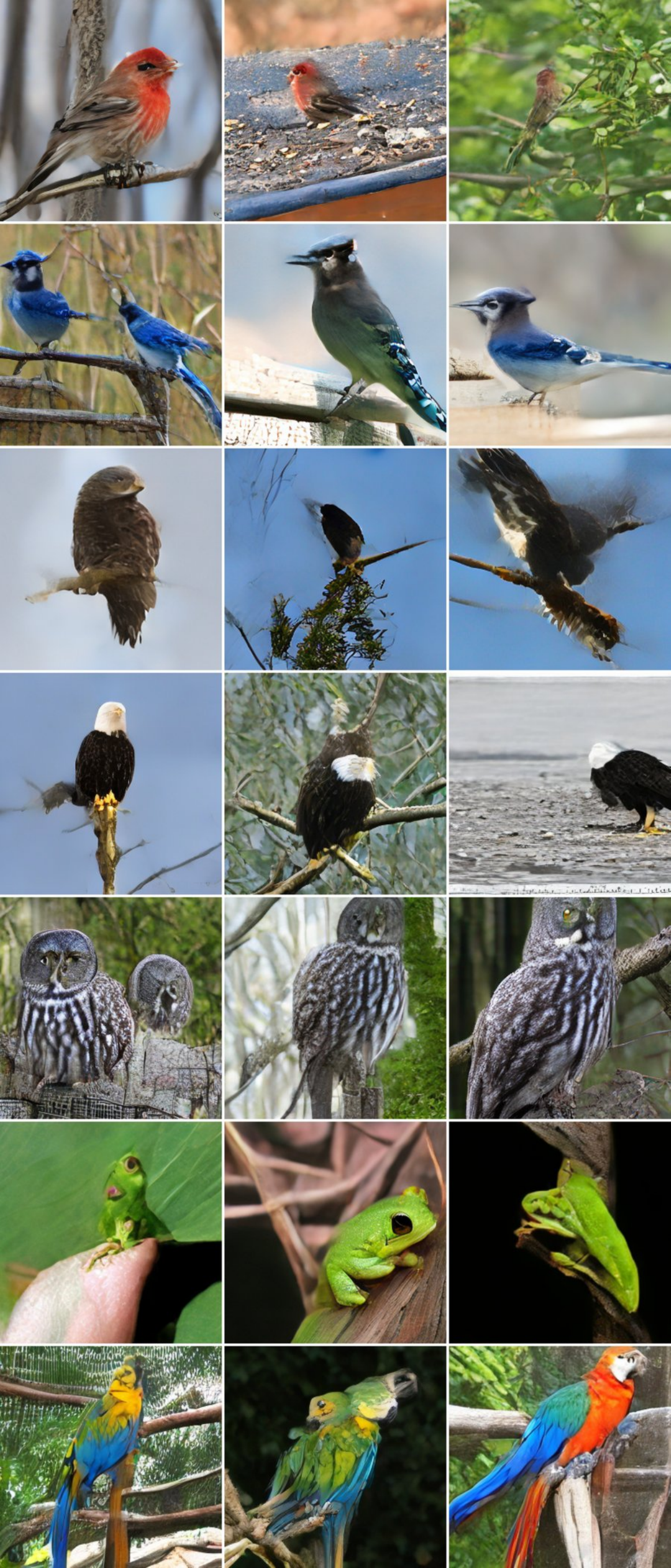} &
\includegraphics[width=0.27\textwidth]{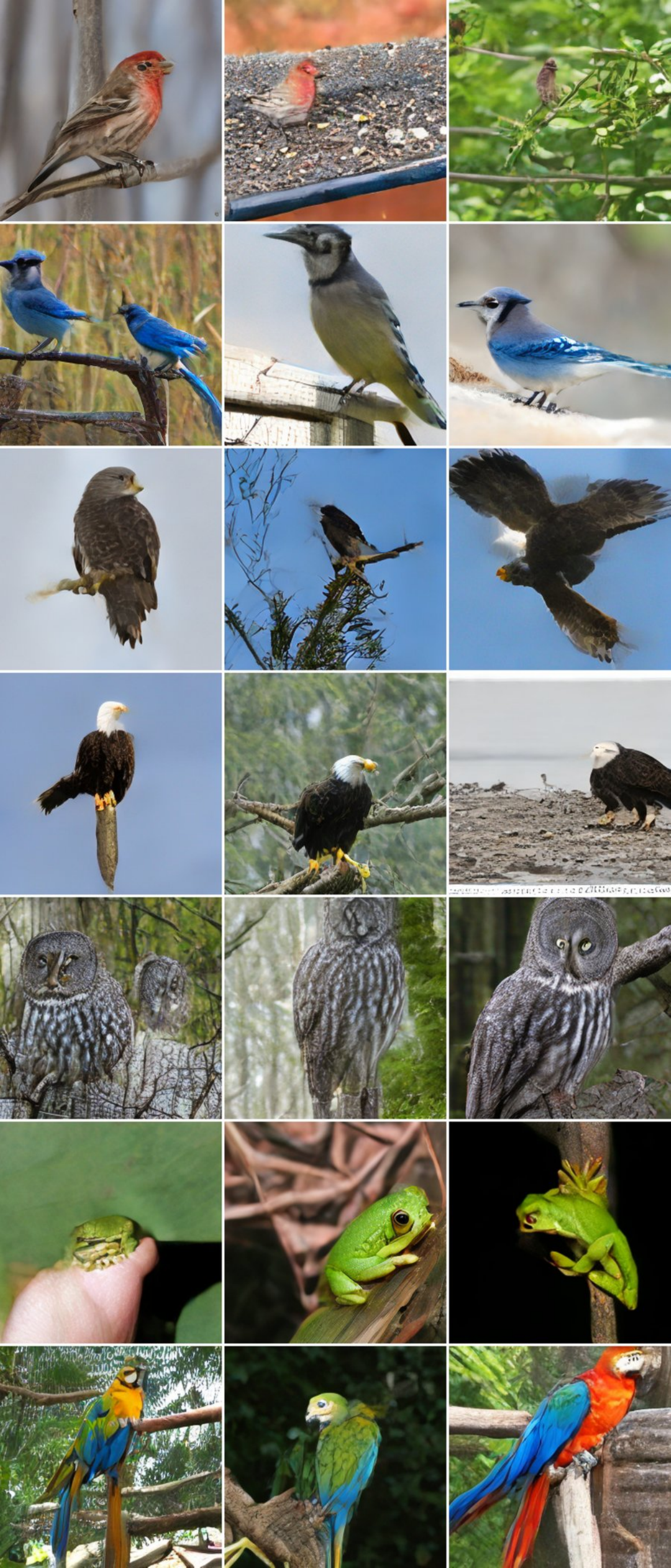} \\[2pt]
\includegraphics[width=0.27\textwidth]{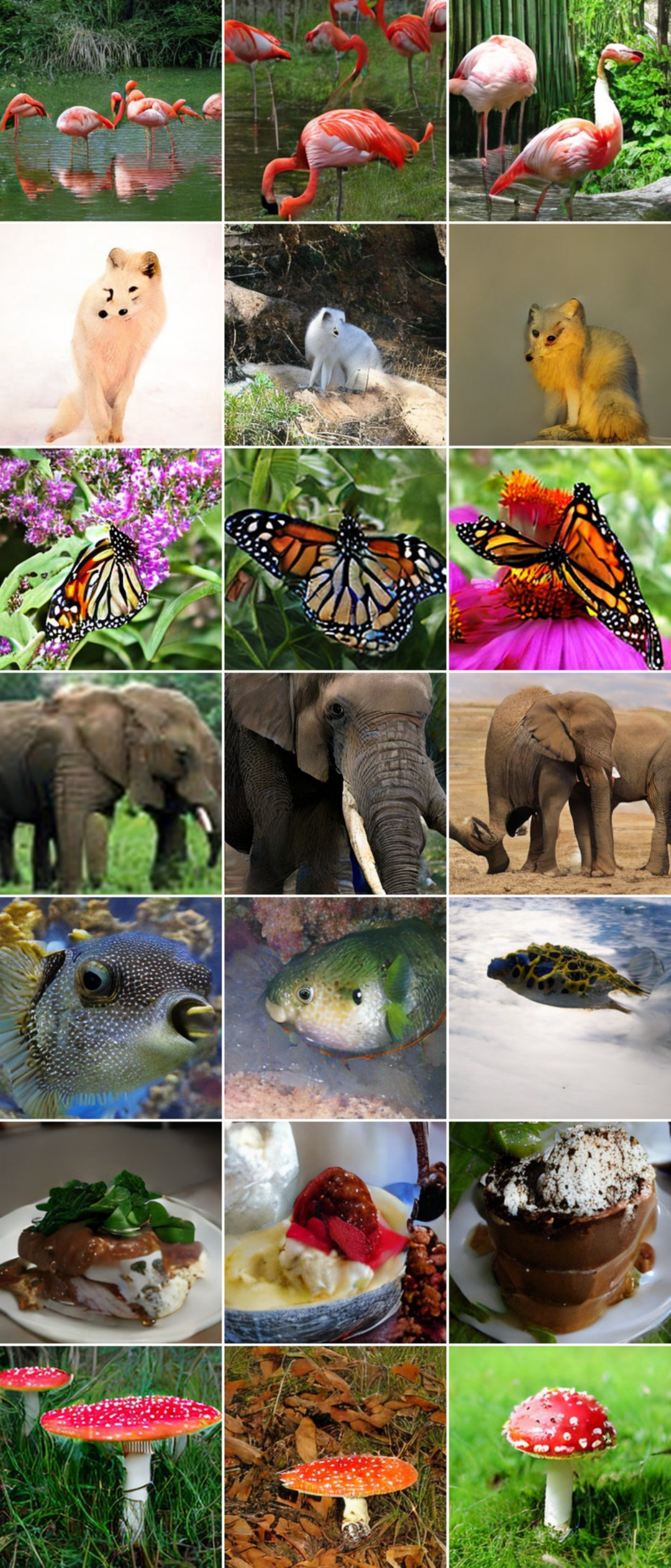} &
\includegraphics[width=0.27\textwidth]{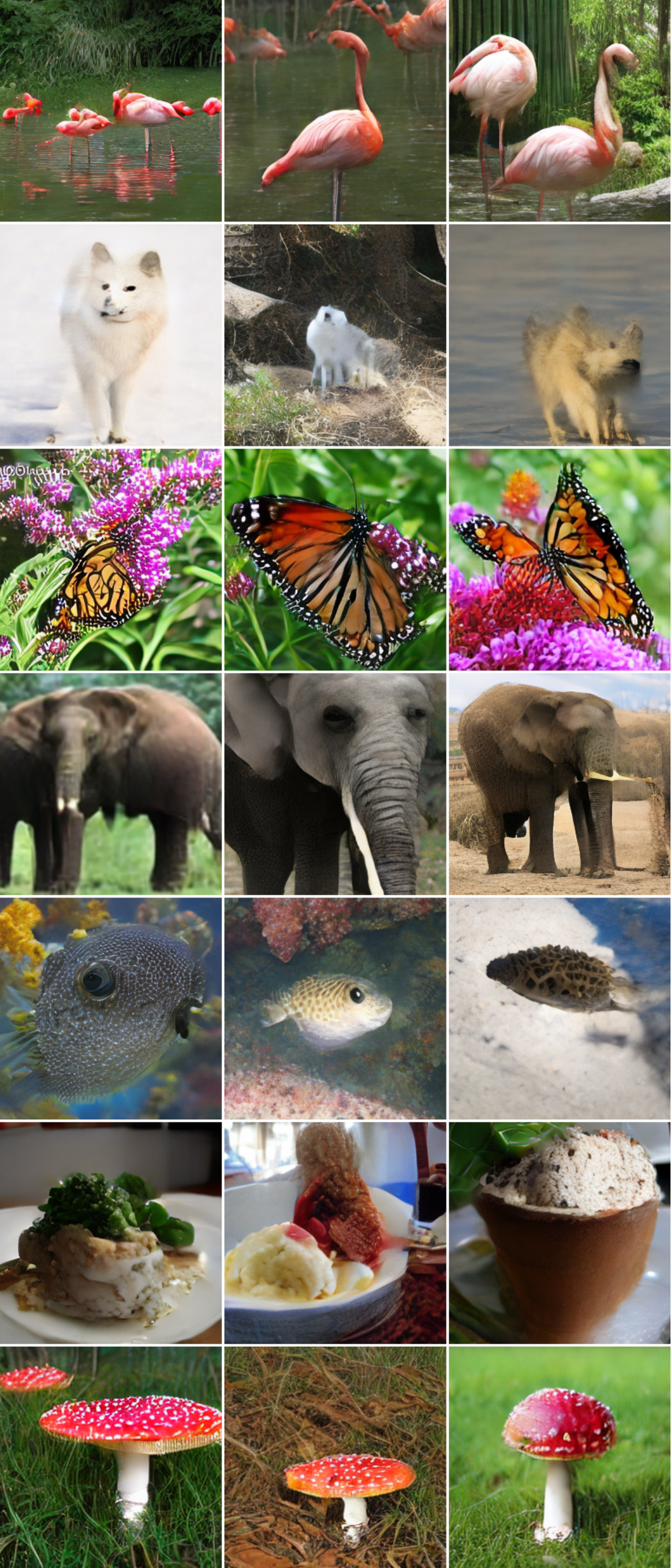} &
\includegraphics[width=0.27\textwidth]{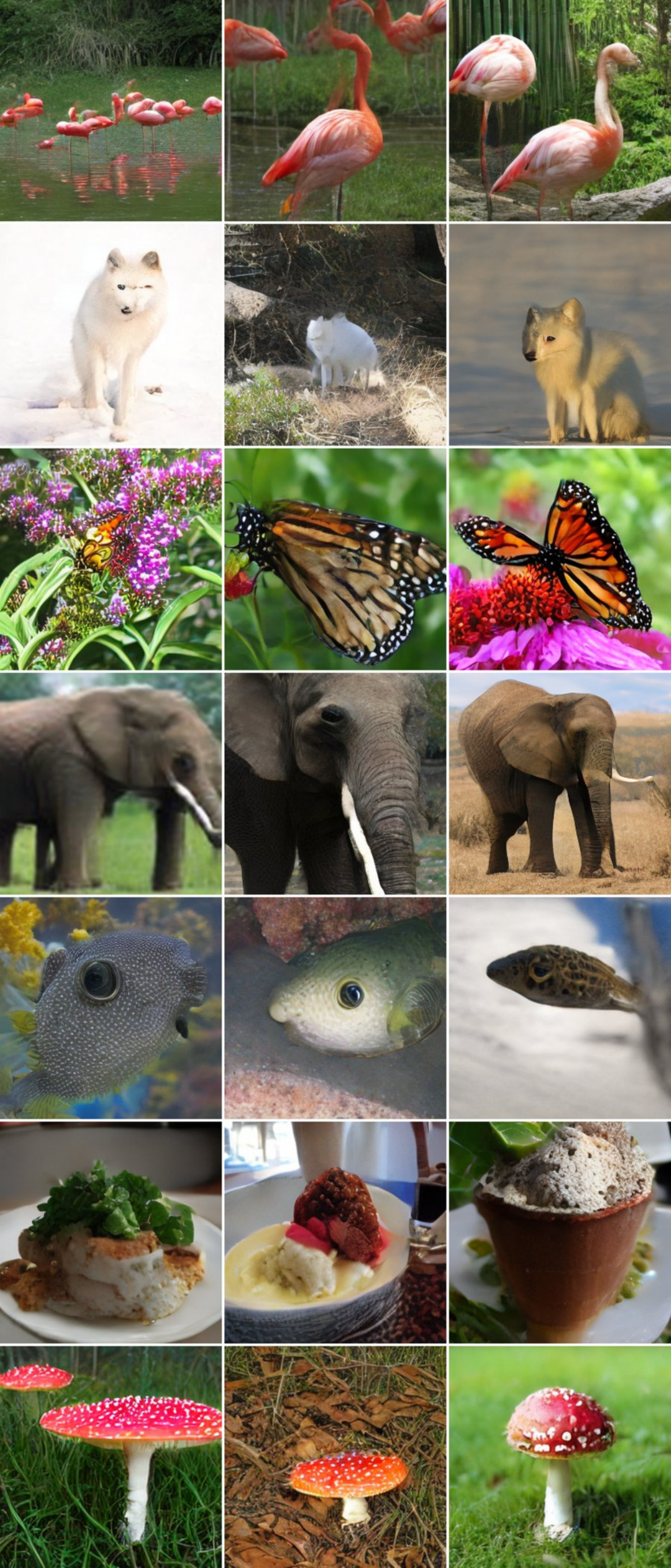} \\
\end{tabular}
\caption{\textbf{Samples from the MeanFlow \texttt{SiT-L/2} teacher and its compressed students} on ImageNet $256{\times}256$. Each row within a panel is one class, and corresponding positions across the three columns use the same initial noise and class label. Students are fixed-$K$ models with $K=24$ after FD-loss post-training (Table~\ref{tab:param_compression}). Despite using $16.6\times$ and $9.9\times$ fewer parameters, the students largely preserve the teacher's composition, pose, and color for each sample, with differences mostly in fine detail.}
\label{fig:sit_l2_samples}
\end{figure*}

\begin{figure*}[p]
\centering
\setlength{\tabcolsep}{2pt}
\renewcommand{\arraystretch}{0.9}
\begin{tabular}{ccc}
{\small Teacher (\texttt{SiT-L/2}, 459M)} &
{\small Ours, 1 unique block (27.6M)} &
{\small Ours, 2 unique blocks (46.5M)} \\[-2pt]
{\scriptsize FID-50k $4.01$} &
{\scriptsize FID-50k $4.87$ ($16.6\times$ fewer params)} &
{\scriptsize FID-50k $4.04$ ($9.9\times$ fewer params)} \\
\includegraphics[width=0.31\textwidth]{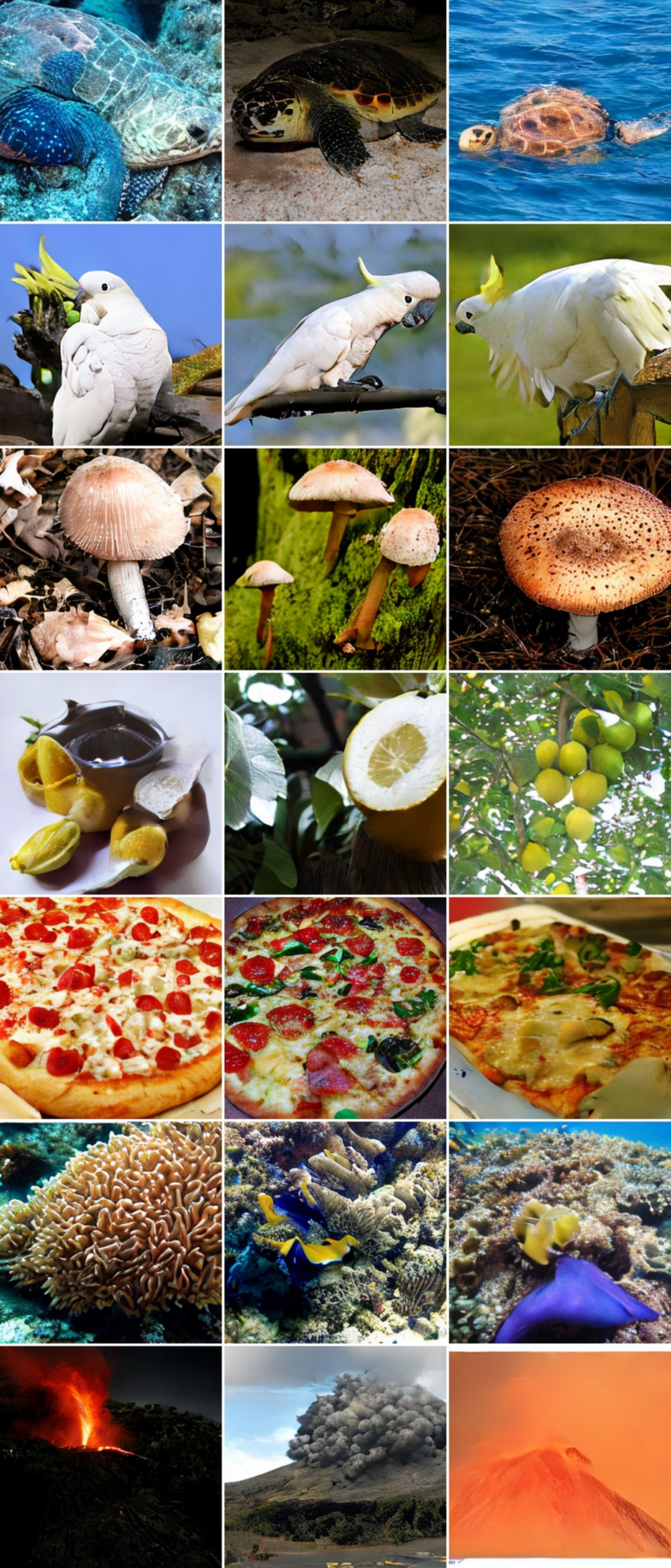} &
\includegraphics[width=0.31\textwidth]{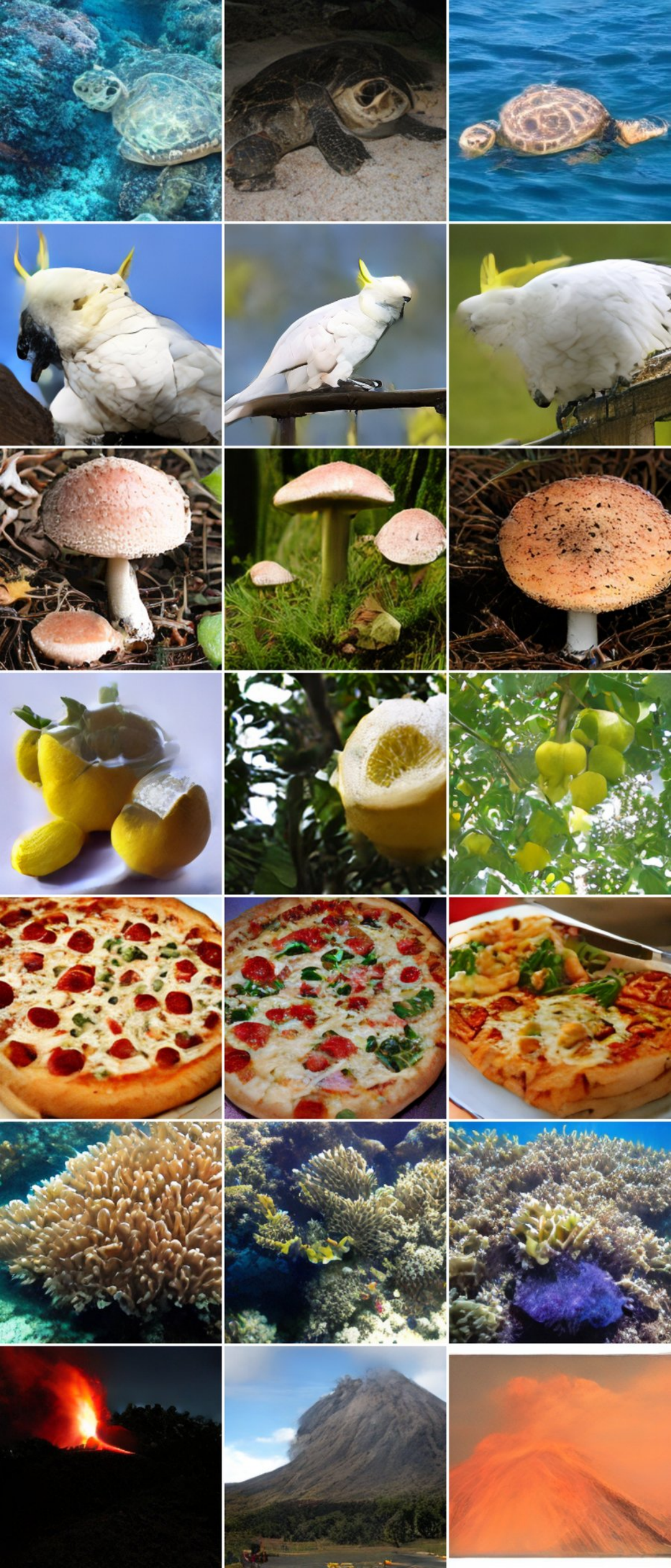} &
\includegraphics[width=0.31\textwidth]{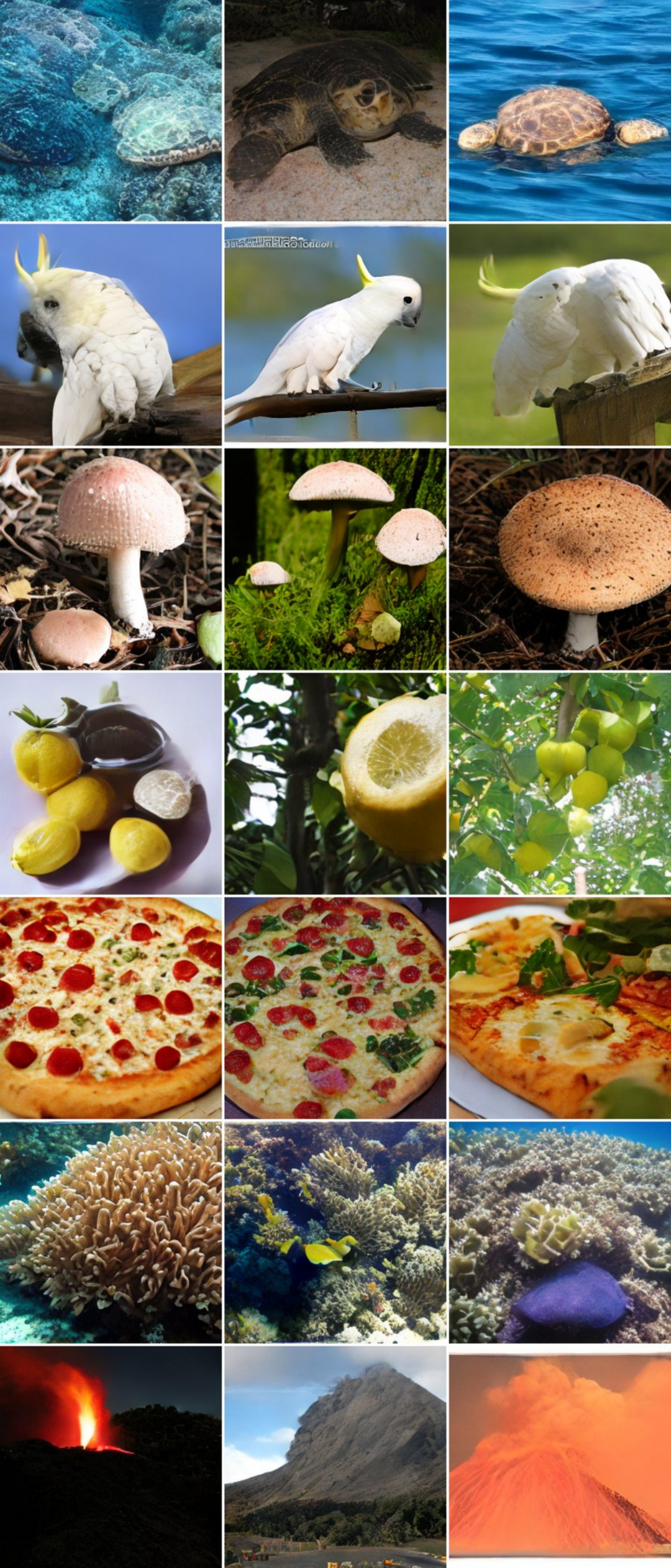} \\
\includegraphics[width=0.31\textwidth]{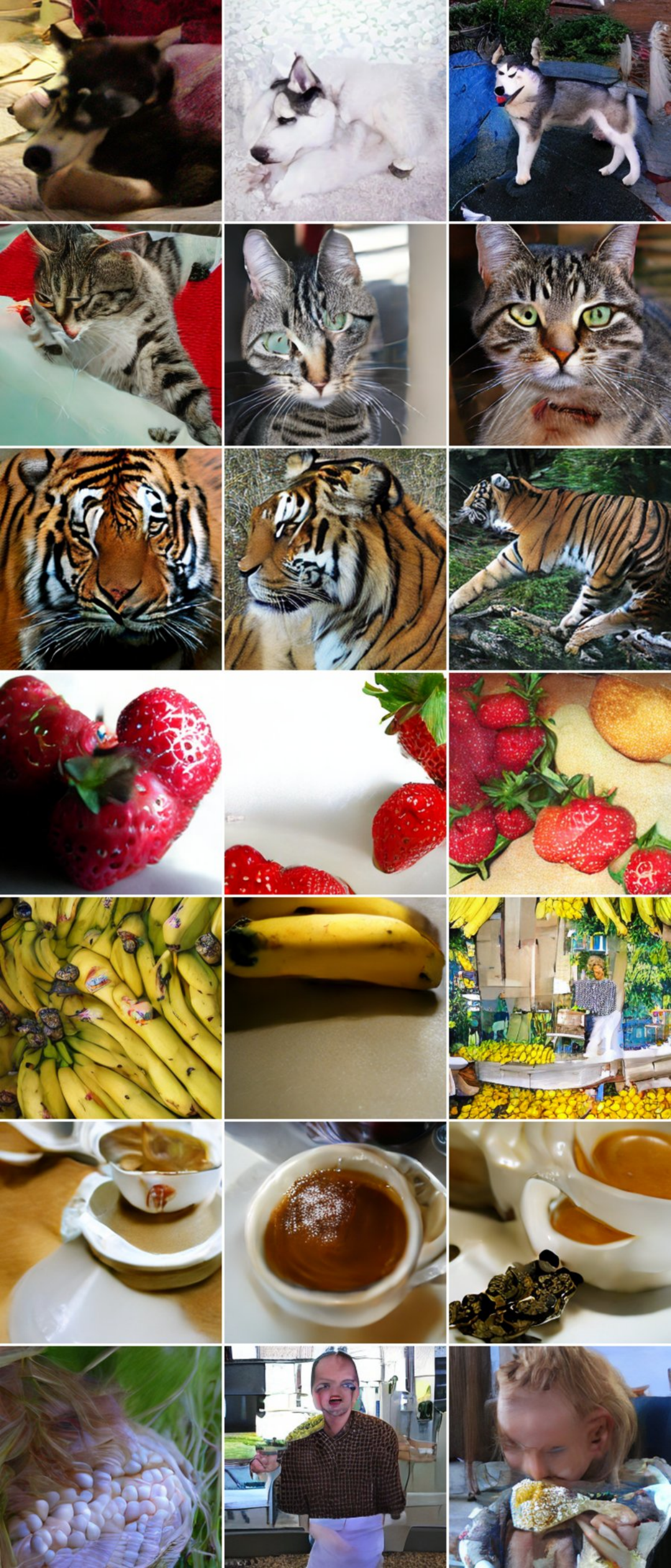} &
\includegraphics[width=0.31\textwidth]{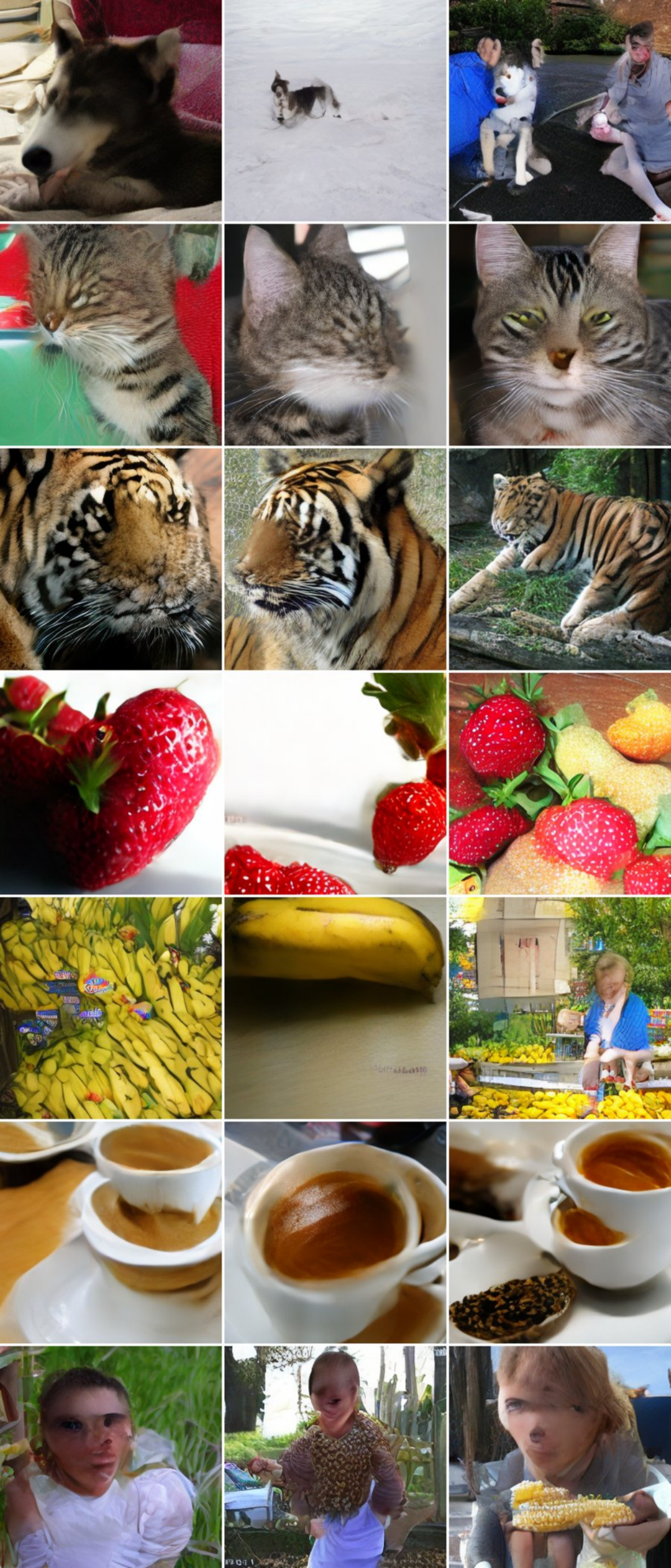} &
\includegraphics[width=0.31\textwidth]{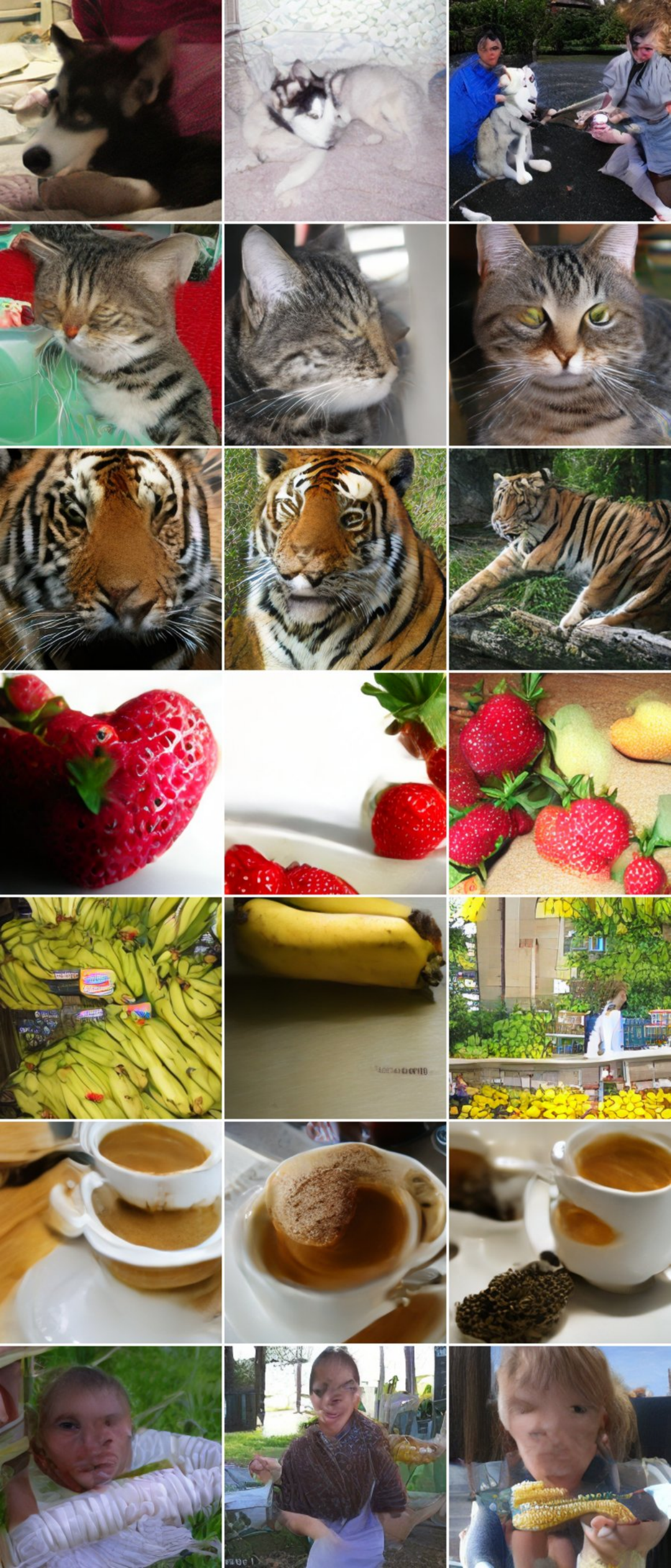} \\
\end{tabular}
\vspace{-0.8em}
\caption{\small\textbf{Additional samples from the MeanFlow \texttt{SiT-L/2} teacher and its compressed students}, arranged as in Figure~\ref{fig:sit_l2_samples}: each row within a panel is one class, and corresponding positions across the three columns use the same initial noise and class label.}
\label{fig:sit_l2_samples_r3}
\end{figure*}

\begin{figure*}[p]
\centering
\setlength{\tabcolsep}{2pt}
\renewcommand{\arraystretch}{0.9}
\begin{tabular}{ccc}
{\small Teacher (\texttt{SiT-L/2}, 459M)} &
{\small Ours, 1 unique block (27.6M)} &
{\small Ours, 2 unique blocks (46.5M)} \\[-2pt]
{\scriptsize FID-50k $4.01$} &
{\scriptsize FID-50k $4.87$ ($16.6\times$ fewer params)} &
{\scriptsize FID-50k $4.04$ ($9.9\times$ fewer params)} \\
\includegraphics[width=0.31\textwidth]{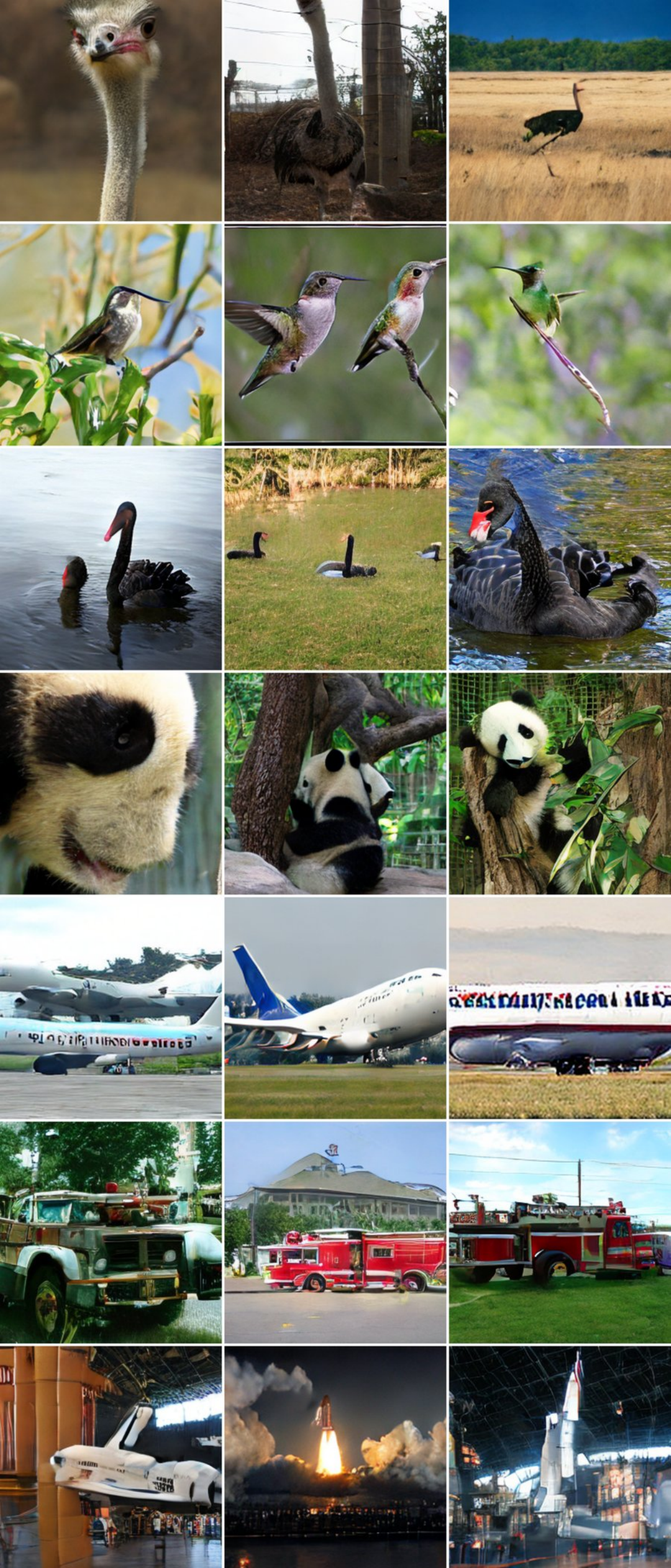} &
\includegraphics[width=0.31\textwidth]{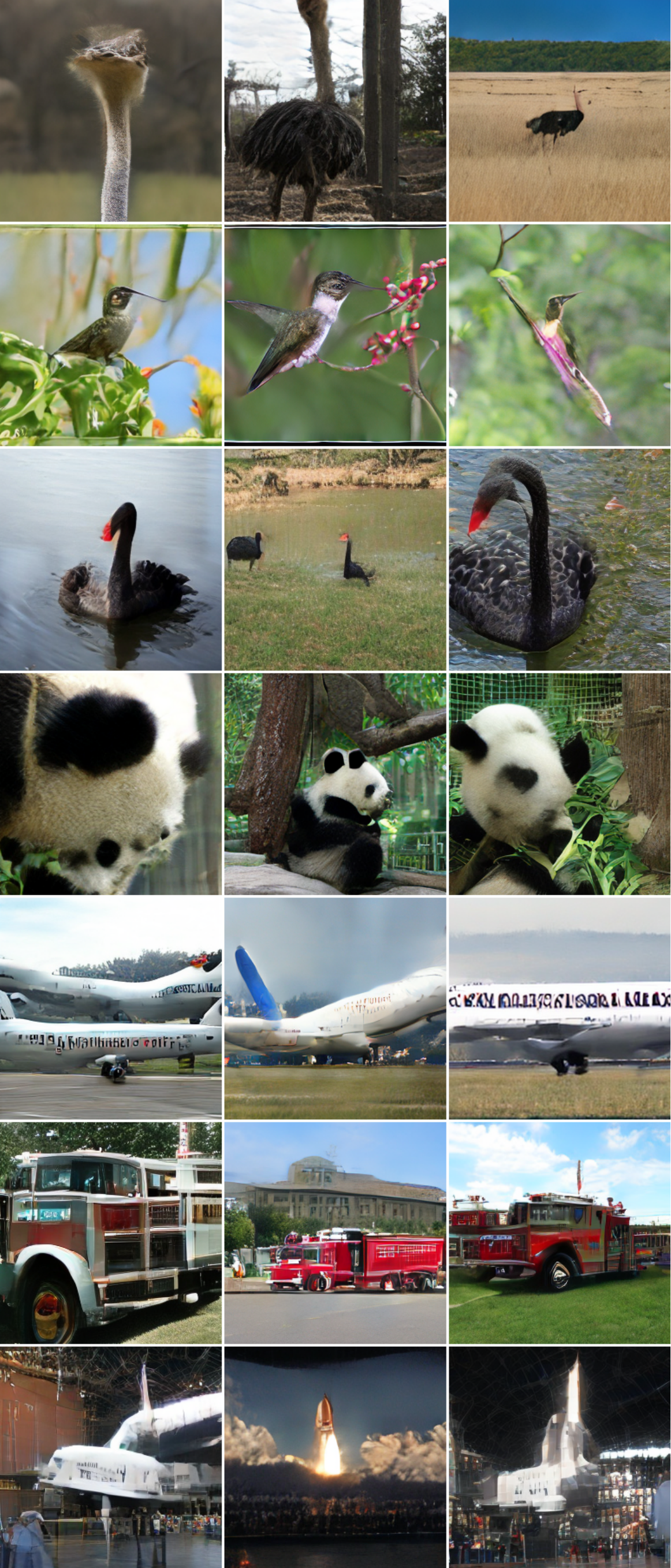} &
\includegraphics[width=0.31\textwidth]{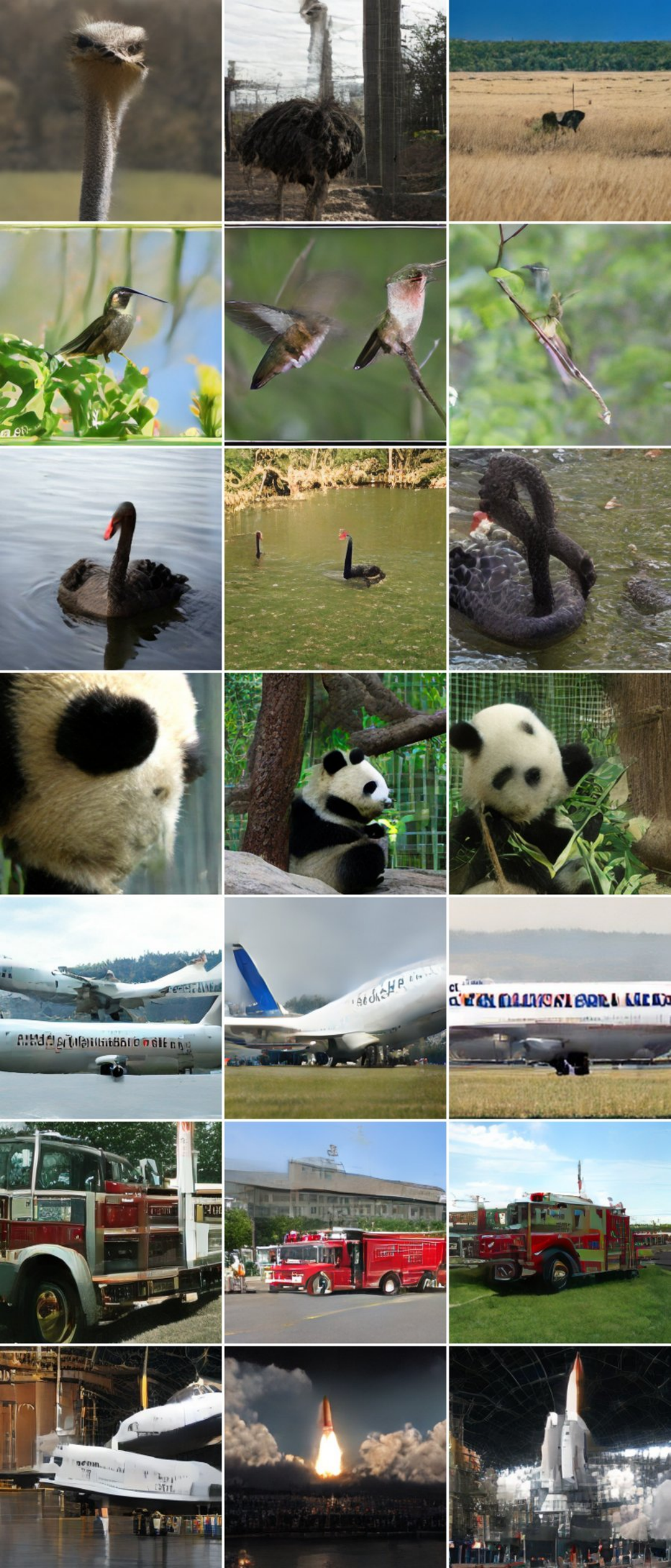} \\
\includegraphics[width=0.31\textwidth]{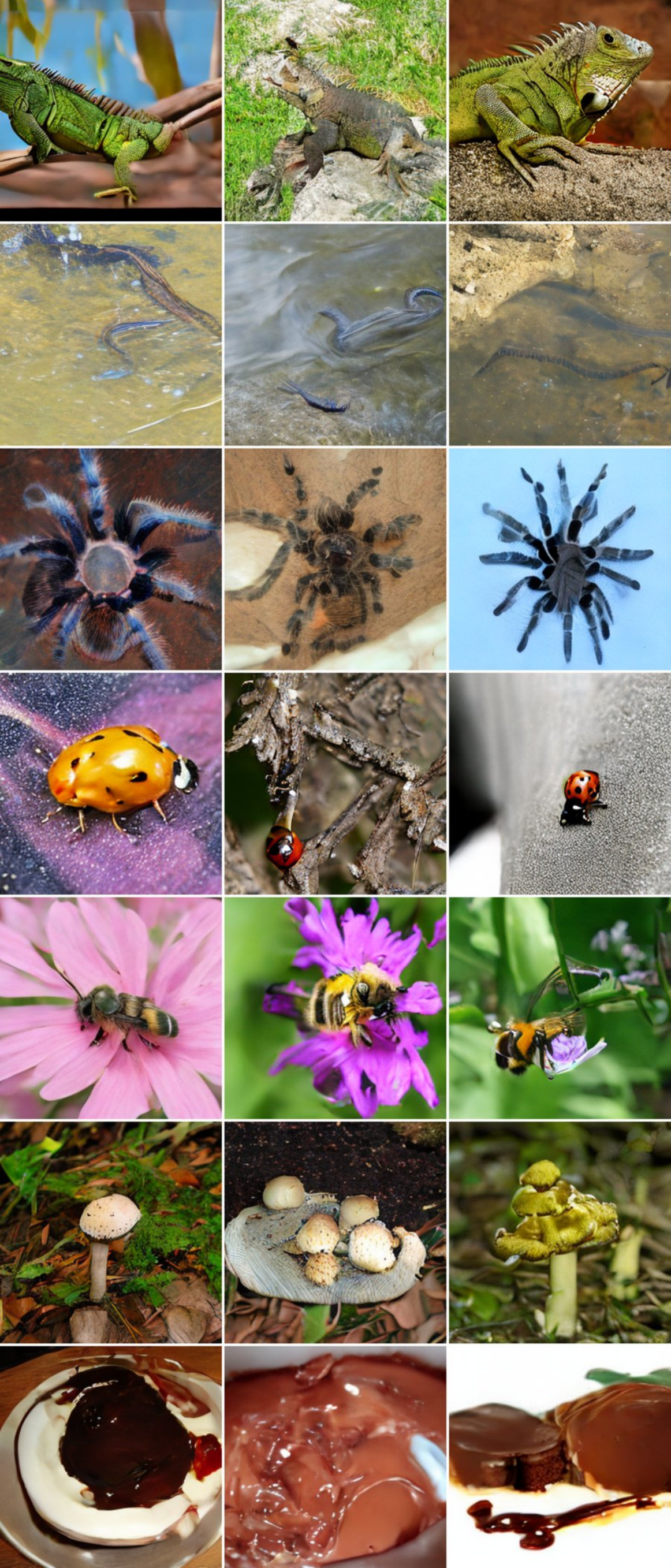} &
\includegraphics[width=0.31\textwidth]{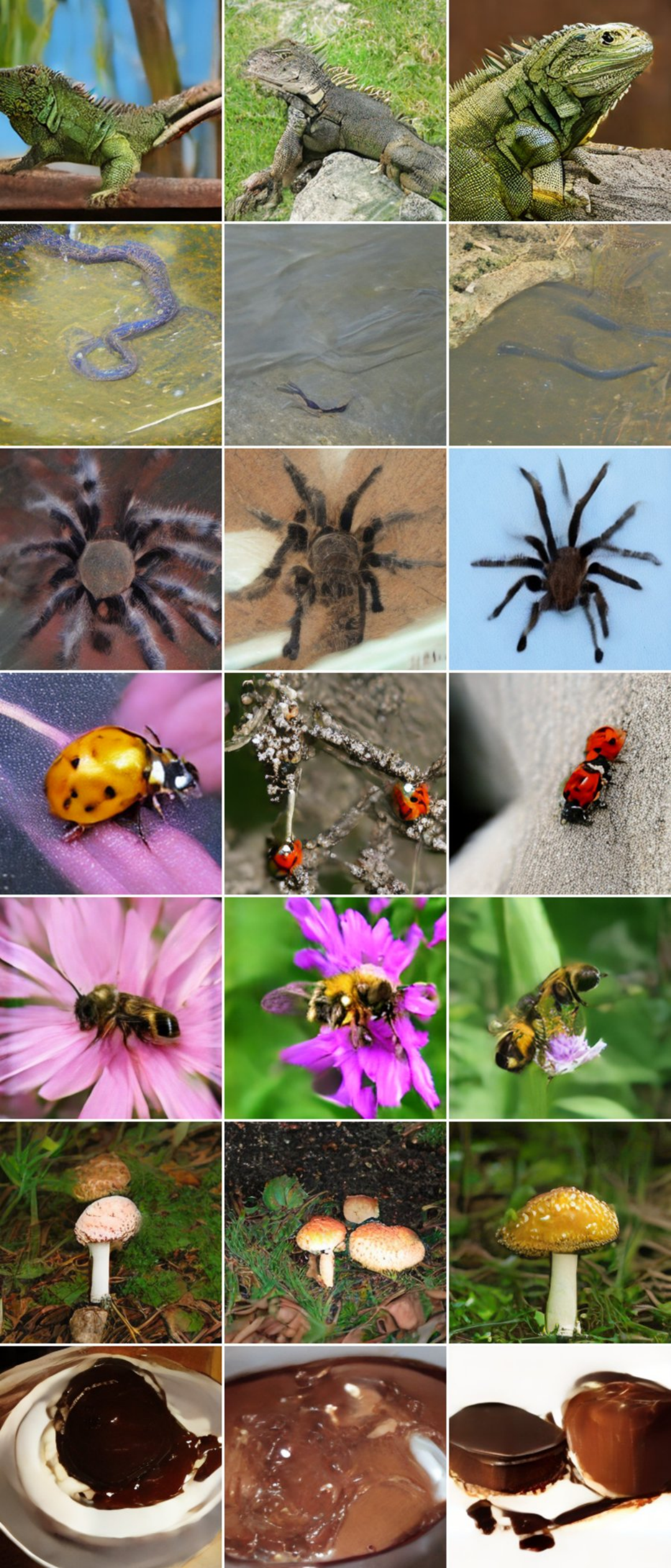} &
\includegraphics[width=0.31\textwidth]{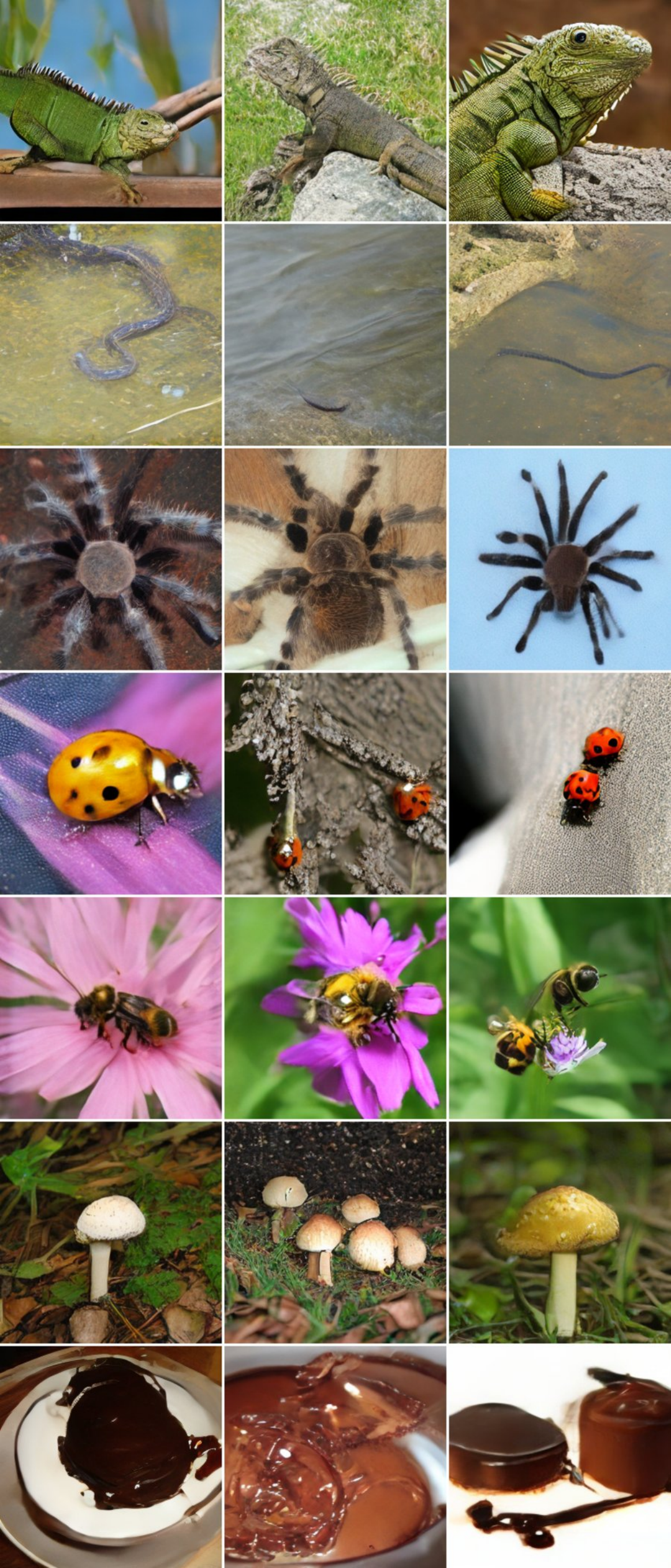} \\
\end{tabular}
\vspace{-0.8em}
\caption{\small\textbf{Additional samples from the MeanFlow \texttt{SiT-L/2} teacher and its compressed students}, arranged as in Figure~\ref{fig:sit_l2_samples}: each row within a panel is one class, and corresponding positions across the three columns use the same initial noise and class label.}
\label{fig:sit_l2_samples_r5}
\end{figure*}

\newpage
\section{Extra Background on One-Step Generators}
\label{app:extra_background}

This appendix reviews the training objectives of the one-step generators analyzed in the main paper, following the taxonomy of Section~\ref{sec:preliminary}. We first cover distillation, where a one-step student is trained against a multi-step teacher (App.~\ref{app:bg_distillation}). We then cover flow maps trained with self-consistency objectives, both boundary-anchored and interval-conditioned (App.~\ref{app:bg_flowmaps}), and step-conditioned (App.~\ref{app:bg_shortcut}). Throughout, we use the convention of Section~\ref{subsec:layer-comparisons}: $\x_t=(1-t)\x_0+t\boldsymbol{\epsilon}$, so that $t=0$ corresponds to data and $t=1$ to noise, with instantaneous velocity $v(\x_t,t)$ and flow map $\Phi_{t\to r}(\x_t)\coloneqq \x_t-\int_r^t v(\x_\tau,\tau)\,d\tau$.

% ---------------------------------------------------------------------
\subsection{Distillation}
\label{app:bg_distillation}
% ---------------------------------------------------------------------

Distillation methods train a one-step generator $G_\theta(z)$ to reproduce the output distribution of a multi-step teacher. A representative example is distribution matching distillation (DMD)~\citep{yin2024one}, commonly expressed as a reverse KL divergence between the noised student and teacher distributions,
$
\mathcal{L}_{\mathrm{DMD}}(\theta)
\coloneqq
\mathbb{E}_{t}\!\left[
\omega(t)\,
\mathbb{E}_{\x_t^\theta \sim p_t^\theta}\!\left[
\log p_t^\theta(\x_t^\theta)
-
\log p_t(\x_t^\theta)
\right]
\right],
$
where $\omega(t)$ is a time-dependent weighting function. The distributions $p_t$ and $p_t^\theta$ use the same Gaussian perturbation kernel $p_t(\x_t\mid\x_0)$ but differ in their clean-data distributions: $p_{\mathrm{data}}$ for real samples and $p_0^\theta$ for samples produced by the student. The optimum is attained when $p_0^{\theta^*}=p_{\mathrm{data}}$. Because neither $\log p_t$ nor $\log p_t^\theta$ is available in closed form, DMD estimates the generator gradient from the difference between real and fake score functions,
\begin{equation}
\nabla_\theta \mathcal{L}_{\mathrm{DMD}}
=
\mathbb{E}_{z,\,t,\,\epsilon}\!\left[
\textcolor[HTML]{808080}{\omega(t)}
\,
\textcolor[HTML]{1E4E8C}{
\big(
s_{\mathrm{fake}}(\x_t^\theta,t)
-
s_{\mathrm{real}}(\x_t^\theta,t)
\big)^{\!\top}
}
\,
\textcolor[HTML]{B02418}{
\frac{\partial G_\theta(z)}{\partial\theta}
}
\right],
\label{eq:dmd_grad}
\end{equation}
where $\x_t^\theta=\alpha_tG_\theta(z)+\sigma_t\boldsymbol{\epsilon}$. Here, $s_{\mathrm{real}}$ and $s_{\mathrm{fake}}$ estimate the scores of $p_t$ and $p_t^\theta$, respectively, and $s_{\mathrm{fake}}$ is detached when differentiating the generator. The score difference \textcolor[HTML]{1E4E8C}{$s_{\mathrm{fake}}-s_{\mathrm{real}}$} is contracted with the student Jacobian \textcolor[HTML]{B02418}{$\partial G_\theta/\partial\theta$}, supervising the student across noise levels $t$.

The distilled models we analyze use related objectives. FLUX-schnell is trained with latent adversarial diffusion distillation (LADD)~\citep{sauer2024fast}, whose discriminator operates on noised samples at varying diffusion times, and SANA-Sprint~\citep{chen2025sana} combines a continuous-time consistency objective with LADD. Although these methods match distributions rather than explicitly reproducing the teacher's sampling trajectory, their learning signal remains indexed by diffusion time. This is one reason to expect their students to organize computation along a denoising-like trajectory across depth, as we observe in Sections~\ref{sec:depth_as_time} and~\ref{subsec:distillation}.

% ---------------------------------------------------------------------
\subsection{Boundary-anchored and interval-conditioned flow maps}
\label{app:bg_flowmaps}
% ---------------------------------------------------------------------

Consistency models~\citep{song2023consistency} learn a flow map whose endpoint is fixed to the clean sample, $f_\theta(\x_t,t)\approx\Phi_{t\to0}(\x_t)$. Subsequent work lifts this restriction and learns transport between arbitrary pairs of times $t$ and $r$. Consistency trajectory models (CTM)~\citep{kim2024consistency} and MeanFlow (MF)~\citep{geng2025mean} are two instances of this more general formulation.

\textbf{Parameterization.} CTM parameterizes the map from $t$ to $r$ with a displacement $g_\theta$, whereas MeanFlow parameterizes the same transport with an interval-averaged velocity
$u_\theta(\x_t,t,r) \approx \frac{1}{t-r}\int_r^t v(\x_\tau,\tau)\,d\tau$.
Both approximate the flow map:
\begin{equation}
\Phi_{t \to r}(\x_t)
\approx
\underbrace{
\tfrac{r}{t}\,\x_t
+
\tfrac{t-r}{t}\,g_\theta(\x_t,t,r)
}_{\mathrm{CTM}}
=
\underbrace{
\x_t
-
(t-r)\,u_\theta(\x_t,t,r)
}_{\mathrm{MF}},
\label{eq:ctm_mf_relationship}
\end{equation}
mapping the current state $\x_t$ to the endpoint $\x_r$.\footnote{We refer the reader to Chapter~11 of~\citet{lai2025principles} for a detailed relationship between the two parameterizations.}

\textbf{Training.} Both methods construct their targets largely from the model's own predictions. Writing $G_\theta(\x_t,t,r)$ for CTM's estimate of $\Phi_{t\to r}(\x_t)$, CTM enforces self-consistency by requiring a direct map from $t$ to $r$ to agree with one that passes through an intermediate time $s\in[r,t]$,
$
\mathcal{L}_{\mathrm{CTM}}(\theta)
=
\mathbb{E}_{t,s,r}\,
\mathbb{E}_{\x_t\sim p_t}\!\left[
\left\|
G_\theta(\x_t,t,r)
-
G_{\theta^-}\!\left(
\hat\Phi_{t\to s}(\x_t),s,r
\right)
\right\|_2^2
\right],
$
where $\hat\Phi_{t\to s}$ is obtained by numerically solving a pretrained teacher's ODE from $t$ to $s$, and $\theta^-$ denotes detached parameters. MeanFlow instead trains the interval-averaged velocity using the instantaneous velocity together with a correction involving the total derivative of the interval average,
$
\mathcal{L}_{\mathrm{MF}}(\theta)
=
\mathbb{E}_{t,r}\,
\mathbb{E}_{\x_t\sim p_t}\!\left[
\left\|
u_\theta(\x_t,t,r)
-
\mathrm{sg}\!\left(
v(\x_t,t)
-
(t-r)\frac{d}{dt}u_\theta(\x_t,t,r)
\right)
\right\|_2^2
\right],
$
where $\mathrm{sg}(\cdot)$ denotes the stop-gradient operator.

\textbf{Clean versus interval endpoints.} The distinction between clean-endpoint and interval-endpoint prediction is central to the depthwise behavior we observe. When $r=0$, the model transports $\x_t$ directly to the clean sample $\x_0$. In a multi-step flow map, however, $r$ may be greater than zero, so the target $\x_r$ remains a noisy intermediate state. MeanFlow is therefore trained to predict an average velocity anchored to the interval endpoint $\x_r$, rather than always to the clean image $\x_0$. As we show in Section~\ref{subsec:flow_map}, this difference in the training objective changes the structure of the trajectory across depth, producing the denoise-then-renoise pattern.

% ---------------------------------------------------------------------
\subsection{Step-conditioned flow maps: Shortcut models}
\label{app:bg_shortcut}
% ---------------------------------------------------------------------

Shortcut models~\citep{frans2025one} condition the network not only on the current time $t$ but also on a desired step size $d$. The model learns a \emph{shortcut} $s_\theta(\x_t,t,d)$ that transports $\x_t$ to the state a distance $d$ closer to the data,\footnote{We state Shortcut models in our time convention, in which $t=0$ is data; the original formulation uses the reverse convention.}
$
\x'_{t-d}
=
\x_t
-
d\,s_\theta(\x_t,t,d),
$
so that $r=t-d$ in the notation of Section~\ref{sec:preliminary}. As $d\to 0$, the shortcut reduces to the instantaneous velocity.

Rather than regressing onto a teacher-defined target, Shortcut models exploit a self-consistency relation: one step of size $2d$ should agree with two consecutive steps of size $d$,
$
s_\theta(\x_t,t,2d)
=
\tfrac{1}{2}s_\theta(\x_t,t,d)
+
\tfrac{1}{2}s_\theta(\x'_{t-d},t-d,d).
$
This yields a joint objective that combines flow matching at $d=0$ with self-consistency at $d>0$,
\begin{equation}
\mathcal{L}_{S}(\theta)
=
\mathbb{E}_{\x_0\sim p_{\mathrm{data}},\,\boldsymbol{\epsilon}\sim\mathcal{N}(\mathbf{0},\mathbf{I}),\,(t,d)}
\left[
\left\|
s_\theta(\x_t,t,0)
-
(\boldsymbol{\epsilon}-\x_0)
\right\|_2^2
+
\left\|
s_\theta(\x_t,t,2d)
-
s_{\mathrm{target}}
\right\|_2^2
\right],
\label{eq:shortcut_objective}
\end{equation}
where
$
s_{\mathrm{target}}
=
\mathrm{sg}\!\left(
\tfrac{1}{2}s_\theta(\x_t,t,d)
+
\tfrac{1}{2}s_\theta(\x'_{t-d},t-d,d)
\right)
$
is constructed from the model's own predictions. The flow-matching term anchors the model to the instantaneous velocity $\boldsymbol{\epsilon}-\x_0$, while the self-consistency term propagates this capability from short steps to progressively larger ones, including the one-step setting.

\section{Extra Analyses on Flow Maps}
\label{app:flow_maps}

\subsection{Shortcut models}
\label{app:shortcut}

\textbf{Setup.}

\begin{wrapfigure}{r}{0.4\linewidth}
\vspace{-1em}

\phantomsection
\label{alg:shortcut_probe}
\noindent{\footnotesize\bfseries Algorithm E.1} \footnotesize Shortcut Depth Probe\par\vspace{0.15em}

\begin{tcolorbox}[
  enhanced,
  colback=white,
  colframe=white,
  boxrule=0pt,
  borderline north={0.8pt}{0pt}{black},
  borderline south={0.8pt}{0pt}{black},
  left=4pt,
  right=4pt,
  top=4pt,
  bottom=4pt,
  before skip=0pt,
  after skip=0pt,
]

{\scriptsize\ttfamily
\renewcommand{\arraystretch}{1.03}
\begin{tabular}{@{}>{\color{gray}}r@{\hspace{0.55em}}l@{}}
1 & $\z \leftarrow \z_1$ \\
2 & \textbf{for} $\, (t,d) \in
      \{(1,0.5),(0.5,0.5)\}\,$ \textbf{do} \\
3 & \quad $\{\textcolor[HTML]{B02418}{\mathbf{h}_i}\}_{i=1}^L
      \leftarrow s_\theta(\z;\textcolor[HTML]{B02418}{t,d})$ \\
4 & \quad \textbf{for} $\,i=1,\ldots,L\,$ \textbf{do} \\
5 & \quad\quad $\textcolor[HTML]{B02418}{s_i}
      \leftarrow \mathbf{W}_{\mathrm{out}}
      (\textcolor[HTML]{B02418}{\mathbf{h}_i};t,d)$ \\
6 & \quad\quad $\textcolor[HTML]{B02418}{\tilde{\x}_{t-d}^{\,i}}
      \leftarrow
      \z - d\,\textcolor[HTML]{B02418}{s_i}$ \\
7 & \quad\quad $\textcolor[HTML]{B02418}{\tilde{\x}_{0}^{\,i}}
      \leftarrow
      \z - t\,\textcolor[HTML]{B02418}{s_i}$ \\
8 & \quad \textbf{end for} \\
9 & \quad $\z \leftarrow
      \z - d\,\textcolor[HTML]{B02418}{s_L}$ \\
10 & \textbf{end for}
\end{tabular}
}
\end{tcolorbox}
\vspace{-1.2em}
\end{wrapfigure}
Shortcut models~\citep{frans2025one} differ from MeanFlow in their task conditioning: instead of an endpoint time $r$, the network receives the current noise level $t$ and a step size $d$, so that $\boldsymbol{\tau}_\mathrm{task}=(t,d)$ and the endpoint is $r=t-d$ (Appendix~\ref{app:bg_shortcut}). We compute layerwise predictions (Eq.~\eqref{eq:layer_velocity_readout}) for a DiT-B/2 Shortcut model $s_\theta$ trained on ImageNet~\citep{russakovsky2015imagenet}, using the two-step schedule $1\rightarrow0.5\rightarrow0$, which corresponds to the two network evaluations $s_\theta(\z_1,1,0.5)$ and $s_\theta(\z_{0.5},0.5,0.5)$ (Alg.\hyperref[alg:shortcut_probe]{~E.1}). From the shortcut predicted at block $i$, $s_i$, we decode the endpoint prediction with the model's own update rule, $\textcolor[HTML]{B02418}{\tilde{\x}_{t-d}^{\,i}}=\z_t-d\,s_i$, and the diagnostic clean prediction $\textcolor[HTML]{B02418}{\tilde{\x}_{0}^{\,i}}=\z_t-t\,s_i$. The first evaluation therefore decodes blockwise predictions of the noisy state $\z_{0.5}$, and the second decodes blockwise predictions of $\z_0$.

\begin{figure}[t]
  \centering
  \includegraphics[width=0.9\textwidth]{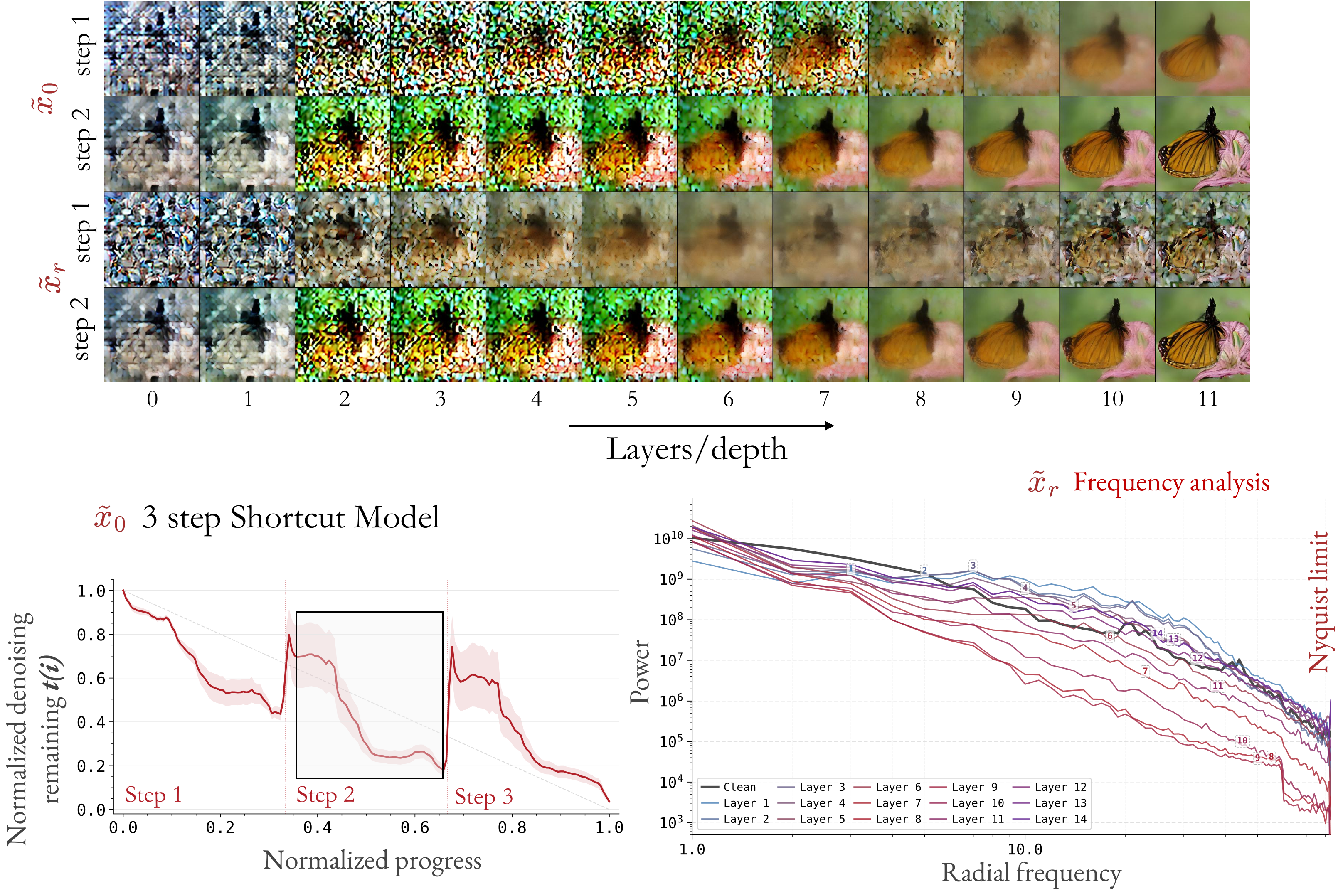}
  \caption{\textbf{Depth as time in a Shortcut model.}
  \emph{Top:} endpoint predictions $\textcolor[HTML]{B02418}{\tilde{\x}_{t-d}^{\,i}}$ across depth during a two-step rollout. Within the first step, whose endpoint is noisy, predictions first reveal cleaner structure and then return toward the prescribed endpoint.
  \emph{Bottom left:} normalized distance $\rho(i)$ computed from the diagnostic clean predictions $\textcolor[HTML]{B02418}{\tilde{\x}_{0}^{\,i}}$.
  \emph{Bottom right:} radially averaged power spectra of the endpoint predictions in the first step. Intermediate blocks move toward the clean-image spectrum before later blocks return toward the noisy endpoint spectrum.}
  \label{fig:shortcut}
  \vspace{-1.2em}
\end{figure}

\textbf{Observations.}
The Shortcut model shows the same denoise-then-renoise pattern as MeanFlow (Section~\ref{subsec:flow_map}). In the first step, whose endpoint $\x_{0.5}$ is noisy, the layerwise predictions first reveal cleaner structure and then return toward the endpoint in the final blocks (Fig.~\ref{fig:shortcut}, top), and $\rho(i)$ correspondingly decreases and then increases within the step (Fig.~\ref{fig:shortcut}, bottom left). Depth as time thus extends to step-conditioned flow maps.

\textbf{Frequency-domain signature.}
We also compute the radially averaged power spectrum of each endpoint prediction in the first step~\citep{field1987relations,simoncelli2001natural}: for block $i$, $P_i(\omega)=|\mathcal{F}[\textcolor[HTML]{B02418}{\tilde{\x}_{t-d}^{\,i}}](\omega)|^2$, averaged over frequencies of equal magnitude $\kappa=\|\omega\|_2$ to give $\bar{P}_i(\kappa)$. Intermediate blocks shift power toward the clean-image spectrum, and later blocks shift it back toward the noisier endpoint spectrum (Fig.~\ref{fig:shortcut}, bottom right), confirming the pattern in frequency space.

\end{document}